\documentclass[twoside,11pt]{article}
\usepackage{amsthm}  
\usepackage[abbrvbib]{jmlr2e}

\usepackage[english]{babel}
\usepackage[T1]{fontenc} 
\usepackage[utf8]{inputenc} 
\usepackage[babel]{microtype} 

\usepackage{amsmath,bm,mathtools}  
\usepackage{amsfonts}  
\usepackage{bbm}
\usepackage[dvipsnames]{xcolor}
\graphicspath{{fig/}}
\usepackage{enumitem}
\usepackage{booktabs}
\usepackage{siunitx}
\usepackage{wrapfig}
\usepackage{verbatim}  
\usepackage{lastpage}  

\usepackage{algorithm}
\usepackage[noend]{algpseudocode}

\newcommand{\ind}{\mathbbm{1}}

\newcommand{\Uc}{\mathcal{U}}

\providecommand{\assumptionname}{Assumption}

\providecommand{\factname}{Fact}
\providecommand{\conditionname}{Condition}
\providecommand{\theoremname}{Theorem}

\newcommand{\indep}{\perp\!\!\!\!\perp} 

\newcommand{\E}{\mathbf{E}}
\providecommand{\assumptionname}{Assumption}

\providecommand{\factname}{Fact}
\providecommand{\conditionname}{Condition}
\providecommand{\theoremname}{Theorem}

\theoremstyle{plain}
\newtheorem{Theorem}{\protect\theoremname}
\theoremstyle{plain}

\theoremstyle{plain}
\newtheorem{assumption}{\protect\assumptionname}
\theoremstyle{plain}

\def\balign#1\ealign{\begin{align}#1\end{align}}
\def\baligns#1\ealigns{\begin{align*}#1\end{align*}}
\def\balignat#1\ealign{\begin{alignat}#1\end{alignat}}
\def\balignats#1\ealigns{\begin{alignat*}#1\end{alignat*}}
\def\bitemize#1\eitemize{\begin{itemize}#1\end{itemize}}
\def\benumerate#1\eenumerate{\begin{enumerate}#1\end{enumerate}}

\newenvironment{talign*}
 {\csname align*\endcsname}
 {\endalign}
\newenvironment{talign}
 {\csname align\endcsname}
 {\endalign}

\def\balignst#1\ealignst{\begin{talign*}#1\end{talign*}}
\def\balignt#1\ealignt{\begin{talign}#1\end{talign}}

\newcommand{\pirnd}{\pi^{\TN{TS-rnd}}}
\newcommand{\piPS}{\pi^{\TN{TS}}}
\newcommand{\p}{p^{\TN{TS}}}
\newcommand{\prnd}{p^{\TN{TS-rnd}}}
\newcommand{\epsilonrnd}{\epsilon_{\TN{rnd}}}
\newcommand{\Sigmapost}{\Sigma_{\TN{post}}}
\newcommand{\mupost}{\mu_{\TN{post}}}

\newcommand{\NN}{\mathbb{N}}

\newcommand{\cdotspace}{\, \cdot \,}

\newtheorem{innercustomgeneric}{\customgenericname}
\providecommand{\customgenericname}{}
\newcommand{\newcustomtheorem}[2]{%
  \newenvironment{#1}[1]
  {%
   \renewcommand\customgenericname{#2}%
   \renewcommand\theinnercustomgeneric{##1}%
   \innercustomgeneric
  }
  {\endinnercustomgeneric}
}
\newcustomtheorem{customthm}{Theorem}
\newcustomtheorem{customlemma}{Lemma}
\newcustomtheorem{customprop}{Proposition}
\newcustomtheorem{customcond}{Condition}
\newcustomtheorem{customcor}{Corollary}

\DeclareMathOperator*{\argmax}{argmax} 

\newcommand{\TN}{\textnormal}
\newcommand{\argmin}{\TN{argmin}}
\newcommand{\iidsim}{\overset{\text{i.i.d.}}{\sim}}
\newcommand{\MC}{\mathcal}

\newcommand{\bo}{\textbf}
\newcommand{\bs}{\boldsymbol}

\newcommand{\real}{\mathbb{R}}

\newcommand{\by}{\times}
\newcommand{\under}{\underline}

\newcommand{\PP}{\mathbb{P}}

\newcommand{\HH}{\MC{H}}

\newcommand{\Var}{\TN{Var}}

\newlist{enuminline}{enumerate*}{1}
\setlist[enuminline,1]{label=\itshape\alph*\upshape)}

\usepackage{lastpage}
\jmlrheading{27}{2026}{1-\pageref{LastPage}}{1/25}{3/26}{25-0030}{Zhang, Baldwin-McDonald, Ciosek, Maystre, and Russo}

\ShortHeadings{Optimizing for the Long-Term Without Delay}{Zhang, Baldwin-McDonald, Ciosek, Maystre, and Russo}
\firstpageno{1}

\begin{document}

\title{Impatient Bandits:\\
Optimizing for the Long-Term Without Delay}

\author{%
  \name Kelly W. Zhang \email kelly.zhang@imperial.ac.uk \\
  \addr Imperial College London
  \AND 
  \name Thomas Baldwin-McDonald\thanks{Research done as part of an internship at Spotify.} \email tommcdonald955@gmail.com \\
  \addr University of Manchester
  \AND
  \name Kamil Ciosek \email kamilc@spotify.com  \\
  \addr Spotify
  \AND
  \name Lucas Maystre\thanks{Research done while at Spotify.} \email lucas@reflection.ai \\
  \addr Reflection AI
  \AND 
  \name Daniel Russo \email djr2174@gsb.columbia.edu\\
  \addr Columbia University
}

\editor{Tor Lattimore}

\maketitle

\begin{abstract}
Increasingly, recommender systems are tasked with improving users' long-term satisfaction. In this context, we study a content exploration task, which we formalize as a bandit problem with delayed rewards. There is an apparent trade-off in choosing the learning signal: waiting for the full reward to become available might take several weeks, slowing the rate of learning, whereas using short-term proxy rewards reflects the actual long-term goal only imperfectly. First, we develop a predictive model of delayed rewards that incorporates all information obtained to date. Rewards as well as shorter-term surrogate outcomes are combined through a Bayesian filter to obtain a probabilistic belief. Second, we devise a bandit algorithm that quickly learns to identify content aligned with long-term success using this new predictive model. We prove a regret bound for our algorithm that depends on the \textit{Value of Progressive Feedback}, an information-theoretic metric that captures the quality of short-term leading indicators that are observed prior to the long-term reward. We apply our approach to a podcast recommendation problem, where we seek to recommend shows that users engage with repeatedly over two months. We empirically validate that our approach significantly outperforms methods that optimize for short-term proxies or rely solely on delayed rewards, as demonstrated by an A/B test in a recommendation system that serves hundreds of millions of users.

\end{abstract}


\section{Introduction}

The multi-armed bandit problem stands as a cornerstone in machine learning, statistics, and operations research, crystallizing the challenge of learning to make effective decisions through intelligent trial and error. In its classical formulation, the model assumes immediate reward observation following an action, facilitating rapid adaptation. However, this instantaneous feedback structure often fails to capture the complexity 
of real-world applications, particularly in the realm of digital platforms serving millions or billions of users.

\begin{wrapfigure}{r}{0.45\textwidth}
    \vspace{-8mm}
  \begin{center}
    \includegraphics[width=0.4\textwidth]{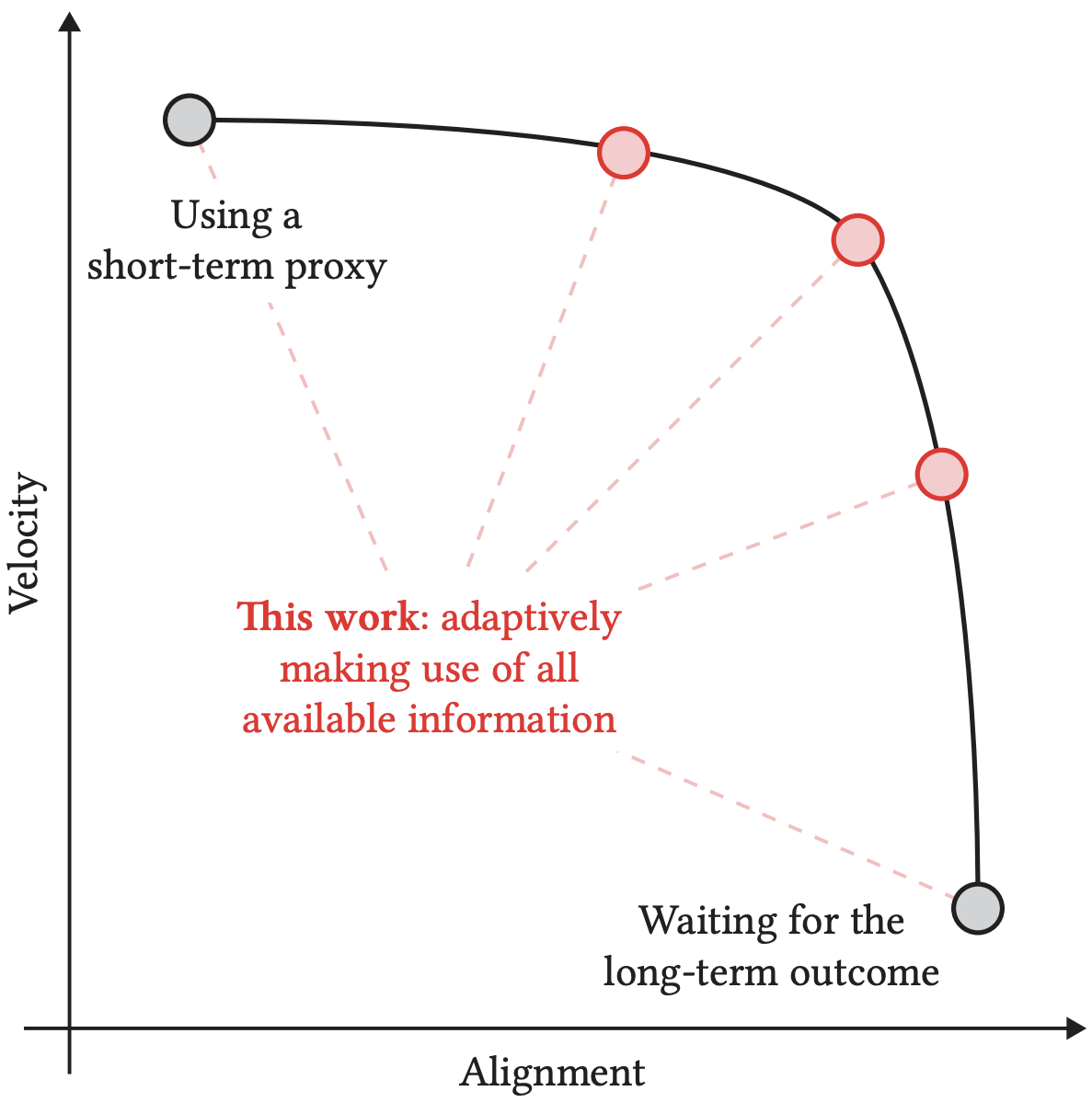}
  \end{center}
  \vspace{-5mm}
  \caption{Short-term outcomes enable a rapid feedback loop, but might be poorly aligned with long-term success metrics, which take longer to realize. Our method bypasses this apparent trade-off by adaptively making use of all available information at a given time.}
  \label{fig:tradeoff-curve}
  \vspace{-5mm}
\end{wrapfigure}

On these large-scale platforms, algorithms need to make decisions at a rapid pace, since the time between the start of one user's interaction and the start of the next user's interaction is extremely brief. Consequently, decision-making algorithms that optimize for rewards that are observed before the start of the subsequent user interaction (or even before a few interactions) inevitably skew towards extremely short-term goals, such as whether a user clicked on an advertisement. Selecting longer-term outcomes for the reward (like long-term engagement with a recommendation) better aligns with genuine long-term goals. However, waiting even a brief amount of time to observe the reward following each decision can greatly hinder the efficacy of bandit algorithms, since these brief delays can correspond to waiting an enormous number of interactions or “rounds” in classical bandit formulations. In Figure \ref{fig:tradeoff-curve}, we depict this apparent trade-off between speed of learning versus the alignment of the reward with the long-term outcome of interest.

In this work, we investigate the problem of optimizing for true long-term goals while mitigating the impact of delay. Our key insight is that most long-term outcomes become increasingly predictable over time. In the context of digital platforms, a user's long-term engagement is revealed 
incrementally. For instance, total spending is signaled by initial purchases, and the likelihood of conversion following a marketing message diminishes as time passes without a response. 
We term this phenomenon “progressive feedback,” distinguishing it from the widely studied delayed feedback, where no information is gleaned until long after an action is taken.

\begin{figure}
    \centering
    \includegraphics[width=0.85\linewidth]{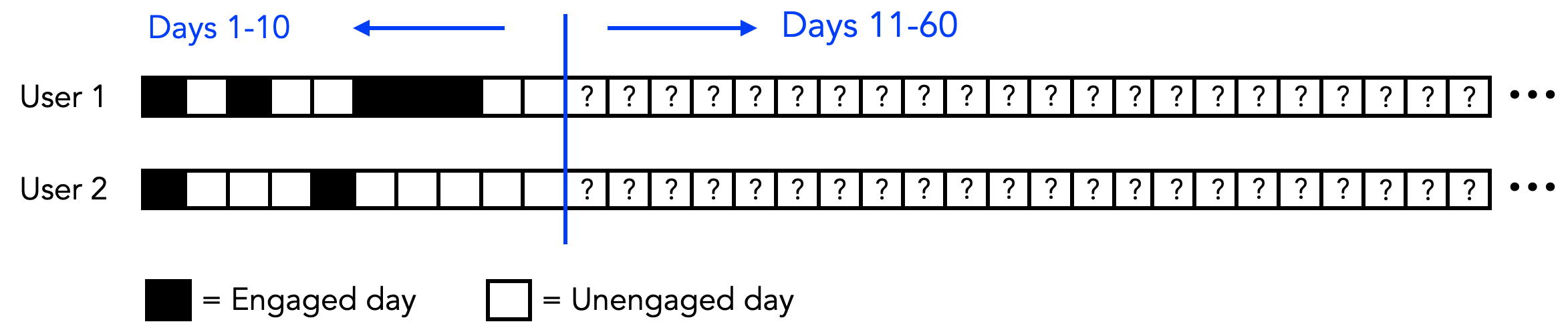}
    \caption{Progressive engagement feedback from two hypothetical users. The long-term outcome of interest is users' average engagement over the $60$ days following a recommendation. Each user's long-term outcome is increasingly predictable as more engagement information is revealed.}
    \vspace{-5mm}
    \label{fig:engagementOverTime}
\end{figure}

Our interest in this problem stems from a real challenge encountered at Spotify, a leading audio streaming service. Following a content recommendation decision, engagement feedback from each user is progressively revealed; we illustrate examples of the progressive engagement feedback in Figure \ref{fig:engagementOverTime}. Recent efforts to optimize for longer-term metrics, such as user engagement over a 60-day period, brought substantial benefits to the recommendation system, as compared to optimizing for shorter-term outcomes \citep{maystre2023optimizing,minmin-surrogates,rl-longterm-engagement,yang2024targeting}. 
However, these long-term metrics are only realized after a significant delay, meaning such data is never available for recently released content. 

In this context, we focus on the cold-start problem---the challenge of rapidly learning to make effective recommendations for newly released content with little to no historical data. The core challenge lies in optimizing for the 
long-term outcome 
and identifying content with superior long-term performance, but doing so without waiting for these long-term metrics to become fully revealed. 
We address this problem in a novel way, by training a (empirical) Bayesian filtering model that draws inferences about content quality and appeal from progressively revealed user engagement. A/B tests have shown that incorporating progressive feedback has large benefits in practice (see Section \ref{sec:abtest}), and this idea is now used in production as a core part of the recommender system at Spotify, which powers personalized audio recommendations for hundreds of millions of users.

This paper develops a general model of multi-armed bandit problem with progressive feedback, which distills the core challenge of the problem encountered at Spotify. We discuss the contributions of the paper in detail in the next subsection.

\subsection{Contributions}
\label{subsec:contributions}

This work expands upon the algorithm originally presented in \cite{impatientKDD}. Key additions include
\begin{enuminline}
\item a formalization of the model,
\item a novel regret analysis,
\item A/B test results from an implementation of the algorithm at Spotify, and
\item several additional synthetic experiments.
\end{enuminline}

\paragraph{Model of bandits with progressive feedback.}
Motivated by this real-world problem faced at Spotify and recognizing its broader relevance, we formulate a model termed \emph{bandits with progressive feedback} (Section~\ref{sec:problemFormulation}).
In this setting, the reward from selecting an ``arm'' is fully observed only after a significant delay but becomes progressively more predictable over time. We primarily focus on a key special case where rewards are a linear function of a set of correlated Gaussian engagement outcomes.
The crux of the problem lies in the interplay between the delay structure, where some outcomes are revealed much earlier than others, and the correlation structure, which determines the extent to which initial observations are predictive of later ones.
In our problem setting, the algorithm confronts two sources of limited feedback. The first is \emph{bandit feedback} (in contrast to \emph{full feedback}), where the algorithm only has data on items it has recommended in the past, necessitating active exploration for learning.
The second is \emph{progressive feedback}, where engagement outcomes are revealed gradually over time, introducing delay in the resolution of uncertainty regardless of the number of users receiving recommendations.

\paragraph{Impatient bandit algorithm.}
To address the above challenges within our model, we propose an algorithm that integrates two key components: Thompson sampling and a (empirical) Bayesian filter (Section~\ref{sec:algorithm}).
Thompson sampling \citep{thompson1933likelihood} is employed to tackle the bandit feedback challenge. 
The Bayesian filter is designed to mitigate the issues arising from progressively revealed feedback.
The Bayesian filtering model projects the true average reward of each based on partially observed engagement trajectories, maintaining appropriate residual uncertainty.
Both mean and uncertainty estimates are incrementally updated as further feedback is observed.
Our model assumes this filter is computed based on highly informed prior beliefs.
In practice, this rich prior comes from historical data, by training the hyperparameters of the Bayesian filter on complete trajectories of user interactions with historical content (Section \ref{sec:prior}).

\paragraph{Regret analysis and the value of progressive feedback.} 
We derive a novel regret bound for this algorithm that provides rigorous insight into the value of progressive feedback (Section~\ref{sec:theory}).
When there is no useful progressive feedback, our regret bound reduces to a standard bound for Thompson sampling with delayed feedback.
With progressive feedback, our regret bound improves based on what we term the \emph{value of progressive feedback}.
This term equals the mutual information between censored (i.e., delayed) user-engagement trajectories and the true reward observations that will eventually materialize.
Note that this mutual information term is entirely a property of the underlying data generating environment itself, and is not impacted by the decision-making algorithm.
High mutual information indicates that early feedback substantially resolves uncertainty in long-term rewards. 

Importantly, our focus differs from traditional bandit analyses that emphasize how quickly regret vanishes as the number of arm selections grows.
In settings with severely delayed rewards, such as our motivating application where the primary reward metric is computed after 60 days, it is critical to learn something meaningful before \emph{any} rewards are fully observed.
This is challenging regardless of the number of arm selections made during that time frame.
Our theoretical and practical objective, therefore, is to take advantage of progressive feedback to significantly improve decision quality compared to scenarios without such feedback.
Our algorithm achieves this:
We prove that it incurs low expected regret in any batch when the value of progressive feedback is high, even before long-term rewards are observed.

\paragraph{Semi-Synthetic Experiments and A/B Test Results.}
Experiments with a synthetic data generating process show that our theory aligns qualitatively with performance across a range of problems with varying true ``values of progressive feedback'' (Section~\ref{sec:experiments}).
We also include experiments that utilize podcast recommendation data from Spotify; these simulation results suggest that there can be substantial value in incorporating progressive feedback in real life applications.
Additionally, we present results from a large-scale A/B test that was run at Spotify (Section~\ref{sec:abtest}).
The A/B test compares a recommendation algorithm that incorporated progressive feedback to one that did not.
The results illustrate the significant benefit that leveraging progressive feedback can have in complex, industrial-scale recommender systems.

\section{Related Work}
\label{sec:relatedWork}

\paragraph{Surrogate Outcomes.}
The \emph{surrogate outcomes} literature uses short-term outcomes as a proxy for long-term outcomes.
For example, for a treatment aimed at preventing heart disease, changes in cholesterol level may be used as a surrogate outcome for the distal outcome of developing heart disease in the next year.
Most of this literature focuses on estimating causal effects, rather than decision-making.
There are three general approaches to using surrogate outcomes:
\begin{enuminline}
\item assume a causal relationship between the surrogate and long-term outcome, and effectively replace the distal outcome with surrogates \citep{athey2016estimating,duan2021online,prentice1989surrogate,fleming2012biomarkers,weintraub2015perils,vanderweele2013surrogate},
\item use surrogates that are predictive of the long-term outcome to improve imputation of missing distal outcomes \citep{kallus2020role,cheng2021robust}, and
\item use surrogates that are predictive of the long-term outcome to fit a joint Bayesian model of these outcomes \citep{tripuraneni2024choosing,richardson2023pareto,anderer2022adaptive}.
\end{enuminline}
Our work is most similar in spirit to that last approach, with a focus on decision-making.

\paragraph{Optimizing for Long-Term Outcomes in Recommendation Systems.}
In the recommendation systems literature, there are two foremost approaches to designing algorithms that optimize for long-term outcomes.
The first focuses on designing algorithms that account for the delayed effects of actions on the same users over time, e.g., by moving beyond bandit algorithms and designing algorithms that account for how a recommendation decision today may affect that same user's future engagement \citep{auction-long-term,returning-is-believing,deeprl-news-rec,lifetime-coldstart}.
The second approach involves designing algorithms that optimize for long-term, delayed rewards and incorporate short-term outcomes in some way to speed up learning, e.g., by using the short-term outcome as a surrogate or combining short-term and long-term outcomes in the definition of the reward \citep{minmin-surrogates,rl-longterm-engagement,short-and-long,counterfactual-delayed};
None of these existing works focus on the cold-start recommendation setting in which there is little to no historical data for certain actions (the focus of this work).
With the exception of \cite{counterfactual-delayed}, which we discuss below, these existing works develop offline RL methods, which use large datasets in which each action has been played many times already.

\paragraph{Online Decision-Making Algorithms that use Intermediate Outcomes.}
Delayed rewards are a significant challenge in online decision-making settings, and many bandit algorithms have been developed specifically to accommodate such delays \citep{howson2023delayed,mandel2015queue,desautels2014parallelizing,NEURIPS2019_ae2a2db4,thune2019nonstochastic,vernade2020linear,pmlr-v80-pike-burke18a,joulani2013online,NEURIPS2019_56cb94cb}.
We approach the delayed rewards problem by incorporating intermediate, surrogate outcomes that may be predictive of the reward. 
Several previous works also consider algorithms with access to intermediate outcomes, but consider different problem settings and/or make different assumptions on the type of intermediate feedback.

\citet{anderer2022adaptive} use surrogate outcomes to design a Bayesian adaptive clinical trial which adaptively decides whether to stop the trial at a pre-specified decision point, subject to a power constraint. \citet{wu2022partial} develop a Thompson Sampling algorithm for survival outcomes based on the proportional hazards model, which leverages incrementally revealed information on how long individuals have survived so far. 
\citet{grover2018best} consider a UCB algorithm for best arm identification in a setting with stochastically delayed rewards in which the next decision cannot be made until the delayed reward is observed; this delay structure differs fundamentally from ours. Moreover, they assume that their algorithm sees i.i.d. intermediate outcomes (which may be biased or unbiased estimates of the mean reward) while it waits, which does not accommodate progressively revealed user engagement feedback (see Figure \ref{fig:engagementOverTime}), which critically are dependent over time.

\citet{yang2024targeting} apply a surrogate index approach \citep{athey2016estimating} to develop a Thompson sampling algorithm that optimizes a fitted function of short-term outcomes instead of the true long-term reward (customer retention). While they assume short-term outcomes act as perfect surrogates for the long-term reward, our approach incorporates \textit{imperfect} leading indicators.
Finally, \cite{counterfactual-delayed} use importance weighting to ensure the short-term surrogate is an unbiased estimate of the mean, long-term reward. Their algorithm (developed for a setting with binary outcomes and a single intermediate outcome) does not easily extend to settings with many intermediate outcomes ($> 50$ in our applications) that are continuous, i.e., where estimating propensities to form importance weights is more difficult.

\paragraph{Distinguishing features of our work.}
Our work advances the literature in several key ways.
First, unlike most existing models which treat feedback as either delayed or immediate, or assume perfect surrogate outcomes, we formalize a model of \emph{progressive feedback} where information becomes incrementally more predictive over time.
Our model addresses the practical challenge of synthesizing information from many ($> 50$) sequentially revealed intermediate signals that serve as imperfect leading indicators rather than perfect surrogates of true rewards.
Second, we introduce a novel Gaussian filtering algorithm specifically designed to synthesize information from numerous progressively revealed outcomes.
This approach not only predicts rewards but crucially maintains appropriate uncertainty estimates throughout the learning process, preventing premature convergence to suboptimal decisions.
While this uncertainty-aware approach shares some features with survival modeling work \citep{wu2022partial}, our method uniquely handles multiple continuous intermediate signals.
Third, we introduce a novel information-theoretic approach to quantifying progressive feedback value and prove novel regret bounds based on this measure.
Finally, our approach has demonstrated significant impact within a large-scale industrial recommender system, validating its practical effectiveness in handling the complexities of real-world progressive feedback at scale.

\section{Problem Formulation}
\label{sec:problemFormulation}
We now formally describe our problem setup, which is motivated by recommender systems applications.
In our problem setup, a large batch of actions (items) are selected at each decision time. 
Specifically, each decision time $t \in \NN$ represents a day during which many users, denoted by the set $\Uc_t$, are simultaneously presented with item recommendations.
Note that the sets $\{ \Uc_t \}_{t \in \NN}$ are disjoint and we use $\Uc \triangleq \cup_{t \in \NN} \ \Uc_t$ to denote the collection of all users.
For simplicity, we assume that the batches $\Uc_t$ are of the same size each day;
We use $m = |\Uc_1| = |\Uc_2| = |\Uc_3| = \dots$ to denote the batch size.
While batching is common in practice for operational reasons, we employ it primarily as a device to clearly distinguish between two crucial aspects of the problem.
The first aspect is the passage of time, which is represented by the number of batches.
The second is the rate at which exploratory recommendations are made, which is determined by the size of each batch.
This model enables us to separately vary the pace of decision-making from the pace at which outcomes are revealed (amount of delay).

Each day $t \in \NN$, for each user $u \in \Uc_t$ in the batch, the system selects an item $A_u$ from a finite collection of items $\MC{A}$ to recommend ($\MC{A}$ is the action space). Each item $a \in \MC{A}$ has an associated item feature vector $Z_a \in \MC{Z}$ for $|\MC{Z}| < \infty$. 
For example, in the podcast recommendation setting, $Z_a$ may include the category of the podcast show (comedy, news, lifestyle). The item features are used as they may be predictive of or correlated with the item's performance. Note that this vector is not a typical context or state vector, rather it represents features of the \textit{item} (action). 
Note that we are able to extend the algorithm we present to settings with user contexts (see Appendix \ref{app:contextVersion} for details); throughout the main body of this paper we focus on the setting without context for clarity.

Upon recommending an item $A_{u} \in \MC{A}$ for a user $u \in \Uc_t$, the algorithm subsequently observes a collection of user outcomes $Y_u^{(1)}, Y_u^{(2)}, \ldots, Y_u^{(J)}$.
We use $Y_u \triangleq (Y_u^{(1)}, Y_u^{(2)}, \dots, Y_u^{(J)}) \in \real^J$ to refer to the entire vector of engagement outcomes observed from selecting action $A_u$.
In our setting, these outcomes are progressively revealed over time.
Specifically, in our application, $\big( Y_u^{(j)} \big)_{j=1}^J$ refer to measures of engagement with the recommended item that are revealed over $J$ days. An engagement trajectory is associated with a long-term reward  $R(Y_u)$ by applying a fixed, known, real-valued, reward function $R(\cdotspace)$. For example, the reward could represent the total user engagement over time, $R(Y_u) = \sum_{j=1}^J Y_u^{(j)}$, or if the outcome of interest is the final engagement, 
one could choose $R(Y_u) = Y_u^{(J)}$. In general, the outcomes $\big( Y_u^{(j)} \big)_{j=1}^J$ could be a series of predictions of a long-term user outcome from a previously trained model or simply leading indicators that are correlated with a long-term metric of interest. 

Formally, we use $d_j$ to refer to the deterministic ``delay'' in seeing the $j^{\TN{th}}$ outcome $Y_u^{(j)}$. For example, if $d_1 = 0, d_2 = 1, \dots, d_J = J-1$ and user $u$ is in batch $t$, then $Y_u^{(1)}$ is revealed immediately after selecting the action $A_u$ at time $t$. Following decision time $t+1$, additional engagement information $Y_u^{(2)}$ is observed, and so on until following decision time $t + J -1$ the final engagement outcome $Y_u^{(J)}$ is revealed.

\subsection{Counterfactual Outcomes and Learning Objective}
\label{sec:counterfactuals}

We use the potential outcomes framework \citep{imbens2015causal,rubin1974estimating} to formally represent counterfactual outcomes. We assume there exist latent variables
\begin{align*}
    \left\{ Y_u(a) = \big( Y_u^{(1)}(a), Y_u^{(2)}(a), \dots, Y_u^{(J)}(a) \big) \right\}_{a\in \MC{A}, u \in \Uc}
\end{align*}
called \emph{potential engagement outcomes}, such that the engagement outcome of selecting action $A_u = a$ for some $a \in \MC{A}$ is $Y_u(a) = \big( Y_u^{(1)}(a), Y_u^{(2)}(a), \dots, Y_u^{(J)}(a) \big)$;  the observed outcome is $Y_u \triangleq Y_u(A_u)$.
This means that the potential engagement outcome is a function of only the recommendation that user receives and not the recommendations received by other users, ruling out, for instance, complicated social network effects.\footnote{%
This assumption is called the Stable Unit Treatment Value Assumption (SUTVA) in the causal inference literature.
A growing literature on experimentation in the presence of interference tries to make sensible inferences when this assumption is relaxed \citep{shuangningInterference,han2023detecting}.}

The next assumption formalizes a symmetry of the distribution of outcomes among users in the cohort $\Uc$.
\begin{assumption}[Exchangeability] 
    \label{assum:exchangeable}
    For any fixed item $a \in \MC{A}$, the potential outcome vectors $\big( Y_u(a) \big)_{u \in \Uc}$ are exchangeable over users $u \in \Uc$ conditional on $Z_a = z$ for any $z \in \MC{Z}$.
    That is, the joint distribution satisfies
    \begin{align*}
        \big( Y_u(a) \big)_{u \in \Uc} \mid (Z_a = z) ~~~
        \overset{\text{D}}{=} ~~~
         \big( Y_{\sigma(u)}(a) \big)_{u \in \Uc} \mid ( Z_a = z ) 
    \end{align*}
    for any permutation $\sigma$ over $\Uc$.
\end{assumption}

\begin{remark}[Interpreting Exchangeability]
    \label{rem:exchangeable} 
    To interpret this assumption, it is helpful to think of $\Uc=\{1,2,3\ldots\}$ as a list of user IDs that are randomly assigned to users in the cohort. Then, because of the random assignment, there is nothing a priori to differentiate the distribution of outcomes for two users who receive the same recommendation at the same time (i.e., $u,u' \in \Uc_t$ with $A_u=A_{u'}$). This part of the assumption is not a major restriction. The meaningful assumed structure is a symmetry in outcomes among users who receive the same recommendation at different points in time. This is critical to allowing us to learn from initial recommendations how to make recommendations to future batches of users.
\end{remark}

By De Finetti's theorem \citep{heath1976finetti,de1937prevision}, the exchangeability assumption is equivalent to assuming that there exists a random, latent (i.e., unobserved) variable $\theta_a$ such that the user's potential engagements $\big\{ Y_u(a) \big\}_{u \in \Uc}$ are i.i.d. given $\theta_a, Z_a$. This means there exists a distribution $P_{\theta_a, Z_a}$ such that 
\begin{align*}
    Y_1(a), Y_2(a), Y_3(a), \dots \mid \theta_a, Z_a \iidsim P_{\theta_a, Z_a}.
\end{align*}
While $Z_a$ captures an item's observed features, we can think of $\theta_a$ as reflecting latent features of the item---like its quality and style---which cannot be inferred prior to recommending it.
Intuitively, user responses $Y_u(a)$ are not i.i.d. (only \emph{conditionally} i.i.d.) because observing $Y_u(a)$ gives one more information about the latent $\theta_a$, which reduces uncertainty in $Y_{u'}(a)$ for some user $u' \not= u$. 

Recall that the reward $R(Y_u)$ is a fixed, known function of the engagement trajectory $Y_u$.
Define the expected reward from deploying an item $a$ in the population by 
\begin{align}
    \bar{R}_a \triangleq \E \left[ R\big( Y_u(a) \big) \mid Z_a, \theta_a \right].
    \label{eqn:RbarDef}
\end{align}
The above expectation averages over the randomness in $Y_u(a)$ given  $Z_a$ and $\theta_a$, which determine the outcome distribution for that item, $P_{\theta_a, Z_a}$. We define the reward-maximizing item by 
\begin{align*}
    A^* \triangleq \arg\max_{a\in \MC{A}} \{ \bar{R}_a \}.
\end{align*}
If the best item is not unique, we simply set $A^*$ to be one of the optimal items in $\MC{A}$.

A \emph{policy} $\pi$ is a (possibly randomized) rule that at each decision time $t$ makes recommendation decisions $A_u$ to the users in batch $t$ as a function of the history of observations available to it, $\HH_{t}$. Due to delayed feedback, we need to develop some extra notation to define $\HH_{t}$ precisely. We use $\tau_u \in \NN$ to refer to the decision time at which the user $u \in \Uc$ receives their recommendation. That is, $u\in \Uc_{\tau_u}$. Using this notation, we define a \emph{censored} version of the potential engagement outcome $Y_{u}(a)$ as follows:
\begin{align}
    \tilde{Y}_{u,t}^{(j)} \triangleq 
    \begin{cases} 
        Y_u^{(j)} & \text{ if } t > \tau_u + d_j,   \\
        \mathrm{Null} & \text{ otherwise}.
    \end{cases} 
    \label{eqn:censoredYs}
\end{align}
Define $\tilde{Y}_{u,t} \triangleq (\tilde{Y}_{u,t}^{(1)}, \tilde{Y}_{u,t}^{(2)}, \dots, \tilde{Y}_{u,t}^{(J)})$.
At decision time $t$, the system must base its decisions on $\{ Z_a \}_{a \in \MC{A}}$ and on the censored history
\[
    \HH_{t} \triangleq
    \bigcup_{u \in \Uc_{<t}} \left\{ A_{u}, \tilde{Y}_{u,t}^{(1)}(A_u), \tilde{Y}_{u,t}^{(2)}(A_u), \dots, \tilde{Y}_{u,t}^{(J)}(A_u) \right\} \cup \{ Z_a : a \in \MC{A} \}.
\]
Above, $\Uc_{<t} \triangleq \bigcup_{t'=1}^{t-1} \Uc_{t'}$. That is, the censored history at decision time $t$ contains the activity measurement $Y_u^{(j)}(a)$ if and only if user $u$ was recommended item $a$ more than $d_j$ periods ago. The above implies that $Y_u^{(j)}$ is observed when $t > \tau_u + d_j$. 
Define the instantaneous regret for a policy $\pi$ at batch $t$,
\begin{align}
    \label{eqn:DeltaDef}
    \Delta_t(\pi) \triangleq \frac{1}{|\Uc_t|} \sum_{u \in \Uc_t} \bigg\{ R \big(Y_u(A^*)\big)  - R\big( Y_u(A_u) \big) \bigg\}.
\end{align}
Our regret bounds will control the “average” regret. Next we discuss exactly what we mean by average regret and how it relates to the cold start problem.

\paragraph{Cold Start and Bayesian Regret.}
The cold-start problem is faced on a recurring basis by the company (e.g., new content is released each day).
When new content is released, we view it as a random draw from some distribution over possible items.
This means that when a new item $a \in \MC{A}$ is released, we can think of observed and latent item features $Z_a$ and $\theta_a$ respectively (which determine $P_{\theta_a, Z_a}$, the response distribution for the item $a$) as being drawn from some underlying distribution.
Assumption \ref{assum:items} below formalizes this.

\begin{assumption}[Randomly Drawn Items]
    The items in $\MC{A}$ (note $|\MC{A}| < \infty$) are drawn i.i.d. from a distribution of possible items. Mathematically, this means that the observed and latent item features, $\{ Z_a, \theta_a \}_{a \in \MC{A}}$, are i.i.d. over $a \in \MC{A}$.
    \label{assum:items}
\end{assumption}

Mathematically, the consequence of Assumption \ref{assum:items} is that the random item response distributions $\big\{ P_{\theta_a, Z_a} \big\}_{a \in \MC{A}}$ are drawn i.i.d. over $a \in \MC{A}$. We call the distribution of latent features $\theta_a$ given the observed features $Z_a$ the \textit{prior} distribution for $\theta_a$. In fact the data generating process can be thought of as a hierarchical model wherein for an item $a \in \MC{A}$, (i) $Z_a$ is sampled, (ii) $\theta_a \mid Z_a$ is sampled from a prior distribution that depends on $Z_a$, and (iii) subsequently potential response outcomes are sampled conditionally i.i.d. as $Y_1(a), Y_2(a), \dots \mid \theta_a, Z_a \iidsim P_{\theta,Z_a}$.

We call the following the expected instantaneous regret (or \emph{Bayesian regret}) for a given policy $\pi$ in batch $t$:
\begin{align}
    \E \left[ \Delta_t(\pi) \right]     &= \E_\pi \left[ \frac{1}{|\Uc_t|} \sum_{u\in \Uc_t} \bigg\{ R \big(Y_u(A^*) \big)  - R\big(Y_u(A_u) \big) \bigg\} \right]    \label{eqn:ExpectDeltaDef} \\
    &= \E \left[ \E_\pi \left[ \frac{1}{|\Uc_t|} \sum_{u\in \Uc_t}  \bigg\{ R \big(Y_u(A^*) \big)  - R\big(Y_u(A_u) \big) \bigg\} ~ \bigg| ~ \{ Z_a, \theta_a \}_{a \in \MC{A}} \right] \right].
    \label{eqn:ExpectDeltaDef2}
\end{align}
Above, we use $\E_{\pi}$ to denote the expectation when policy $\pi$ is used to select actions and the outer expectation in \eqref{eqn:ExpectDeltaDef2} averages over the draw of items $\{ Z_a, \theta_a \}_{a \in \MC{A}}$.
Note that the above expectation averages over the randomness in
\begin{enuminline}
\item the draw of the observed and latent item features, $\{ Z_a, \theta_a \}_{a \in \MC{A}}$, which determines the distribution $\{ P_{\theta_a, Z_a} \}_{a \in \MC{A}}$,
\item the draw of potential outcomes $Y_u(a)$ from $P_{\theta_a, Z_a}$ given $Z_a, \theta_a$, and
\item the recommendation decisions $A_u$ made by the (possibly randomized) policy $\pi$. 
\end{enuminline}

Our Bayesian approach is not merely a philosophical choice, but a pragmatic one, reflecting the recurring nature of cold-start problems in recommender systems. The expected value in \eqref{eqn:ExpectDeltaDef} can be interpreted as the long-run regret incurred across many instances of this problem, faced repeatedly as new content is released. This interpretation aligns with the practical reality of recommendation systems, where new items are continually introduced. This recurring structure also motivates our theoretical assumption that the algorithm has access to an informed prior distribution. In practice, we construct this informed prior by leveraging historical item engagement data from previously recommended items not in the current set $\MC{A}$; See Section \ref{sec:prior}. This approach lets us leverage patterns in historical data to inform current recommendations. For instance, we might infer immediately that an item is likely of low quality given its features $Z_a$, or might infer that an item is likely to offer high long-term reward given the strong recurring engagement patterns users have displayed over a shorter time period.

A common objective, roughly speaking, is to minimize the expected instantaneous regret averaged over $T$ batches:
\begin{align}\label{eq:_per_user_regret}
    \frac{1}{T} \sum_{t=1}^T \E[ \Delta_t(\pi) ].
\end{align}
Our upper bounds are more refined and also bound $\E[\Delta_t(\pi)]$ for \emph{every} batch of users $t$. Rather than focus on attaining vanishing regret as the number of batches scales, our focus is on achieving fairly low regret after a short timespan (low $t$ or $T$), well before all rewards are fully observed. In our recommender system application, this translates to an ability to effectively recommend content soon after its release. Our theory confirms that our algorithm can leverage progressive feedback to learn far more rapidly than would otherwise be possible in settings with severely delayed rewards.

\subsection{Gaussian distribution}
\label{sec:gaussian}

Next, we introduce a Gaussian assumption on the outcomes $Y_u^{(j)}$.
This assumption enables computing a posterior distribution in closed form.
Additionally, we will prove formal regret bounds under this Gaussian assumption.
Note, however, that the algorithm we develop can be generalized beyond Gaussian outcomes. 

\begin{assumption} 
    For any action $a \in \MC{A}$, the potential outcomes $\big( Y_u(a) \big)_{u\in \Uc}$ are a Gaussian process. 
    That is, for any finite subset of  $\Uc' \subset \Uc$, the distribution of $\big( Y_u(a) \big)_{u\in \Uc'}$ is multivariate Gaussian. Furthermore, assume $R(y)=r_0 + r_1^\top y$ for $r_0 \in \real$ and $r_1 \in \real^J$ is an affine, real-valued function so $R(Y_u(a))$ follows a Gaussian distribution.
    \label{assum:gaussian}
\end{assumption}
A Gaussian and exchangeable model is very special and is completely defined by the mean and covariance of the joint distribution of $\left( Y_u(a) , Y_{u'}(a) \right)$ for any $u \neq u'$ \citep{aldous2006exchangeability}. For arbitrary $u, u'\in \Uc$ with $u \neq u'$, we can write 
\begin{align*}
    \begin{bmatrix} Y_u(a) \\ Y_{u'}(a) \end{bmatrix} ~ \bigg| ~ ( Z_a = z )
    \sim N\left( \begin{bmatrix} \mu_{1, z} \\ \mu_{1, z} \end{bmatrix} \, , \, 
    \begin{bmatrix} \Sigma_{1, z} + V_z & \Sigma_{1, z} \\ \Sigma_{1, z} & \Sigma_{1, z} + V_z
    \end{bmatrix}  \right)
\end{align*}
where $\mu_{1, z} \triangleq \E \left[ Y_u(a) \mid Z_a = z \right] \in \real^J$ is the mean conditional on $Z_a = z$ (marginal over the latent features $\theta_a$), $\Sigma_{1, z} + V_z \triangleq {\rm Cov} \left( Y_u(a) \,, \, Y_{u}(a) \mid Z_a = z \right) \in \real^{J \by J}$ is the conditional covariance matrix of a single user's engagement metrics, and $\Sigma_{1, z} \triangleq {\rm Cov} \left( Y_u(a) \,, \, Y_{u'}(a) \mid Z_a = z \right)$ for $u \not= u'$ is the cross-user covariance. 

\begin{lemma}[Multivariate Gaussian Bayesian Model]
    \label{lem:joint_normal}
    Let Assumptions \ref{assum:exchangeable} and \ref{assum:gaussian} hold. The limit $\theta_a \triangleq \lim_{|U| \to \infty} \frac{1}{|U|} \sum_{u \in U} Y_{u}(a)$ exists almost surely. Conditional on $\theta_a \in \real^J, Z_a \in \MC{Z}$, the random variables $\left\{ Y_u(a) \right\}_{u\in \Uc}$ are i.i.d. 
    Moreover,
    \[
        \theta_a \mid ( Z_a = z ) \sim N \left( \mu_{1, z} \,,\, \Sigma_{1, z} \right) \qquad  \text{and} \qquad Y_u(a) \mid (\theta_a, Z_a = z) \sim N( \theta_a \,,\, V_z ). 
    \]
\end{lemma}
By Lemma \ref{lem:joint_normal} above, the latent parameter $\theta_a$ can be interpreted as the (random) almost sure limit of the empirical mean outcomes.
By the above lemma, the outcome distribution $P_{\theta_a, Z_a}$ that we defined earlier is a multivariate Gaussian $N(\theta_a, V_{Z_a})$. The proof of Lemma \ref{lem:joint_normal} is in Appendix \ref{app:gaussianModel}.

In practice, we use historical data from previously recommended items to fit the prior distribution parameters $\mu_{1,z}, \Sigma_{1,z}$, as well as the noise covariance matrix $V_z$;
We discuss our procedure for fitting these parameters in practice in Section \ref{sec:prior}.

\section{Algorithm Overview}
\label{sec:algorithm}

The \emph{impatient bandit} algorithm we develop next builds on the classical Thompson sampling bandit algorithm \citep{russo2020tutorial,thompson1933likelihood}.
However, our approach generalizes the classical Thompson sampling algorithm to take advantage of progressive feedback in settings with delayed rewards.
In particular, our algorithm does this by forming a posterior for the vector $\theta_a$, which, according to Lemma \ref{lem:joint_normal}, defines the mean of all the engagement outcomes $(Y^{(1)}_u(a), \ldots ,Y_u^{(J)}(a))$, rather than just the reward. This allows us to update the posterior distribution even after observing partial feedback. Then at decision time, we sample a random draw from the posterior of $\theta_a$, and use that posterior sample to compute a posterior draw of the mean reward under that arm using the function $R$ (which is linear by Assumption \ref{assum:gaussian}). 
See Algorithm \ref{alg:bandit} for more details.
Note that the posterior update rule used in line 11 of Algorithm \ref{alg:bandit} can be derived using standard formulas (see further discussion in Appendix \ref{app:posterior}). Additionally, we extend our algorithm to settings with user context features, where expected outcomes are linear functions of the context; See Appendix \ref{app:contextVersion} for details.

\begin{algorithm}
    \caption{Impatient Bandit Thompson Sampling $\piPS$}
    \begin{algorithmic}[1]
    \Require Priors and noise covariance matrices $\big\{ \mu_{1,Z_a}, \Sigma_{1,Z_a}, V_{Z_a} \big\}_{a \in \MC{A}}$
    \For{$a \in \MC{A}$}
        \State $(\mu_{1,a}, \Sigma_{1,a}) \gets (\mu_{1, Z_a}, \Sigma_{1, Z_a})$ \Comment{Initialize beliefs.}
    \EndFor
    \State $\MC{H}_1 \gets \{Z_a : a \in \MC{A}\}$ \Comment{Initialize history.}
    \For{$t = 1, 2, \dots, T$}
        \For{$u \in \Uc_t$}
            \State $\tau_u \gets t$ \Comment{Set user decision time.}
            \For{$a \in \MC{A}$} \label{line:startsample}
                \State $\theta_a' \sim N \left( \mu_{t,a} \,,\, \Sigma_{t,a} \right)$ \Comment{Sample from belief.}
            \EndFor
            \State $A_u \gets \argmax_{a \in \MC{A}} \left\{ R( \theta_a' ) \right\}$ \Comment{Take action.}
            \label{line:takeaction}
        \EndFor
        \State $\MC{H}_{t+1} \gets \MC{H}_t \cup \left\{ Y_u^{(j)}(A_u) : u \in \Uc_{\leq t}, \text{$j$ s.t. $t = \tau_u + d_j$} \right\}$ \Comment{Update history.}
        \For{$a \in \MC{A}$}
            \State $(\mu_{t+1,a}, \Sigma_{t+1,a}) \gets \textrm{PosteriorUpdate}(\mu_{t, Z_a}, \Sigma_{t, Z_a}, V_{Z_a}, \MC{H}_{t+1})$ \Comment{Update belief.}
        \EndFor
    \EndFor
    \end{algorithmic}
    \label{alg:bandit}
\end{algorithm}

\subsection{Lower Variance Version of Algorithm}

For our regret bounds, we will focus on proving results for a slightly lower-variance version of the Impatient Thompson Sampling Bandit Algorithm from Algorithm \ref{alg:bandit}. 
We do this mainly to avoid repeating concentration of measure arguments from \cite{qin2023adaptive}, which are quite long and complicated.
We introduce this lower variance version in Algorithm~\ref{alg:banditrnd}.
As $m$ becomes large, the difference between Algorithm~\ref{alg:bandit} and Algorithm~\ref{alg:banditrnd} vanishes.
Later, in Section~\ref{sec:theory-vs-practice}, we show empirically that Algorithm~\ref{alg:bandit} behaves the same way whether $m$ is small or large.
As such, we believe that the theoretical insights we provide into the reduced-variance version carry over to the version presented in Algorithm~\ref{alg:bandit}.

Note that as stated in Algorithm \ref{alg:bandit}, the probability that any user $u \in \Uc_t$ is assigned action $a$ is 
\begin{align*}
    \p_{t,a} \triangleq \PP \bigg( a = \argmax_{\alpha \in \MC{A}} R( \tilde{\theta}_{\alpha}) \, \bigg| \, \HH_{t} \bigg),
\end{align*}
where the probability above averages over the draw of $\theta_a'$ from the posterior $N(\mu_{t,a}, \Sigma_{t,a})$.
Furthermore, lines \ref{line:startsample}--\ref{line:takeaction} in Algorithm \ref{alg:bandit} can be equivalently be written as for each $u \in \Uc_t$,
\begin{align*}
    A_u \gets \begin{cases}
        a & \TN{w.p.~~} \p_{t,a} \TN{~for~all~} a \in \MC{A}
    \end{cases}
\end{align*}
While typical Thompson sampling sets $A_u$ to action $a$ with probability $\p_{t,a}$, for each $u \in \Uc_t$, a lower variance of the algorithm can be defined which assigns exactly $\lvert \Uc_{t,a} \rvert \approx \lvert \Uc_t \rvert \cdot \p_{t,a}$ users to action $a$.
Since $\lvert \Uc_t \rvert \cdot \p_{t,a}$ is not necessarily a whole number, this would involve rounding $\p_{t,a}$ to some $\prnd_{t,a}$ which ensures that $\lvert \Uc_{t,a} \rvert = \lvert \Uc_t \rvert \cdot \prnd_{t,a}$ is a whole number.
This lower-variance version of the algorithm is presented in Algorithm \ref{alg:banditrnd}.

\begin{algorithm}
    \caption{Impatient Bandit Thompson Sampling---Lower Variance Version $\pirnd$}
    \begin{algorithmic}[1]
    \Require Priors and noise covariance matrices $\big\{ \mu_{1,Z_a}, \Sigma_{1,Z_a}, V_{Z_a} \big\}_{a \in \MC{A}}$
    \For{$a \in \MC{A}$}
        \State $(\mu_{1,a}, \Sigma_{1,a}) \gets (\mu_{1, Z_a}, \Sigma_{1, Z_a})$ \Comment{Initialize beliefs.}
    \EndFor
    \State $\MC{H}_1 \gets \{Z_a : a \in \MC{A}\}$ \Comment{Initialize history.}
    \For{$t = 1, 2, \dots, T$}
        \State $\left\{ \prnd_{t,a} \right\}_{a \in \MC{A}} \gets \TN{Round} \left( \left\{ \p_{t,a} \right\}_{a \in \MC{A}}, |\Uc_t| \right)$ \Comment{Apply rounding procedure.}
        \State $\{ \Uc_{t,a} \}_{a \in \MC{A}} \gets \mathrm{Partition} \left(\left\{ \prnd_{t,a}, \Uc_t \right\}_{a \in \MC{A}} \right) $ \Comment{Note that $\lvert \Uc_{t,a} \rvert = \lvert \Uc_{t} \rvert \cdot \prnd_{t,a}$.}
        \For{$a \in \MC{A}$}
            \For{$u \in \Uc_{t, a}$}
                \State $\tau_u \gets t$ \Comment{Set user decision time.}
                \State $A_u \gets a$ \Comment{Take action.}
            \EndFor
        \EndFor
        \State $\MC{H}_{t+1} \gets \MC{H}_t \cup \left\{ Y_u^{(j)}(A_u) : u \in \Uc_{\leq t}, \text{$j$ s.t. $t = \tau_u + d_j$} \right\}$ \Comment{Update history.}
        \For{$a \in \MC{A}$}
            \State $(\mu_{t+1,a}, \Sigma_{t+1,a}) \gets \textrm{PosteriorUpdate}(\mu_{t, Z_a}, \Sigma_{t, Z_a}, V_{Z_a}, \MC{H}_{t+1})$ \Comment{Update beliefs.}
        \EndFor
    \EndFor
    \end{algorithmic}
    \label{alg:banditrnd}
\end{algorithm}

In Algorithm \ref{alg:banditrnd}, $\mathrm{Round}( \left\{ \p_{t,a} \right\}_{a \in \MC{A}}, m )$ is a function that takes the probabilities to round and the total number of samples $m$ as input, and outputs probabilities $\left\{ \prnd_{t,a} \right\}_{a \in \MC{A}}$.
In order to ensure $m \cdot \prnd_{t,a}$ is a whole number, it rounds each $\prnd_{t,a}$ to a value in $\big\{ 0, \frac{1}{m}, \frac{2}{m}, \dots, \frac{m-1}{m}, 1 \big\}$. The rounding procedure also ensures that the rounding maintains a proper probability distribution, i.e., $1 = \sum_{a \in \MC{A}} \prnd_{t,a}$. 
The formal assumptions we make on the rounding procedure are written in Assumption \ref{assump:round}.

\begin{assumption}[Rounding Procedure]
  The rounding procedure is such that there exists a scalar $0.5 \geq \epsilon_{\rm rnd} > 0$ that uniformly bounds rounding error as:
  \[
  \left| \p_{t,a} - \prnd_{t,a} \right| \leq \epsilon_{\rm rnd} \quad \text{and} \quad \prnd_{t,a} \big/ \p_{t,a} \geq 1- \epsilon_{\rm rnd},
  \]
  with probability $1$ for all $t \in [1 \colon T], a \in \MC{A}$.
  \label{assump:round}
\end{assumption}

We provide an example rounding procedure that satisfies Assumption \ref{assump:round} in Appendix \ref{app:rounding}. In fact, this rounding procedure will ensure Assumption \ref{assump:round} holds for $\epsilonrnd = \frac{|\MC{A}|}{ m }$ where $m = \lvert \Uc_1 \rvert = \lvert \Uc_2 \rvert \dots = \lvert \Uc_T \rvert$ as long as $m > 2 \lvert \MC{A} \rvert$. This shows formally that rounding is not a major issue when user batches are large, as they are in the applications we have in mind.

\section{Theoretical Guarantee}
\label{sec:theory}

Nearly all bandit analyses focus on the rate at which regret---the suboptimality of selected arms---vanishes as the number of arm selections grows.
When reward observations are severely delayed, there is a long delay before anything is learned, regardless of how many arm selections are made in the interim;
In our motivating application, for example, the primary reward metric is computed after observing a user's $60$-day engagement with a recommended item.
Our theoretical focus, aligned with the practical objective, is not only on converging to the optimal arm in the limit $T \to \infty$, but also on leveraging progressive feedback to substantially improve decision quality relative to the case where no progressive feedback is available \emph{for every} $t \in \{1, \ldots, T\}$.
Two factors make this learning problem challenging:
\begin{description}
    \item[Delayed feedback.] After a user receives a recommendation, their downstream activity is revealed progressively across time. The resulting delay in information revelation limits how quickly the recommender system can learn and adapt. 
    \item[Bandit feedback.] The recommender system only observes a user's reward for the selected action $A_u$---not for all actions $\MC{A}$ (as is the case for full-feedback learning problems). To learn, the system needs to engage in costly exploration over the actions in $\MC{A}$. 
\end{description}
We provide a regret upper bound for our Impatient Bandit algorithm that aims to reflect the above two challenges in an interpretable manner. By taking advantage of high quality progressive feedback, it is possible for our bandit algorithm to resolve most of the uncertainty of the reward $R(Y_u)$ before one reaches the maximum delay $d_{\max} = \max_{j} d_j$. This progressive feedback can then be used to make higher quality decisions in this setting with bandit feedback. For our results, we prove a regret bound for the slightly lower-variance version of the Impatient Bandit Algorithm from Algorithm \ref{alg:banditrnd}.

\subsection{Warm-up: Setting without Action Features $Z_a$}
In this section we present a regret bound for the Impatient Bandit Algorithm (Algorithm \ref{alg:banditrnd}) in a setting in which action features $Z_a$ are not used to tailor the prior distribution, nor the noise covariance matrix $V_z$ (i.e., $\mu_z, \Sigma_z, V_z$ are constant across $z \in \MC{Z}$). See Section \ref{sec:generalizationWithZ} for the generalization to the setting with action features $Z_a$.

\subsubsection{Defining the Value of Progressive Feedback}
The regret of the Impatient Bandit Algorithm fundamentally depends on the \emph{quality} of the intermediate progressive feedback that the algorithm receives.
The intermediate feedback is of higher quality if it provides more information on the true mean reward.
To formalize this, below, we define a metric that captures the benefit of the progressive feedback in an information-theoretic sense.
We later use this metric in our regret bound. 

We define the ``Value of Progressive Feedback'' (VoPF) to equal the following conditional mutual information between the mean reward and the progressive (potential) outcomes,  conditional on the delayed (potential) outcomes for any action $a \in \MC{A}$ (recall $\MC{A}$ is sampled according to Assumption \ref{assum:items}):
\begin{align}
\textrm{VoPF}(t) = I \big(
    \bar{R}_a ;
    \underbrace{ \big\{ \tilde{Y}_{u,t}(a) : u \in \Uc_{< t} \big\} }_{ \TN{includes progressive feedback} } \mid
    \underbrace{ \big\{ \tilde{Y}_{u,t}(a) : u \in \Uc_{< (t-d_{\rm max})} \big\} }_{  \TN{only delayed outcomes}  }
\big).
\label{eqn:priceofcensoringNoZ}
\end{align}
Above, the outcomes $\big\{ \tilde{Y}_{u,t}(a) : u \in \Uc_{< (t - d_{\max})} \big\} = \big\{ Y_{u,t}(a) : u \in \Uc_{< (t - d_{\max})} \big\}$ refer to the \textit{entire} vector of full-feedback outcomes, which are not censored due to a deterministic delay and are not incrementally revealed. Also note that $\big\{ \tilde{Y}_{u,t}(a) : u \in \Uc_{< t} \big\}$ contains much more information about $\bar{R}_a$ than the history $\HH_{t}$. 
Here, $\big\{ \tilde{Y}_{u,t}(a) : u \in \Uc_{< t} \big\}$ represents the outcomes one would observe in a \textit{full-feedback} setting, where the outcomes of all arms are revealed after taking an action, rather than just the selected arm's outcomes are revealed (bandit-feedback).

Note that by design, the VoPF metric is entirely a property of the underlying environment, i.e., the potential outcomes distributions of the rewards and intermediate outcomes. This means that the VoPF metric \textit{does not} change depending on what decision-making algorithm is used to select actions. 

By well-known properties of mutual information, we can rewrite the VoPF as a difference of two conditional entropies:
\begin{align*}
    {\rm VoPF}(t)
    = H \big( \bar{R}_a \mid \underbrace{ \big\{ \tilde{Y}_{u,t}(a) : u \in \Uc_{< (t-d_{\rm max})} \big\} }_{  \TN{only delayed outcomes}  } \big)
    - H \big( \bar{R}_a \mid \underbrace{ \big\{ \tilde{Y}_{u,t}(a) : u \in \Uc_{< t} \big\} }_{ \TN{includes progressive feedback} }
    \big).
\end{align*}
By the above, we can interpret the expected value of the VoPF as the reduction in entropy in $\bar{R}_a$ by including progressive feedback.

\begin{remark}[Interpreting the Value of Progressive Feedback in Simple Examples]
\label{rem:interpretVoPF}
Consider the setting in which there are just two outcomes.
For any action $a \in \MC{A}$, selecting $A_u = a$, the first outcome $Y_u^{(1)}(a)$ is revealed immediately ($d_1 = 0$), the second (and final) outcome $Y_u^{(2)}(a)$ is revealed with delay $d_2 = d_{\max}$, and the reward is the final outcome, $R\big(Y_u(a)\big) = Y_u^{(2)}(a)$.
Assume that
\begin{align*}
\Sigma_1 = \begin{bmatrix}1 & \rho \\ \rho & 1\end{bmatrix} \qquad \TN{and} \qquad
V = \sigma^2_R \begin{bmatrix}1 & \rho \\ \rho & 1\end{bmatrix},
\end{align*}
for some $\rho \in [0, 1]$.
Then, $\TN{Corr}\big( Y_u^{(1)}(a), Y_u^{(2)}(a) \big) = \rho$ and one can show that
\begin{align*}
\TN{VoPF}(t) &= \frac{1}{2} \log \left( \frac{\Var(\bar{R}_a \mid \{ Y_u(a) : u \in \Uc_{<t-d_{\max}} \} )}{\Var(\bar{R}_a \mid \{ \tilde{Y}_{u,t}(a) \}_{u \in \Uc_{< t}}) } \right) \\
&= \frac{1}{2} \log\left( \frac{\sigma^2_R + m(t-d_{\max}) + 1}{\sigma^2_r + m(t-d_{\max}) + 1 - \rho^2} \right) \quad \forall t \geq d_{\max}.
\end{align*}
If the intermediate feedback is ``perfect'', i.e., $\rho = 1$, then 
\begin{align*}
\Var(\bar{R}_a \mid \{ \tilde{Y}_{u,t}(a) \}_{u \in \Uc_{< t} } )
    = \Var(\bar{R}_a \mid \{ Y_u(a) \}_{u \in \Uc_{<t} } ),
\end{align*}
and the Impatient Bandit algorithm will act as if the reward is observed immediately (no delay at all).
Alternatively, if the intermediate feedback is completely uninformative, i.e., $\rho = 0$, then $\TN{VoPF}(t) = 0$. In this case, the Impatient Bandit algorithm will act as if there is no intermediate feedback at all. See Appendix \ref{app:derivationRemark} for a derivation.
\end{remark}

\subsubsection{Regret Bound (Without Action Features $Z_a$)}

Corollary \ref{corr:mainNoZ} below follows from our main result. 
This corollary bounds the regret of the Impatient Bandit Thompson Sampling algorithm ($\pirnd$ from Algorithm \ref{alg:banditrnd}) in any batch in terms of (i) the reward variance $\sigma_{R}^2 \triangleq \Var \big( R(Y_u(a)) \mid \bar{R}_a \big)$,  (ii) the total number of previous user interactions over the number of actions $|\Uc_{< t}| / |\MC{A}|$, (iii) the Value of Progressive Feedback, and (iv) rounding error due to implementing the Thompson sampling allocation in a finite batch. 

Recall that $| \Uc_{<t}| = m (t-1)$ where $m=|\Uc_1|=|\Uc_2|= \dots = |\Uc_T|$ is the user batch size. Summing across $t$ gives a cumulative regret bound, which is more conventional. We prefer this, stronger, bound which indicates how low regret is in every batch. 

\begin{corollary}[Regret of Impatient Bandit Algorithm without Action Features $Z_a$]
\label{corr:mainNoZ}
Let Assumptions \ref{assum:exchangeable}-\ref{assump:round} hold, and let $\Sigma_{1}$ and $V$ be invertible. Then under the Impatient Bandit Thompson Sampling algorithm $\pirnd$ (Algorithm \ref{alg:banditrnd}), for any $t>1$,
\begin{align}
    \label{eqn:surrogateRegretNoZ}
    \E \left[ \Delta_t(\pirnd) \right]  \leq
    \underbrace{%
    \vphantom{ \sqrt{ \frac{ \lvert \MC{A} \rvert}{\sigma_{R}^2} } }
    \exp \big[ - {\rm VoPF}(t) \big]
}_{\substack{\TN{Decrease from} 
 \\ \TN{Progressive Feedback}}}
\cdot \underbrace{%
    \sigma_{R} \sqrt{ \frac{ 2 \lvert \MC{A} \rvert \log \lvert \MC{A} \rvert}{\sigma_{R}^2 (r_1^\top \Sigma_1 r_1)^{-1} + \lvert \Uc_{<(t-d_{\rm max})}\rvert} }
}_{\TN{Regret under Delayed Rewards}}
+ \underbrace{%
    \vphantom{ \sqrt{ \frac{ \lvert \MC{A} \rvert}{\sigma_{R}^2} } }
    O(\epsilon_{\rm rnd})
}_{\TN{Rounding error}}.
\end{align}
\end{corollary}
\noindent We discuss each term in this bound:
\begin{itemize}[leftmargin=5.5mm]
    \item {\bf Regret under Delayed Rewards:} Above, we can interpret the term \\
    $\sigma_{R} \cdot \sqrt{ \frac{ 2 |\MC{A}| \log(|\MC{A}|)}{\sigma_{R}^2 (r_1^\top \Sigma_1 r_1)^{-1} +| \Uc_{<(t-d_{\rm max})}|} }$ as the regret the Thompson Sampling algorithm would achieve with delayed rewards. In this setting, the system has observed rewards associated with recommendations that were made more than $d_{\max}$ periods ago, amounting to $| \Uc_{<(t-d_{\rm max})}|$ observed interactions in total.
The system has observed no feedback on more recent recommendations. This term can also be interpreted as a bound on the regret of an algorithm that 
\emph{ignores incremental feedback} even though it is available. 

The factor of the number of actions $|\MC{A}| $ in the bound can be thought of as the price of bandit feedback. It appears in our analysis because the system only observes a user's interaction \emph{with the single item recommended to them}. In a model in which we saw each user's interaction with all items in $\MC{A}$, this term would be replaced with $1 $ in our bound. The $\log(|\MC{A}|)$ term appears because it is used as a coarse upper bound for the entropy of the optimal action $A^*$ given the history. 
 
\item {\bf Decrease in Regret due to Progressive Feedback:} When using progressive feedback, the regret upper bound gets rescaled by a function of the of value of progressive feedback, $\exp \big[ - {\rm VoPF}(t) \big]$. Note in \eqref{eqn:surrogateRegretNoZ} above that when ${\rm VoPF}(t) = 0$, i.e., the intermediate outcomes are completely uninformative, the regret bound equals the standard regret bound for Thompson Sampling one would expect without any progressive feedback and just delayed rewards.

\item {\bf Rounding error:} Our analysis makes the $O(\epsilon_{\rm rnd})$ term explicit. We hide it in a big-O because we do not consider it to be an interesting feature of this analysis. In our application, the batch size $m = |\Uc_t|$ is large (hence $\epsilon_{\rm rnd}$ is not a big concern), but so is $\sigma_{R}$, so many batches are still required.
\end{itemize}

\paragraph{Per-User Regret Across Batches In Extreme Cases.}
In Proposition \ref{prop:extremes} below, we explicitly bound the average per-user regret across batches $\E \left[ \frac{1}{T} \sum_{t=1}^T \Delta_t(\pirnd) \right]$ under extreme values of $\TN{VoPF}(t)$. At one extreme, we consider the case in which the value of progressive feedback is zero (intermediate outcomes are completely uninformative). Regardless of the number of users per-batch $m$, regret can only be small after waiting to observe delayed rewards ($T>d_{\rm max}$); This takes 60 days in our main motivating application. At the other extreme, we consider the case in which there are “perfect'' intermediate outcomes (as if there is no delay). 
We can interpret Corollary \ref{corr:mainNoZ} as interpolating between these two extreme cases when $\TN{VoPF}(t)$ takes on different values.  
\begin{proposition}[Regret Under Extremes] \label{prop:extremes}
Let the conditions of Corollary \ref{corr:mainNoZ} hold.

\noindent \under{\TN{(1) Completely Uninformative Intermediate Feedback (Delayed Rewards).}} In the case that the intermediate feedback is completely uninformative, i.e., ${\rm VoPF}\big( t \big) = 0$ for all $t$,
    \begin{align}
        \E \bigg[ \frac{1}{T} \sum_{t=1}^T \Delta_t(\pirnd) \bigg] \, \leq \,
        \frac{d_{\max}}{T} \sqrt{ 2 (r_1^\top \Sigma_1 r_1) \log (|\MC{A}|) } 
        + 3 \sigma_R \sqrt{ \frac{ 2 |\MC{A}| \log (|\MC{A}|)}{Tm} } 
        + O(\epsilon_{\rm rnd}).
        \label{eqn:regretPt1}
    \end{align}

\noindent \under{\TN{(2) Perfect Intermediate Feedback (No Delay).}} If we have a perfect surrogate, i.e., \\
$\TN{Corr}\big(Y_u^{(1)}, R(Y_u) \big) = 1$ 
    and $d_1 = 0$, \vspace{-2mm}
    \begin{align}
        \E \bigg[ \frac{1}{T} \sum_{t=1}^T \Delta_t(\pirnd) \bigg]
        \, \leq \,  3 \sigma_R \sqrt{ \frac{ 2 |\MC{A}| \log (|\MC{A}|)}{ Tm} } + O(\epsilonrnd).
        \label{eqn:perfect}
    \end{align}
\end{proposition}
The regret bound in \eqref{eqn:perfect} matches (excluding constant factors) those for standard Thompson Sampling, without delayed rewards, from the literature \citep{russo2016information}. Note, the gap in regret between the two extreme settings above is worsened when one considers longer-term outcomes, i.e., as the delay $d_{\rm max}$ grows.

\paragraph{Novelty of our Regret Bound.} Our regret bound is novel because of how it is controlled by the Value of Progressive Feedback (VoPF). The bound directly quantifies the expected benefit of incorporating progressive feedback over just using delayed rewards. It is notable that the regret bound applies to settings in which rewards are completely delayed (intermediate outcomes have no useful information and the VoPF is zero), settings in which there is no delay (intermediate outcomes are perfect), as well as settings in between these two extremes. Moreover, in practice, the VoPF is a useful metric that can be estimated from historical data and could be used to compare the quality of different progressive feedback signals. In Section \ref{sec:spotify-prior}, we demonstrate how we estimate the VoPF on a Spotify podcast recommendation dataset.

To provide a regret bound in terms of the Value of Progressive Feedback metric, we had to overcome substantial conceptual and mathematical subtleties. The challenge lies in the nature of VoPF(t). This metric measures the value of early feedback on an arm that \emph{could be} available at decision-time $t$, but only \emph{if the algorithm had chosen to repeatedly recommend that arm in earlier periods}. When the VoPF(t) is low, progressive feedback would not be especially useful even if the algorithm opted to explore aggressively. But to upper bound the regret incurred at the $t^{\rm th}$ decision time, we need to reason about the volume of information the algorithm \emph{did}  acquire (i.e. its information gain)  rather than the volume of information it \emph{could have} acquired (the VoPF). The  mismatch between these notions is severe, since our Thompson sampling based algorithm will choose to select some arms very rarely. That is, it often gathers far less information about some arms than the VoPF measure suggests. At a high level, there is hope of overcoming this mismatch because Thompson sampling selects arms infrequently only when they are unlikely to be optimal and hence do not matter much to decision-making. 

To formalize this, we deviate from conventional analyses in the literature and provide intricate proofs that use inverse-propensity weights to analyze the evolution of posterior beliefs, building on \cite{qin2023adaptive}. It is unclear if there is any hope of adapting more conventional analyses, like the use confidence interval-based analyses \citep{russo2014learning,lattimore2020bandit} and those that utilize the \textit{information ratio}   \citep{russo2016information,lattimore2019information,lattimore2020bandit}, which were not designed to cope with delayed feedback.

\paragraph{Overview of Proof Approach.}
We now provide a brief overview of the key steps used in the proof of our regret bound from Corollary \ref{corr:mainNoZ}.
Let $\sigma_{t,a}^2 \triangleq {\rm Var}(\bar{R}_a \mid \HH_{t})$, where recall from \eqref{eqn:RbarDef} that $\bar{R}_a \triangleq \E \left[ R\big( Y_u(a) \big) \mid \theta_a \right]$.  By Lemma \ref{lem:regret-to-estimation}, which uses the probability matching property of Thompson Sampling and rounding Assumption \ref{assump:round}, we have that
\begin{align*}
    \E \left[ \Delta_t(\pirnd) \right] \leq \sqrt{ 2 \log(|\MC{A}|) \E_{\pirnd} \bigg[ \sum_{a\in \MC{A}} p_{t,a}^* \sigma^2_{t, a} \bigg] } 
    + O(\epsilonrnd)
\end{align*}
where $p_{t,a}^* = \PP( A^* = a \mid \HH_{t} )$. Thus, the instantaneous regret is controlled by a function of the posterior uncertainty of the optimal arm under the Thompson sampling algorithm. 

The crux of the proof is showing the following inequality holds:
\begin{align*}
    \E_{\pirnd} \bigg[ \sum_{a\in \MC{A}} p_{t,a}^* \sigma^2_{t, a} \bigg] 
    \leq (1+2\epsilon_{\rm rnd}) \sum_{a \in \MC{A}} \E \left[ \Var \left( \bar{R}_a \mid Z_a, \{ \tilde{Y}_{u,t}(a) \}_{u \in \Uc_{< t} } \right) \right]
\end{align*}
Crucially above, the left hand side is an expectation related to the actions selected under the rounded Impatient Bandit algorithm, but the right hand side is a function of the potential outcomes and not the algorithm itself. The proof of this inequality uses inverse-propensity weighting together with an inequality from \cite{qin2023adaptive}.

Finally, the corollary holds by Lemma \ref{lemma:simplifyingCov}, which shows the following equality
\begin{align*}
    \E \left[ \Var \left( \bar{R}_a \mid Z_a, \{ \tilde{Y}_{u,t}(a) \}_{u \in \Uc_{< t} } \right) \right]
    = \E \left[ \frac{ \sigma_{R}^{2}(Z_a) \cdot \exp \big( -2 \cdot {\rm VoPF}(t, Z_a) \big) }{ \sigma_{R}^2(Z_a) \cdot (r_1^\top \Sigma_{Z_a} r_1)^{-1} + |\Uc_{< (t-d_{\rm max})}| } \right].
\end{align*}

\subsection{General Setting with Action Features $Z_a$}
\label{sec:generalizationWithZ}

We now characterize the regret of the Impatient Bandit Thompson Sampling algorithm that uses action features $Z_a$. First, we define a generalized version of the Value of Progressive Feedback that depends on $Z_a$:
\begin{align}
    {\rm VoPF}(t, z) \triangleq I \bigg( \bar{R}_a ; \underbrace{ \big\{ \tilde{Y}_{u,t}(a) : u \in \Uc_{< t} \big\} }_{ \TN{includes progressive feedback} } \mid \underbrace{ \big\{ \tilde{Y}_{u,t}(a) : u \in \Uc_{< (t-d_{\rm max})} \big\} }_{  \TN{only delayed outcomes}  }, Z_a = z \bigg).
    \label{eqn:priceofcensoring}
\end{align}

Using this generalized version of the Value of Progressive Feedback, we can now present our generalized regret bound. This result bounds the regret of Impatient Bandit Thompson sampling in terms of (i) the variance of rewards conditional on action features $\sigma^2_{R}(z) = \Var \big( R(Y_u(a)) \mid \bar{R}_a, Z_a = z \big)$, (ii) the number of previous recommendations $|\Uc_{<t}|$, (iii) the number of actions/items $|\MC{A}|$, (iv) the generalized definition of the value of progressive feedback (VoPF), and (v) a rounding error term.
\begin{Theorem}[Regret with Progressive Feedback]
\label{thm:main}
Let Assumptions \ref{assum:exchangeable}-\ref{assump:round} hold, and let $\Sigma_{1,z}$ and $V_z$ be invertible for all $z \in \MC{Z}$. Then under Impatient Bandit Thompson Sampling algorithm $\pirnd$ (Algorithm \ref{alg:banditrnd}), for any $t>1$,
\begin{align*}
    \E \left[ \Delta_t(\pirnd) \right]  
    &\leq \sqrt{ \E \left[ \frac{ \exp \big( -2 \cdot {\rm VoPF}(t, Z_a) \big) \cdot \sigma^2_{R}(Z_a) }{ \sigma_{R}^2(Z_a) \cdot (r_1^\top \Sigma_{Z_a} r_1)^{-1} + |\Uc_{< (t-d_{\rm max})}|} \right] } \times \sqrt{ 2 |\MC{A}| \log(|\MC{A}|) } + O(\epsilon_{\rm rnd}). 
\end{align*}
\end{Theorem}
Note that the expectation $\E \left[ \frac{ \exp \left( -2 \cdot {\rm VoPF}(t, Z_a) \right) \cdot \sigma^2_{R}(Z_a) }{ \sigma_{R}^2(Z_a) \cdot (r_1^\top \Sigma_{Z_a} r_1)^{-1} + |\Uc_{< (t-d_{\rm max})}|} \right]$ above averages over the draw of action features $Z_a$.

\section{Fitting the Prior}
\label{sec:prior}

We now discuss how to use historical data (training set), collected prior to the deployment of the Impatient Bandit algorithm, to fit the prior distribution parameters $\big\{ \mu_{1, z}, \Sigma_{1,z} \big\}_{z \in \MC{Z}}$ and noise covariance matrices $\{ V_z \}_{z \in \MC{Z}}$. We use $\Uc^{\TN{hist}}$ to denote the historical users (training set) and $\MC{A}^{\TN{hist}}$ to denote the historical items which were recommended to users $\Uc^{\TN{hist}}$. We assume items $\MC{A}^{\TN{hist}}$ are drawn according to the process from Assumption \ref{assum:items}. Throughout, we assume that there are a finite number of action features, i.e., $| \MC{Z} | < \infty$. In the podcast recommendation case, $Z_a$ could be the category of the podcast (e.g., comedy, news).

\paragraph{Fitting the Noise Covariance Matrices.}
Recall that $Y_u(a) \mid (\theta_a, Z_a = z) \sim N( \theta_a, V_z)$. We estimate the $V_z$ by averaging the noise covariance matrices $\hat{V}^{(a)}$ estimates for each action $a \in \MC{A}^{\TN{hist}}$:
\begin{align*}
    \hat{V}_z \triangleq \frac{ \sum_{a \in \MC{A}^{\TN{hist}}} \ind_{Z_a = z} \hat{V}^{(a)} }{\sum_{a \in \MC{A}^{\TN{hist}}} \ind_{Z_a = z}} \quad \TN{where} \quad \hat{V}^{(a)} \triangleq \frac{1}{ N^{(a)} } \sum_{u \in \Uc^{\TN{hist}}} \ind_{A_u = a} \big( Y_u - \bar{Y}^{(a)} \big) \big( Y_u - \bar{Y}^{(a)} \big)^\top.
\end{align*}
Above, we use, $N^{(a)} \triangleq \sum_{u \in \Uc^{\TN{hist}}} \ind_{A_u = a}$ and $\bar{Y}^{(a)} \triangleq \frac{1}{ N^{(a)} } \sum_{u \in \Uc^{\TN{hist}}} \ind_{A_u = a} Y_u$.

\paragraph{Fitting the Prior Distribution Parameters.}

In the statistics and machine learning literature, a common approach to fitting an informative prior distribution using data is type II maximum likelihood \citep[page 173]{pml1Book}---also called empirical Bayes \citep{casella1985introduction}.  For the Impatient Bandit algorithm, the prior hyperparameters we want to fit using historical data are $\big\{ \mu_{1, z}, \Sigma_{1,z} \big\}_{z \in \MC{Z}}$, i.e., the mean vectors and covariance matrices for the multivariate Gaussian prior for $\theta_a \mid Z_a$. For now, we assume that we already have estimators of the noise covariance matrices $\{ V_z \}_{z \in \MC{Z}}$.

The basic idea behind type II maximum likelihood is to choose the prior hyperparameters $\{ \mu_{1, z}, \Sigma_{1,z} \}$ that maximize the \textit{marginal} likelihood of the data.
In our Gaussian prior case, the maximum marginal likelihood criterion can be written as the following minimization problem: \vspace{-2mm}
\begin{multline*}
    (\hat{\mu}_{1,z}, \hat{\Sigma}_{1,z} ) \triangleq \argmin_{\mu, \Sigma} \\
    \sum_{a \in \MC{A}^{\TN{hist}}} \ind_{Z_a = z} \bigg\{ \log \left( \Sigma + V_z / N^{(a)} \right) + \underbrace{ 
    (\bar{Y}^{(a)} - \mu)^\top \left[ \Sigma + V_z / N^{(a)} \right]^{-1} (\bar{Y}^{(a)} - \mu) }_{\TN{Weighted Least Squares}} 
    \bigg\}
\end{multline*}
Note that by the above criterion, it is clear that if $\Sigma_{1,z}$ was known, then $\hat{\mu}_{1,z}$ would be the solution to a weighted least squares criterion. 
In general, when $\Sigma_{1,z}$ is unknown, one can minimize the above criterion 
using Newton-Raphson or Expectation maximization \citep{normand1999meta,SASpackage}.

In the special case that $N^{(a)} = n$ for all $a \in \MC{A}^{\TN{hist}}$ such that $Z_a = z$, then the solution to the above $\hat{\mu}_{1,z}$ is a simple mean over $\bar{Y}^{(a)}$'s (no matter the value of $\Sigma_{1,z}$):
\begin{align}
    \label{eqn:muhat}
    \hat{\mu}_{1,z} = \frac{1}{\sum_{a \in \MC{A}^{\TN{hist}}} \ind_{Z_a = z}} \sum_{a \in \MC{A}^{\TN{hist}}} \ind_{Z_a = z} \bar{Y}^{(a)}.
\end{align}
For our experiments, we approximately solve  by estimating $\mu_{1,z}$ using \eqref{eqn:muhat} (even though $N^{(a)}$ varies across actions). We estimate $\hat{\Sigma}_{1,z}$ using $\hat{\Sigma}_{1,z} \triangleq \frac{1}{\sum_{a \in \MC{A}^{\TN{hist}}} \ind_{Z_a = z}} \sum_{a \in \MC{A}^{\TN{hist}}} \ind_{Z_a = z} \big( \bar{Y}^{(a)} - \hat{\mu}_{1,z} \big) \big( \bar{Y}^{(a)} - \hat{\mu}_{1,z} \big)^\top$.

\begin{figure}[b!]
    \centering
    \includegraphics{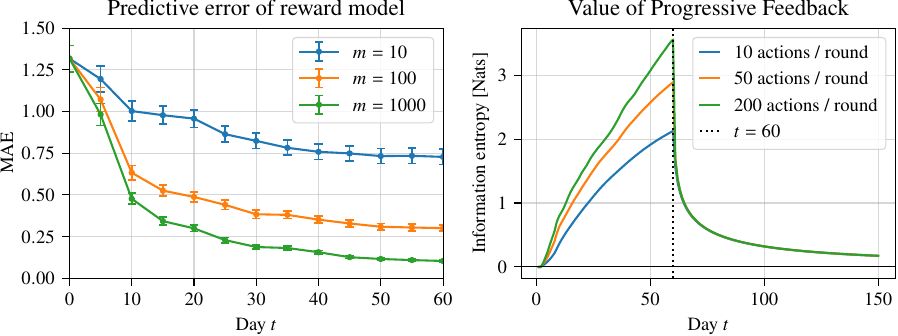}
    \vspace{-5mm}
    \caption{Prior learned on Spotify interaction data.
Left: predictive error of the reward model as a function of the number of traces $m \in \{10, 100, 1000\}$ and the number of observed days $t$.
The mean and standard errors are computed over 200 actions in the test set.
Right: estimated Value of Progressive Feedback for $m \in \{10, 50, 200\}$.
    }
    \label{fig:spotify-prior}
\end{figure}

\subsection{Example on Real-World User Interaction Data}
\label{sec:spotify-prior}
To illustrate prior fitting in practice, we consider a dataset of podcast consumption traces collected on the Spotify audio streaming platform between September 2021 and May 2022.
The data is divided into a training set and an independent validation set;
The training and validation sets cover podcast shows appearing during the periods September–December 2021 and January–March 2022, respectively.
Each subset consists of a sample of $200$ podcast shows first published on the platform during a given three-month period.
For each of these shows, the data contains a representative sample of users that discover the show during the same three-month period.
For each user, we obtain a longitudinal trace that captures their engagement with the show on each day starting from the discovery, i.e.,
\begin{align*}
Y_u^{(j)} = \mathbbm{1}\{\text{user $u$ engaged with the show on the $j$th day since discovery}\},
\end{align*}
for $j = 1, \ldots, 60$.
By construction, the user engaged with (discovered) the recommended content on the first day, i.e., $Y_u^{(1)} = 1$ for all $u$. 
The reward is defined as the cumulative engagement over the 60 days from the initial discovery, i.e.,
\begin{align*}
R(Y_u) = \sum_{j = 1}^{60} Y_u^{(j)}
\end{align*}
In total, the dataset consists of 8.77M activity traces, corresponding to a total of 26M cumulative active-days.
The number of traces per show ranges between 2.4K and 295K, with a median of 5.8K. 

We examine the prior fitted using the procedure described above using the data from training set.
First, in Figure \ref{fig:spotify-prior} (left), we examine the predictive error of the model on the validation set as we vary the number of days and the batch size.
Specifically, we compute the mean absolute error of the observed rewards and posterior mean of our model, which is initialized with the fitted prior.
We see that the model's predictions can be relatively accurate after observing only 10 days of data.
The predictions improve as time passes, and increasing the batch size further increases predictive accuracy.

In Figure \ref{fig:spotify-prior} (right), we examine the Value of Progressive Feedback according to the fitted prior.
We plot the Value of Progressive Feedback as we vary the batch size $m$.
Notice that as $m$ increases, the value of surrogate feedback increases, as the larger batch sizes reduce noise.

\section{Empirical Evaluation}
\label{sec:experiments}

Next, we investigate different aspects of the performance of our bandit algorithm empirically with two decision-making problems.
After briefly introducing these two problems in Section~\ref{sec:environments}, we address five questions.
\begin{enumerate}
\item How does the empirical improvement from progressive feedback compare to our theory's predictions? (Section \ref{sec:theory-vs-practice})
\item Are there real-world applications that can benefit substantially from progressive feedback? (Section \ref{sec:spotify-benefit})
\item How important is fitting the prior, as described in Section~\ref{sec:prior}? (Section \ref{sec:fittingPriorImportance})
\item How do bandits with and without progressive feedback fare in a realistic setting with perpetually changing actions? (Section \ref{sec:changingActions})
\item How does Thompson sampling compare to sequential elimination in problems with progressive feedback? (Section \ref{sec:successive-elimination})
\end{enumerate}

To this end, we quantify empirically the performance of our impatient bandit algorithm on the two problems.
A natural benchmark for a bandit algorithm that makes use of progressive feedback is to compare it to a copy of the same algorithm, but one where we pretend that all outcomes are revealed at once after delay $d_{\max}$.
Equivalently, we can think of that benchmark as ignoring the outcomes $Y_u^{(1)}, Y_u^{(2)}, \ldots$ as they become available, and only using the final delayed reward $R(Y_u)$.
This benchmark is implicit in Theorem~\ref{thm:main}, and will appear throughout our experiments.

\subsection{Data-Generating Environments}
\label{sec:environments}

Across our experiments, we study two decision-making problems.
The first is an idealized, synthetic problem, where we can explicitly vary the Value of Progressive Feedback.
The second one is inspired by a real-world product problem at Spotify (a problem we will later re-examine in Section~\ref{sec:abtest}).
In this second problem, actions, rewards and progressive feedback represent actual interactions of users with content on Spotify.

\subsubsection{Synthetic Environment}
We postulate a simple, synthetic generative model.
There are $J = 25$ outcomes per decision denoted $Y_u \in \real^J$, and each outcome $j \in [J]$ is revealed with delay $d_j = j-1$ (first outcome is revealed immediately, second outcome is revealed following decision time $t+1$, and so on).
Given a hyperparameter $\alpha \in [0, 1)$, for each action $a \in \mathcal{A}$, we sample a mean outcome vector
\begin{align}
    \label{eqn:synthetic-prior}
    \theta_a \sim N(0, \Sigma_1), \quad \TN{where} \quad  \Sigma_1 = (1 - \alpha^2)\boldsymbol{I} + \alpha^2 \bm{1}\bm{1}^\top,
\end{align}
where $\bm{I}$ is the $J \times J$ identity matrix, and $\bm{1}$ is the all-ones vector.
Then, conditional on $\theta_a$ for each action $a$ and each user $u$, we independently sample potential outcomes
\begin{align*}
Y_u(a) \mid \theta_a \sim N(\theta_a, V), \quad \TN{where} \quad V = m \boldsymbol{I}.
\end{align*}
Note that the dependency of the noise covariance matrix $V$ on the batch size $m$ is chosen such that, informally, the amount of information collected at each time step is independent of $m$.
Furthermore, $\alpha$ controls how much information the first entries of $\theta_a$ contain about future entries.
The higher $\alpha$ is, the more informative the progressive feedback is.
In our experiments, we vary $\alpha$ to vary how informative the intermediate leading indicators are.
Specifically, we vary $\alpha \in \{0.1, 0.4, 0.8\}$.
We refer to these as the \emph{Low}, \emph{Mid}, and \emph{High} information regimes, respectively.

We consider a reward that is the average outcome: $R(Y_u) = c^{-1} \sum_{j=1}^J Y_u^{(j)}$, revealed after delay $J-1$.
We set $c = \sqrt{(1 - \alpha^2) J + \alpha^2 J^2}$ such that the mean reward $\bar{R}_a = \E[R(Y_u(a)) \mid \theta_a ] = c^{-1} \sum_{j=1}^J \theta_a^{(j)}$ has variance $\mathrm{Var}[\bar{R}_a] = 1$ for any choice of parameters.

\begin{figure}[b!]
    \centering
    \includegraphics{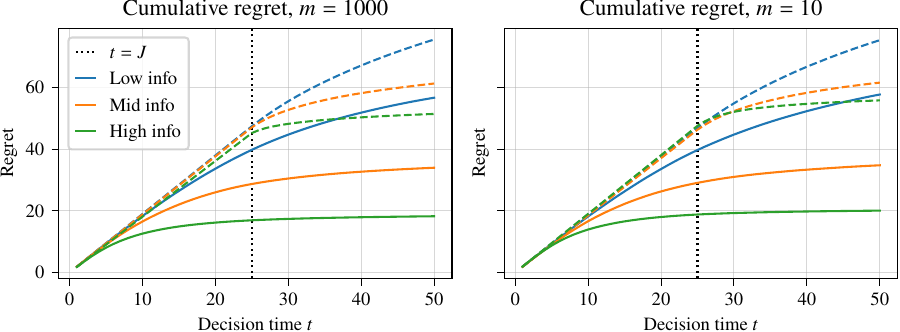}
    \vspace{-5mm}
    \caption{Cumulative regret for three different settings of the generative model, averaged over \num{500} runs.
    Solid lines correspond to the Impatient Bandit algorithm, which uses progressive feedback, and dashed lines correspond to a Thompson Sampling algorithm that ignores progressive feedback and waits for delayed rewards. All standard errors are $\le \num{0.88}$.}
    \label{fig:genmodel-cumul-regret}
\end{figure}

\subsubsection{Spotify Environment}
We reuse the data presented in Section~\ref{sec:spotify-prior}, consisting of discoveries of podcast shows on Spotify, as well as the user's subsequent engagement with the show over the 60 days that follow.
In this semisynthetic environment, actions correspond to podcast shows.
Every time we take an action, we sample a datum $Y_u$ from the user engagement traces of the corresponding show uniformly at random with replacement.
The reward is defined as the cumulative engagement: $R(Y_u) = \sum_{j=1}^J Y_u^{(j)}$.

\begin{figure}[b!]
    \centering
    \includegraphics{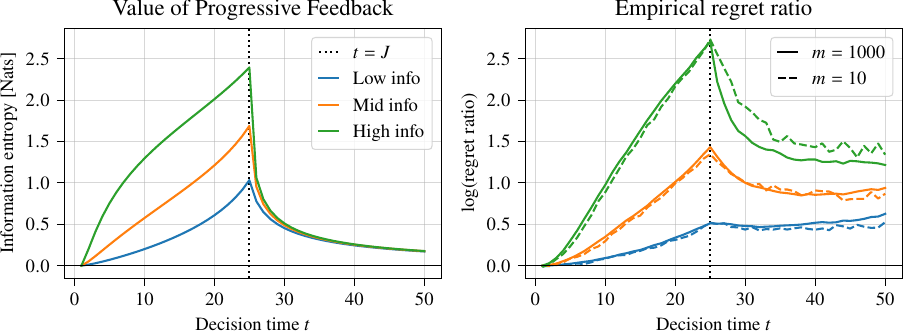}
    \vspace{-5mm}
    \caption{A comparison between the Value of Progressive Feedback, a theoretical quantity (\emph{left)} and the empirical benefit of our algorithm over the delayed bandit algorithm, measured by averaging over 500 simulations (\emph{right}, experimental details provided in the text).
The empirical benefit qualitatively matches the theoretical prediction.}
    \label{fig:genmodel-ratio}
\end{figure}

\subsection{How does the empirical improvement from progressive feedback compare to our theory's predictions?}
\label{sec:theory-vs-practice}
We run simulations on the synthetic environment with $\lvert \mathcal{A} \rvert = 20$ actions.
In Figure \ref{fig:genmodel-cumul-regret} we plot the cumulative regret of the Impatient Bandit algorithm in the low, medium, and high information regimes for $m \in \{10, 1000\}$.
We contrast the performance of our Impatient Bandit algorithm, Algorithm~\ref{alg:bandit} (solid lines), with that of a Thompson Sampling algorithm that only uses full rewards observed after delay $d_{\textrm{max}} = J-1$ (dashed lines).
Note that both the Impatient Bandit and delayed Thompson Sampling algorithms use a correctly specified prior that is derived using \eqref{eqn:synthetic-prior}.

We observe that the benefit of using intermediate outcomes increases with $\alpha$.
Furthermore, the average performance of the algorithms does not change substantially as we reduce the number of actions per round dramatically from $m = 1000$ to $m = 10$, suggesting that the sampling step does not impact the performance in expectation.

In Figure \ref{fig:genmodel-ratio}, on the left we display the Value of Progressive Feedback \eqref{eqn:priceofcensoring} for the low, mid, and high information regimes.
Recall that the VoPF is a property of the underlying data generating environment, and can be computed in closed form given our generative model.
On the right-hand side of Figure \ref{fig:genmodel-ratio} we plot the empirical ratio (on log scale) between the regret of Thompson Sampling using only delayed rewards and the regret of the Impatient Bandit algorithm, which uses progressive feedback.
Based on Corollary \ref{corr:mainNoZ}, we expect that the logarithm of this regret ratio is approximately upper bounded by the Value of Progressive Feedback.
Indeed, we observe that the (log) empirical regret ratio on the right closely resemble the $\TN{VoPF}(t)$ curves on the left.
Again, we observe no significant difference between the large-batch ($m = 1000$) and small-batch ($m = 10$) regimes.

Notice that the VoPF sharply drops at decision time $t = J + 1 = 26$.
This occurs because, when $t = J+1$, the set of delayed outcomes $\big\{ Y_{u}(a) : u \in \Uc_{< (t-J)} \big\}$ begins to include the user potential outcomes from the first batch.

\begin{figure}[b!]
    \centering
    \vspace{-5mm}
    \includegraphics{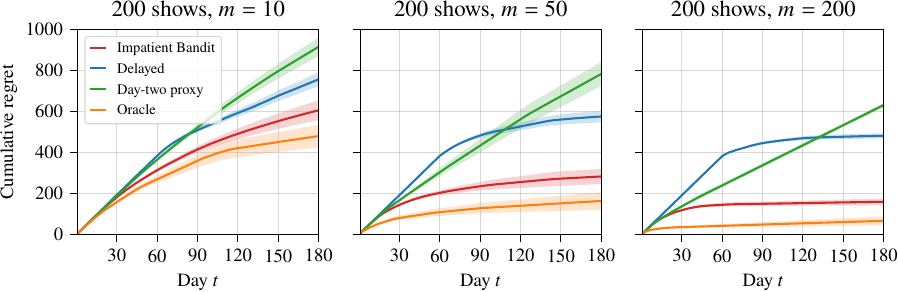}
    \vspace{-5mm}
    \caption{Cumulative regret of different bandit algorithms on the Spotify problem with $|\MC{A}| = 200$ shows and batch sizes $m \in \{10, 50, 200\}$.
The mean (solid line) and standard error (shaded area) are computed over 10 runs.}
    \label{fig:spotify-bandit-200}
\end{figure}

\subsection{Are there real-world applications that can benefit substantially from progressive feedback?}
\label{sec:spotify-benefit}

Next, we focus on the Spotify environment, where intermediate outcomes and rewards are collected from real-world interactions, and reflect a concrete product problem. This setting was first described in Section \ref{sec:spotify-prior}.
We ask: Does the value of progressive feedback show a large benefit on real-world data?

We benchmark our Impatient Bandit algorithm (which we label \emph{Progressive} in the following) against several baselines, all of which are based on variants of Thompson sampling.
In addition to the \emph{Delayed} algorithm we introduced previously, we consider two additional algorithms.
\begin{description}
\item[Day-two proxy.]
Instead of the true delayed reward, this algorithm learns to maximize a short-term proxy.
Specifically, the algorithm uses $Y_u^{(2)}$ as a proxy for the reward, and discards all subsequent information $(Y_u^{(j)})_{j \geq 3}$.
This baseline captures an intuitive outcome that is clearly related to the goal of maximizing habitual engagement: Does the user return to the show the day after first discovering it?
This baseline is representative of short-term proxies widely used in recommender systems, such as the click-through-rate, the dwell time, or the conversion rate \citep{lalmas2014measuring, dedieu2018hierarchical, bogina2017incorporating}.

\item[Oracle.]
This algorithm assumes that all outcomes are observed immediately following an action, i.e., that $d_j = 0$ for all $j$.
This is clearly unrealistic, but it provides a natural upper-bound on attainable performance.
\end{description}
Note, all the algorithms use the same prior learned via the procedure described in Section \ref{sec:prior}.

We consider a setting with $|\MC{A}| = 200$ and we vary the size of the batches ($m$).
We run 10 simulations for each algorithm, and we plot the resulting (average) cumulative regret in Figure \ref{fig:spotify-bandit-200}.
These regret plots suggest that using the day-two outcome is a poor proxy for the true reward, since the Day-two proxy algorithm appears to be the only one that does not have sublinear regret.
The performance of the Delayed algorithm is comparable to that of the Day-two proxy until the day $60$, at which the algorithm begins to observe the delayed rewards.
The Progressive algorithm accumulates much less regret during the first $60$ decision times compared to the Delayed algorithm.
Strikingly, the Progressive algorithm's regret appears to be only slightly worse than that of the oracle.

\begin{figure}[b!]
    \centering
    \includegraphics{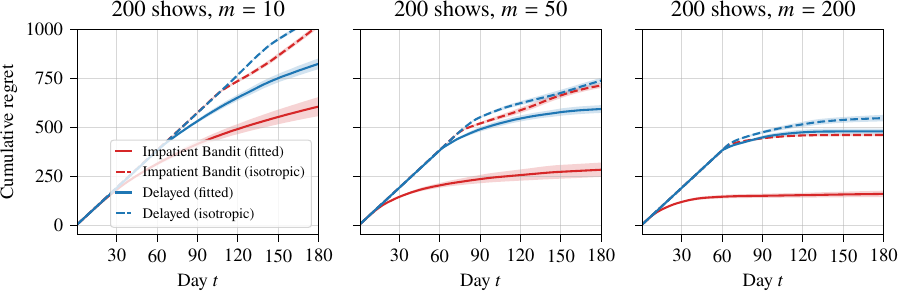}
    \vspace{-5mm}
    \caption{Cumulative regret of two variants of our bandit algorithm using different prior and covariance matrices, and a delayed bandit algorithm.
The mean and standard error are computed over 10 runs.
Accurately estimating hyperparameters is instrumental to achieving good performance.}
    \label{fig:spotify-bandit-priors}
\end{figure}

\subsection{How important is fitting the prior, as described in Section~\ref{sec:prior}?}
\label{sec:fittingPriorImportance}

We now examine the importance of using a learned prior, in the context of the Spotify decision-making problem.
To this end, we include an additional baseline, which we call \emph{Progressive (isotropic)}.
Specifically, this variant uses the Impatient Bandit algorithm, but instead of using the fitted prior it uses the following uninformative prior:
\begin{itemize}
    \item the prior mean is the all-zeros vector: $\mu_1 = \bs{0}$,
    \item the prior covariance is isotropic (multiple of identity matrix): $\Sigma_1 = 100 \cdot I$, and
    \item the noise covariance matrix is the identity matrix: $V = I$.
\end{itemize}
In Figure \ref{fig:spotify-bandit-priors} we see that the cumulative regret of the progressive algorithm with the uninformative prior is comparable to that of the Delayed algorithm.
Both the Progressive (uninformative prior) and Delayed algorithms are significantly outperformed by the Progressive algorithm with the fitted prior.
These results suggest that the prior is really \emph{driving} the good performance of the Impatient Bandit algorithm.

\subsection{How do progressive feedback bandits fare in a realistic setting with perpetually changing actions?}
\label{sec:changingActions}

We consider a non-stationary variant of the Spotify environment, where the set of shows which the algorithm recommends from gradually changes at each decision time.
This is designed to reflect how in the real podcast recommendation problem, new podcast shows are continually being released over time.
Specifically, in our experiment, the number of available actions is equal to $60$ throughout. 
After each decision-time, we drop the oldest action, and add a new action.

\begin{figure}[h!]
    \centering
    \includegraphics{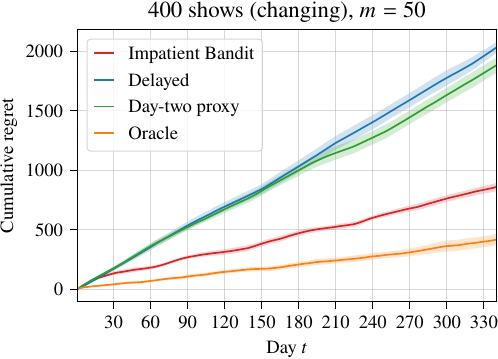}
    \caption{Cumulative regret of different bandit algorithms on a variant of the Spotify problem, where the action set changes slightly at each decision time.
The mean (solid line) and standard error (shaded area) are computed over 10 runs.}
    \label{fig:spotify-bandit-nonst}
\end{figure}

In Figure \ref{fig:spotify-bandit-nonst} we plot the cumulative regret of our algorithm over time up until $t=340$ days. Note in this setting there are 400 shows in the action space in total over time; this means that each action is selected less than once on average.
In this setting, it is particularly striking that the regret of the Delayed algorithm matches or is slightly worse than that of the Day-two proxy algorithm.
Since the set of shows to recommend is changing, the Delayed algorithm's regret does not begin to improve at day $60$, as we saw earlier in Figure \ref{fig:spotify-bandit-200}.
These results suggest that in settings where the cold start problem is at the forefront (e.g., a recommendation problem where the set of items over which to recommend changes frequently~\citep{baran2023accelerating}), the Delayed algorithm can perform very poorly.
In contrast, the algorithm that incorporates progressive feedback incurs less than half of the cumulative regret of the Delayed and Day-two proxy algorithms.

\vspace{-8mm}
\subsection{How does Thompson sampling compare to sequential elimination?}
\label{sec:successive-elimination}

In this section, we compare our algorithm to a Sequential Elimination based approach \citep{even2006action,shahrampour2017sequential,karnin2013almost}.
Sequential elimination is a natural algorithm to compare to since it is conceptually very similar to Thompson Sampling, and it can also take advantage of the Bayesian filter we introduce.
The sequential elimination algorithm we use maintains a posterior distribution over the mean reward using our progressive feedback Bayesian model.
Under Sequential Elimination, when an action's probability of being optimal according to the posterior distribution falls below a certain threshold, that action is eliminated; Actions are selected uniformly over the set of non-eliminated actions.
In contrast, Thompson Sampling gradually reduces the probability of selecting arms that are unlikely to be optimal, which makes it often more difficult to analyze.

\begin{figure}[h!]
    \centering
    \includegraphics{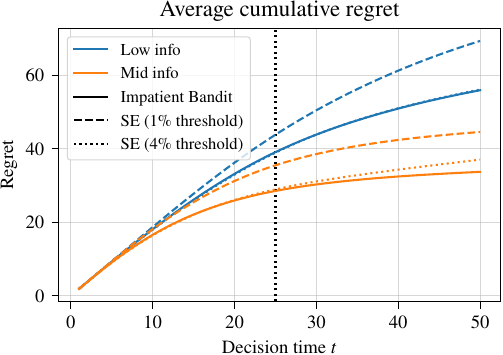}
    \caption{Cumulative regret of our Thompson Sampling Impatient Bandit algorithm (solid lines) and two variants of Sequential Elimination with different elimination thresholds (dashed and dotted lines), averaged over \num{5000} runs.
All standard errors are $\le \num{0.28}$.  The dashed vertical line indicates the value of $d_{\max}$, the delay before any reward is fully observed. }
    \label{fig:genmodel-cumul-regret-SE}
\end{figure}

For the purposes of this comparison, we use the synthetic environment and focus on the low and mid-information regimes.
We run simulations with two variants of Sequential Elimination, with threshold values of $1\%$ and $4\%$, respectively.
We plot the cumulative regret of both successive elimination algorithms and our Impatient Bandit Thompson Sampling based algorithm in Figure \ref{fig:genmodel-cumul-regret-SE}.
Similar to our Impatient Bandit algorithm, the regret of the successive elimination algorithms improve in the mid-information environment, as compared to the low information environment.
We find that the Impatient Bandit Thompson Sampling based algorithm matches the performance of the best successive elimination algorithm we consider in all settings, and has even better performance in the mid-information regime.
Overall, while successive elimination may have some advantages, it does not  quite get the same performance as Thompson Sampling.

\vspace{-5mm}
\section{A/B Test}
\label{sec:abtest}

Our work addresses a key challenge encountered at Spotify. While \cite{maystre2023optimizing} demonstrated substantial improvements from optimizing recommender systems for long-term outcomes, their methods introduce significant delays in evaluating new content, as they require waiting to observe whether users form lasting listening habits with the content. This section describes how progressive feedback, combined with the Gaussian filtering procedure introduced in Sections \ref{sec:problemFormulation} and \ref{sec:algorithm}, alleviates this challenge. It also presents evidence of significant impact from an A/B test involving hundreds of millions of users over a two-month period. Based on this and subsequent tests, the method was rolled out in production to power personalized podcast recommendations for hundreds of millions of listeners worldwide.

While our model of progressive feedback in multi-armed bandits aims to distill a core challenge that arises in recommender systems, rather than attempt to realistically model all intricacies of a particular system, the test results demonstrate that incorporating progressive feedback via Gaussian filtering can have enormous impact even in a complex industrial-scale system. The primary differences between the algorithm introduced in previous sections and the A/B test implementation are that \begin{enuminline}
\item the Gaussian filtering model is used to rank items on a shelf rather than make single recommendations, and
\item the algorithm tested does not perform purposeful exploration, which was not a priority for our test partners. 
\end{enuminline}
These differences are discussed in Section \ref{sec:variants}. Despite these differences, the substantial improvements observed in the A/B test validate the fundamental insight that progressive feedback can be leveraged to accelerate learning in recommendation systems.

\subsection{Recommendation algorithms for long-term rewards}
\label{subsec:intro_practical}

While recommender systems often have recurring interactions with individual users across months or years, the machine learning systems that power personalized recommendations are often optimized for short-term measures of success. For instance, it is common in the industry to train machine learning algorithms that predict the chance a user will (briefly) engage with content if it is shown and then to display whichever content has the greatest predicted chance of engagement.

\cite{maystre2023optimizing} detail efforts to optimize key components of Spotify's recommender systems for personal listening journeys that unfold over months. Beginning with a fairly general reinforcement learning (RL) model, they derive a pragmatic solution by imposing structural assumptions on certain value functions (or $Q$-functions). They describe the motivation of their modeling as follows:
\begin{quote}
   \emph{Our approach to modeling the $Q$-function is driven by a key hypothesis recommendations significantly contribute to long-term user satisfaction by fostering the formation of specific, recurring engagement patterns with individual pieces of content, which we call “item-level listening habits”. In the context of audio streaming, these habits can manifest in various forms: a user might develop a regular routine with a specific podcast show, returning to it as new episodes are released; form an attachment to a particular creator’s content; or integrate a curated playlist into their daily activities. Here, we use the term ``item'' to refer to any distinct piece of content that can be recommended, such as a podcast show, an album, or a playlist.}
\end{quote}
Their approach ultimately breaks down a complex RL problem into the problem of building two models of the affinity between a user and item:
\begin{enumerate}
    \item A pre-existing short-term model predicts the chance a user will briefly engage with an item if it is recommended to them. We refer to this as a \emph{clickiness} score.
    \item A long-term \emph{stickiness} model forecasts the strength of users' future listening habits with an item as a function of features of the item, features of the user, and the user's historical engagement with that specific item. In particular, the final implementation predicts 60-day return days: the number of days over the next 60 days in which a user will return to listen to the item.
\end{enumerate}
The researchers derive an overall score for each potential recommendation by combining the predictions of clickiness and stickiness models. Repeated A/B tests show this methodology helps foster lasting relationships between users and creators and leads to increased engagement with the app. At the time of their publication, this approach was being used in production in several parts of the app.

Our work addresses a major shortcoming of the stickiness models: they exacerbate the challenge of rapidly evaluating fresh content. 
Directly observing whether users tend to ``stick to'' a newly released podcast show requires waiting 60 days after they first discover the show. This introduces a delay of at least 60 days before the models can identify that newly released content is unusually sticky (and for whom it is sticky). We leverage progressive feedback to accelerate inferences about the stickiness of newly released content.

\begin{wrapfigure}{r}{0.45\textwidth}
    \vspace{-10mm}
  \begin{center}
    \includegraphics[width=0.3\textwidth]{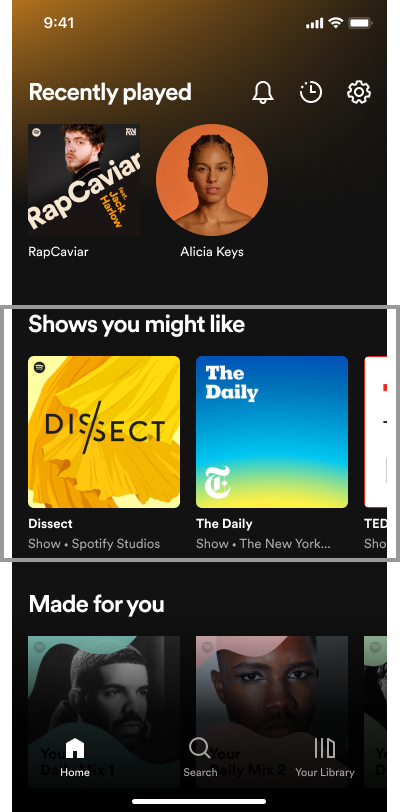}
    \vspace{-6mm}
  \end{center}
      \caption{Screenshot of Spotify mobile application. The highlighted podcast discovery shelf contains up to 20 cards that the user can scroll through.}
  \label{fig:screenshots}
  \vspace{-13mm}
\end{wrapfigure}

\subsection{A specific recommendation problem: ranking a shelf of discovery recommendations}

The methodology we describe now governs podcast recommendations in several parts of the app, but we focus here on results from an A/B test that intervenes on the ranking logic for a specific home page shelf. Figure \ref{fig:screenshots} shows an example of such a recommendation shelf, which helps users discover new content by displaying only podcast shows they haven't listened to before. Like many other personalized recommendation systems, content selection for this shelf follows a two-step process: first, a candidate generation step filters millions of podcast shows down to a manageable personalized list; then, a ranking step determines the order in which these candidates are displayed. The ranking step has a large impact on the likelihood that a recommendation is seen (or ``impressed'') by the user. Our A/B test intervenes only on this ranking logic, leaving the candidate generation step unchanged.

\subsection{Some metrics of interest}
\label{sec:metrics}
We now define the metrics used in our algorithms and results. 
It is worth emphasizing that the first four metrics, around discoveries and their durability, relate to user behavior that could happen anywhere on the app. 
For instance, a user can discover an item by searching actively for it.  Only the final metric attempts to directly link user behavior to actions of the recommender system. 
\begin{description}
\item[Discovery] When a user listens to part of an episode of a previously undiscovered podcast show, they are said to have "discovered" the podcast. On the "Shows you might like" shelf, all recommended podcasts are new to the user. 
\item[60-day minutes (following a discovery)] The total number of minutes spent listening to the podcast during the 60 days following a discovery (includes the day it was discovered).
\item[60-day return days (following a discovery)] The number of days a user returns to the podcast during the 60-day period after discovering it, excluding the discovery day itself; this metric takes values in $[0, 59]$. 
\item[60-day active days (following a discovery)] The total number of days a user listens to the podcast during the 60-day period following the discovery (includes the day it was discovered). Equal to return days plus one. 
\item[Impression] A user has an "impression" of a podcast recommendation when the recommendation appears on their screen. A user viewing the screen in Figure \ref{fig:screenshots} has an impression of two shows on the "Shows you might like" shelf, with additional impressions occurring as they scroll right on the shelf.
\item[60-day engagement attributable to an impression] When an impression leads directly to a discovery (through clicking and listening to that recommendation), the subsequent 60-day minutes are \emph{attributable} to the impression. If an impression does not directly result in a discovery, zero minutes are attributable, even if the user later discovers the item through another source. The 60-day return days and 60-day active days attributable to an impression are defined analogously.
\end{description}
In evaluating recommendation efficacy across users, we primarily measure success on a \emph{per-impression} basis, such as discoveries per impression or 60-day minutes per impression. We also examine metrics that focus on the durability of discoveries, such as 60-day return days \emph{per-discovery}.

\subsection{Recommendation Algorithm Variants}
\label{sec:variants}

In the A/B test, each user is randomized to one of the following recommendation ranking algorithms for the duration of the test. Note that the algorithm variants all employ deterministic ranking strategies.
Each algorithm is periodically refit using available batch data on the same regular cadence (e.g. daily).

\paragraph{(1) Legacy Approach: Optimizing for the Number of Discoveries.}
The legacy approach aimed to maximize the probability that users would "discover" a new podcast given an impression. A machine learning model was trained to fit these probabilities from features of the user and item, including powerful learned embeddings of each. Items in the shelf were ranked in descending order of these probabilities. This approach can be viewed as an effort to maximize a short-term proxy reward.
Note that we do not directly compare to these legacy approaches in this work; previous work has already demonstrated the benefit of modeling a long-term reward, as opposed to using discoveries as a proxy reward \citep{maystre2023optimizing}. The approach to modeling long-term rewards taken in this previous work serves as the control policy we compare to in the A/B test (discussed next).

\paragraph{(2) Control Policy: Delayed Rewards Model of Stickiness.}
The control policy used in our A/B test was learned using the approach introduced in \cite{maystre2023optimizing}. For discovery recommendations, the authors suggest scoring a potential recommendation to a user via the metric
\begin{equation}\label{eq:overall_scores}
    \texttt{score} = \underbrace{P(\text{discovery} \mid \text{impression} )}_{\text{clickiness}} \times \underbrace{\mathbb{E}\left[ \text{60-day active days}  \mid \text{discovery} \right]}_{\text{stickiness}},
\end{equation}
which equals the expected number of 60-day active days attributable to an impression. The personalized probability of a discovery given an impression was modeled using the legacy approach. Through the logic above, this is augmented by incorporating personalized stickiness estimates. For every given item $a$, they gather a dataset 
\[
\mathcal{D}_a = \bigg\{ \bigg( X_u, \, Y_u := (Y_u^{(1)}, \ldots, Y_u^{(60)}), \, R_u :=\sum_{j=1}^{60} Y_u^{(j)} \bigg) \bigg\}, 
\]
which ranges over users $u$ who discovered item $a$ at least 60 days ago. Discoveries that happened anywhere on the app over a long time-span are pooled together. 
For each such user, the dataset contains a tuple consisting of a pre-trained vector representation of user tastes $X_u$, an activity trace $Y_u$ where $Y_u^{(j)}\in \{0,1\}$ indicates whether user $u$ listened to the item $j$ days post-discovery, and the \emph{post-discovery} engagement $R_u=\sum_{j=1}^{60} Y_u^{(j)}$ equal to the total number of active days within 60 days. Stickiness is estimated through a linear model
\begin{equation}\label{eq:stickiness_regression}
\text{stickiness}(a, u) = \theta_a^\top X_u  \quad \text{where} \quad \theta_a = \argmin_{\theta} \sum_{(X,Y, R) \in \mathcal{D}_a} (X^\top \theta  - R)^2.
\end{equation}
We refer to a $\theta_a$ as the stickiness vector (or stickiness embedding) of item $a$. Using observed data on post-discovery listening habits enables learning of items that are unusually ``sticky'' (and for whom they are sticky). Note, however, that this estimation procedure cannot learn from discoveries that occurred within the last 60 days. Newer content cannot be evaluated, and the system defaults to a fixed, content-agnostic stickiness in this case. This is the main limitation we seek to address.

\paragraph{(3) Treatment Policy: Progressive Feedback Model of Stickiness.}
The treatment policy in the A/B test uses progressive feedback to draw faster inferences about items' stickiness vectors $\theta_a$; all other aspects of the algorithm, including the scoring formula \eqref{eq:overall_scores} and clickiness model are the same as in the control policy. Observe that the post-discovery engagement $R_u=\sum_{j=1}^{60} Y_u^{(j)}$ 
is only fully observable after a 60-day delay, but the activity trace $Y_u=(Y_u^{(1)}, \ldots, Y_u^{(60)})$ on which it depends is revealed progressively. We apply a contextual generalization of the Gaussian filtering algorithm (see Appendix \ref{app:contextVersion}) to update beliefs about an item's stickiness vectors $\theta_a$ from censored activity traces. Potential recommendations are scored using the posterior mean of stickiness predictions in the score \eqref{eq:overall_scores}, and, like the control policy, items are ranked in decreasing order of their (personalized) scores. The A/B test isolates the benefit of incorporating progressive feedback into stickiness estimates, showing large benefits even in a highly complex industrial-scale system.

\subsection{A/B Test Results}

The use of progressive feedback is especially important when recommending recently released shows, because they lack of historical data and since there are fewer long-term engagement outcomes observed due to the delay structure. For this reason, we discuss the A/B test results at two scales: impact when recommending recently released shows versus impact on the quality of recommendations across all podcasts.

\subsubsection{Efficacy When Recommending Recently Released Shows}

\begin{figure}[t]
    \centering
    \includegraphics{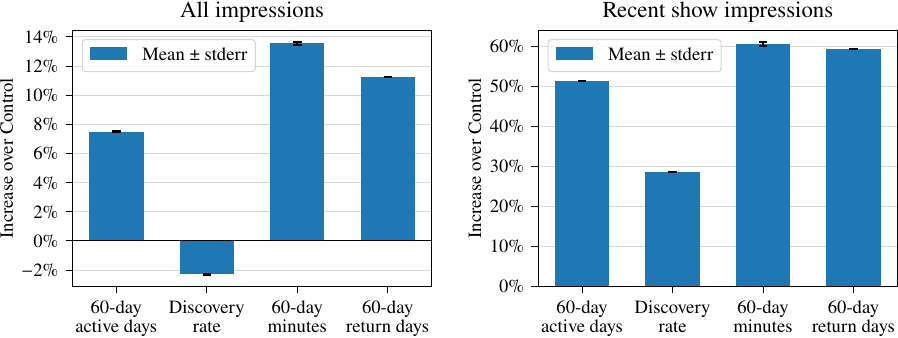}
    \vspace{-5mm}
    \caption{Results of treatment policy compared to control policy for recommending all shows (left) and for recently-released shows only (right). Note that all metrics are computed using only engagement attributable to an impression (as defined in Section \ref{sec:metrics}).}
    \label{fig:spotify-cst}
\end{figure}

Table~\ref{tab:spotify-vol} and Figure~\ref{fig:spotify-cst} (right) summarize the  effects of the A/B test. The key findings are as follows:
\begin{enumerate}
    \item {\bf No change in the volume of recommendations of recent releases.} 
    The total number of impressions of recently released shows is virtually the same between control and treatment groups. The treatment policy alters \emph{which} recent shows are recommended and \emph{to whom} they are recommended, while keeping the overall frequency of recent show recommendations unchanged.
	\item {\bf Huge increases in the quality of recommendations of recent releases.} When recommending recently released shows, the discovery rate (or number of discoveries per impression) was nearly 30\% higher under treatment than control.  These discoveries were also more lasting, with 60-day active days per-impression, 60-day minutes per-impression, and 60-day return days per-impression increased by over $50\%$ as compared to control.\footnote{The metric 60-day minutes per-impression is equal to the total number of 60-day minutes attributable to impressions  during the test (summed over all impressions of recent releases on the shelf by any user), divided by the number of impressions.} Note that these are \textit{huge} improvements in the context of A/B testing \citep{zhang2023evaluating,azevedo2020b,azevedo2019empirical,rosenthal1994parametric}.
\end{enumerate}

\begin{table}[t]
  \caption{In the Spotify A/B test, we observe no substantial difference in the volume of recommendations, recent or otherwise, across cells.}
  \vspace{2mm}
  \label{tab:spotify-vol}
  \centering
  \begin{tabular}{crr}
    \toprule
    & All shows & Recently-released shows only \\
    \midrule
    Relative difference, treatment vs. control
        & \num[retain-explicit-plus]{-0.02}\%
        & \num[retain-explicit-plus]{+2.15}\% \\
    \bottomrule
  \end{tabular}
\end{table}

\begin{figure}[h]
    \centering
    \includegraphics{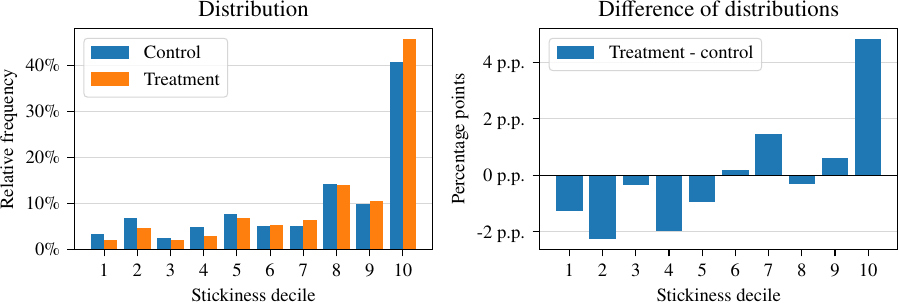}
    \vspace{-5mm}
    \caption{Stickiness of discoveries of recently released shows. 
    }
    \label{fig:spotify-cst-mechanism}
\end{figure}

Evidence that incorporating progressive feedback improved long-term recommendations is further supported by Figure \ref{fig:spotify-cst-mechanism}, where we examine how the recommendation policy increased the 60-day activity per recent show impression. In hindsight\footnote{The test results are being written up 6 months after the test.}, we have sufficient data to accurately estimate the average (unpersonalized) long-term value of each show that was recent during the test period. We leverage this data in Figure \ref{fig:spotify-cst-mechanism}, partitioning recent shows by their average long-term stickiness (measured in hindsight).

The analysis reveals that, relative to the control policy (delayed rewards), the treatment policy (progressive feedback) decreased impressions to less sticky shows, particularly among the 50\% least sticky shows. Instead, the algorithm allocated these impressions to highly sticky shows, increasing the relative fraction of impressions these shows receive. A plausible interpretation is that the treatment policy successfully identified and redistributed impressions based on early indicators of shows' long-term value, which proved to correlate with the shows' eventual stickiness.

\subsubsection{Recommendations Beyond Recently Released Shows}
We have seen that incorporating progressive feedback into stickiness estimates led to substantial improvements in the quality of recommendations for recently released shows, a critical part of the system's overall promise for users and creators.

The changes also led to meaningful improvements in metrics for the shelf overall, beyond recently released shows. Table~\ref{tab:spotify-vol} and Figure \ref{fig:spotify-cst} (left) summarize the findings. The number of impressions on the shelf remained virtually identical between treatment and control groups, suggesting that differences in other metrics are primarily attributable to the quality of the personalized recommendations.

Compared to the control algorithm, the treatment algorithm led to a 7\% increase in 60-day active days per impression and over 10\% increases in both 60-day minutes per impression and 60-day return days per impression. In line with findings in \cite{maystre2023optimizing}, these improvements in long-term metrics came at the expense of a slight degradation in short-term metrics: the treatment group had slightly fewer discoveries per impression than the control group.

\subsection{Subsequent testing and rollout}
 Following the initial A/B test, the method underwent further testing in more diverse parts of the app, building trust in the initial findings. First, a subsequent A/B test applied the model to several additional shelves and to recommendations on the podcast sub-feed, allowing for broader impact. Then, a method leveraging our Bayesian filtering procedure was rolled out to all users. The rollout process itself served as another controlled experiment by gradually ramping up the proportion of users receiving the treatment algorithm \citep[Chapter 15]{kohavi2020trustworthy}.
\section{Discussion}

In this work, we develop a Thompson sampling based algorithm that can take advantage of intermediate progressive feedback in a setting in which observing the true reward is delayed. We found in both simulation settings and a real A/B test for Spotify podcast recommendations, that the algorithm outperformed methods that do not take advantage of progressive feedback. Given this work, there are several directions for future research, which we discuss below.

The Impatient Bandit algorithm assumes that the prior distribution is learned correctly. A limitation of this work is that we did not provide any guarantees regarding under what settings one would expect to learn an accurate prior from previously collected data. We found in our experiments that we were able to learn an effective prior from data. Recent works have begun to characterize the regret of Thompson Sampling with learned priors \citep{simchowitz2021bayesian,zhang2024posterior}.

We discuss how the Impatient Bandit algorithm is able to incorporate user context features, when expected outcomes are linear functions of these contexts (see Appendix \ref{app:contextVersion}). A limitation of our current algorithm is that we are restricted to linear and kernel type outcome models, which enables us to use Gaussian Bayesian inference methods. An open question is how to computationally efficiently update the posterior distribution for other types of outcomes. Another direction for future research is to extend our regret analysis to settings with more general outcome models.
Finally, it would be of interest to extend our algorithm to handle Markov Decision Process environments with progressive feedback and/or delayed rewards.

\section*{Acknowledgements}
We thank the reviewers for their thoughtful comments, which helped us improve the paper. We also thank Mounia Lalmas-Roelleke for feedback and discussions on this work.

\clearpage
\section*{Overview of Appendices}
\begin{itemize}
    \item Appendix \ref{app:gaussianModel}: Proof of Lemma \ref{lem:joint_normal} (Properties of Multivariate Gaussian Model)
    \item Appendix \ref{app:posterior}: Deriving Incremental Posterior Updates
    \item Appendix \ref{app:contextVersion}: Impatient Bandit Algorithm with Context
    \item Appendix \ref{app:regretBounds}: Proof of Theorem \ref{thm:main} and Corollary \ref{corr:mainNoZ} (Regret Bounds)
    \item Appendix \ref{app:rounding}: Rounding Procedure
\end{itemize}

\appendix
\section{Multivariate Gaussian Bayesian Model: Proof of Lemma \ref{lem:joint_normal}}
\label{app:gaussianModel}

\begin{customlemma}{\ref{lem:joint_normal}}[Multivariate Gaussian Bayesian Model]
    Let Assumptions \ref{assum:exchangeable} and \ref{assum:gaussian} hold. The limit $\theta_a \triangleq \lim_{|U| \to \infty} \frac{1}{|U|} \sum_{u \in U} Y_{u}(a)$ exists almost surely. Conditional on $\theta_a, Z_a$, the random variables $\left\{ Y_u(a) \right\}_{u\in \Uc}$ are i.i.d. 
    Moreover,
    \begin{align}
        \theta_a \mid ( Z_a = z ) \sim N \left( \mu_{1, z} \,,\, \Sigma_{1, z} \right) \qquad  \text{and} \qquad Y_u(a) \mid (\theta_a, Z_a = z) \sim N( \theta_a \,,\, V_z ). 
        \label{eqn:hierarchical}
    \end{align}
\end{customlemma}

\begin{proof}
As discussed earlier below Remark \ref{rem:exchangeable}, by exchangeability Assumption \ref{assum:exchangeable} and De Finetti's theorem \citep{heath1976finetti,de1937prevision}, for any $a \in \MC{A}$, there exists some (potentially) random outcome distribution which we denote using $P_a$ that may depend on $Z_a$ that can be used to describe the data generating process for $\{ Y_u(a) \}_{u \in U}$:
\begin{align}
    \label{eqn:condIIDdefinetti}
    \TN{sample } P_a \mid Z_a, \TN{ then draw } Y_1(a), Y_2(a), Y_3(a), \dots \iidsim P_a.
\end{align}
Note that $\E[ |Y_u(a)| \mid P_a ] < \infty$ w.p. $1$. If this were not the case, then $\E[ |Y_u(a)| \mid P_a ]$ could be infinite with non-zero probability, which would imply $\E[ |Y_u(a)| ] = \E[ \E[ |Y_u(a)| \mid P_a ] ]$ is infinite and violate Assumption \ref{assum:gaussian}.
Thus, we can apply the strong Law of Large Numbers after conditioning on the draw of $P_a$ to get the following result:
\begin{align*}
    \PP \left( \lim_{|U| \to \infty} \bar{Y}_U(a) = \E \big[ Y_{u}(a) \mid P_a \big] ~ \bigg| ~ P_a, Z_a 
    \right) = 1 ~~ \TN{w.p.}~1,
\end{align*}
where $\bar{Y}_U(a) \triangleq \frac{1}{|U|} \sum_{u \in U} Y_{u}(a)$. Since the above holds with probability $1$, it implies that
\begin{align}
    \PP \left( \lim_{|U| \to \infty} \bar{Y}_U(a) = \E \big[ Y_{u}(a) \mid P_a \big] ~ \bigg| ~ Z_a 
    \right) = 1 ~~ \TN{w.p.}~1.
    \label{eqn:SLLNresult}
\end{align}
Thus we have shown that, with probability 1, $\theta_a \triangleq\lim_{|U| \to \infty} \bar{Y}_{U}(a)$ exists and is equal to $\E \big[ Y_{u}(a) \mid P_a \big]$.  

Almost sure convergence implies convergence in distribution. Therefore, $\lim_{|U|\to \infty} \mathcal{L}( \bar{Y}_{U}(a) \mid Z_a=z) = \mathcal{L}(\theta_a \mid Z_a=z)$ where $\mathcal{L}(X)$ denotes the law (or probability distribution) of a random variable $X$.

To show \eqref{eqn:hierarchical},  we calculate $\mathcal{L}( \bar{Y}_{U}(a) \mid Z_a=z)$ and then take the limit as $|U| \to \infty.$ Note that by Assumptions \ref{assum:exchangeable} and \ref{assum:gaussian}, the joint distribution of $\left( Y_u(a) \right)_{u \in U}$ for any $U$ must be of the following form:
\begin{align*}
    \begin{pmatrix}
        Y_1(a) \\
        Y_2(a) \\
        Y_3(a) \\
        \vdots \\
        Y_{|U|}(a)
    \end{pmatrix} \bigg| Z_a = z
    \sim N \left( 
    \begin{bmatrix} \mu_{1,z} \\ \mu_{1,z} \\ \mu_{1,z} \\ \vdots \\ \mu_{1,z} \\ \end{bmatrix},
    \begin{bmatrix} 
    \Sigma_{1,z} + V_z &  \Sigma_{1,z} & \Sigma_{1,z} & \hdots & \Sigma_{1,z} \\ 
    \Sigma_{1,z} &  \Sigma_{1,z} + V_z & \Sigma_{1,z} & \hdots & \Sigma_{1,z} \\ 
    \Sigma_{1,z} &  \Sigma_{1,z} & \Sigma_{1,z} + V_z & \hdots & \Sigma_{1,z} \\ 
    \vdots & \vdots & \vdots & \ddots & \vdots \\
    \Sigma_{1,z} & \Sigma_{1,z} & \Sigma_{1,z} & \hdots & \Sigma_{1,z} + V_z
    \end{bmatrix}
    \right)
\end{align*}
Using that the sum of jointly Gaussian random variables is Gaussian and linearity properties of Gaussian random variables, we find,
\begin{align*}
    \bar{Y}_{U} \mid Z_a = z \, \sim \, N \left( \mu_{1,z}, \, \frac{(|U|-1)}{|U|} \Sigma_{1,z} + \frac{1}{|U|} V_z \right).
\end{align*}
Thus, $\lim_{|U| \to \infty} \mathcal{L}\left( \bar{Y}_{U} \mid Z_a = z \right) = N(\mu_{1,z}, \Sigma_{1,z})$. Thus, we have shown that  $\theta_a  \sim \, N(\mu_{1,z}, \Sigma_{1,z})$. 

One can generalize the above argument to show that $\begin{pmatrix}
        \theta_a \\
        \{ Y_u(a) \}_{u \in \Uc}
    \end{pmatrix} \bigg| Z_a = z$ is a Gaussian process.
Specifically, for any $u, u' \in \Uc$ with $u \not= u'$,
\begin{align*}
    \begin{pmatrix}
        \theta_a \\
        Y_u(a) \\
        Y_{u'}(a) 
    \end{pmatrix} \bigg| Z_a = z
    \sim N \left( 
    \begin{bmatrix} \mu_{1,z} \\ \mu_{1,z} \\ \mu_{1,z} \end{bmatrix},
    \begin{bmatrix} 
    \Sigma_{1,z} & \Sigma_{1,z} & \Sigma_{1,z} \\
    \Sigma_{1,z} & \Sigma_{1,z} + V_z & \Sigma_{1,z} \\ 
    \Sigma_{1,z} & \Sigma_{1,z} & \Sigma_{1,z} + V_z
    \end{bmatrix}
    \right)
\end{align*}
By properties of conditioning on Gaussian random variables,
\begin{align*}
    \begin{pmatrix}
        Y_u(a) \\
        Y_{u'}(a)
    \end{pmatrix} \bigg| (Z_a = z, \theta_a)
    \sim N \left( 
    \begin{bmatrix} \theta_a \\ \theta_a \end{bmatrix},
    \begin{bmatrix} 
    V_z & 0 \\ 
    0 & V_z 
    \end{bmatrix}
    \right)
\end{align*}
By the above, we have that $Y_u(a) \mid (\theta_a, Z_a = z) \sim N( \theta_a \,,\, V_z )$. Moreover, $Y_u(a) \indep Y_{u'}(a) \mid (Z_a = z, \theta_a)$, since for jointly Gaussian random variables, zero correlation implies independence. Moreover, under Assumption \ref{assum:exchangeable}, $\{ Y_u(a) \}_{u \in \Uc}$ are exchangeable conditional on $Z_a$, thus we have that conditional on $\theta_a, Z_a$, the random variables $\left\{ Y_u(a) \right\}_{u\in U}$ are i.i.d.
\end{proof}


\section{Incremental Posterior Updates}
\label{app:posterior}

Since the outcomes $Y_u^{(1)}, Y_u^{(2)}, \dots, Y_u^{(J)}$ are revealed incrementally over time after selecting action $A_u = a$, the posterior distribution for $\theta_a$ must accordingly be updated with this incremental information. In this Gaussian case, there is a closed form formula for the posterior distribution. Moreover, the posterior updates can happen incrementally, as we get more information on the outcomes $Y_u^{(1)}, Y_u^{(2)}, \dots, Y_u^{(J)}$.

Updating the posterior distribution in this setting is slightly different from standard Gaussian bandit setting because the outcomes $Y_u^{(1)}, Y_u^{(2)}, \dots, Y_u^{(J)}$ from a single user $u$ over time are correlated according to the covariance matrix $V_z$ (see the end of Section \ref{sec:gaussian}). Specifically, the dimensions of the “noise'' vector $Y_u - \theta_a \in \real^J$ are correlated according to $V_z$:
\begin{align*}
    \quad Y_u - \theta_a \mid ( A_u = a, Z_a = z ) \sim N( 0, V_z ) 
    \quad \TN{where} \quad 
    \theta_a \mid (Z_a = z) \sim N( \mu_{1,z}, \Sigma_{1,z} ).
\end{align*}

Incremental updating of the posterior can be done computationally inexpensively with a clever use of the Cholesky Decomposition \citep{Haddad2009}.
Assume the covariance matrix $V_z$ is invertible. Otherwise, one of the $J$ engagement observations provides redundant information and could be ignored without loss. By the Cholesky decomposition, since $V_z$ is a symmetric positive definite matrix, $V_z = L_z L_z^\top$ for a unique lower triangular matrix $L_z \in \real^{J \by J}$. The inverse matrix $L^{-1}_z$ is also lower triangular. Let $\ell_z^{(j)}$ denotes row $j$ of $L_z^{-1}$. Then define the transformed engagement observations, 
\begin{align}
\MC{Y}_u \ =  \begin{pmatrix}
\MC{Y}_u^{(1)} \\
\MC{Y}_u^{(2)} \\
\vdots \\
\MC{Y}_u^{(J)}  
\end{pmatrix} \triangleq  
\begin{pmatrix}
(\ell_z^{(1)})^\top  Y_u  \\
(\ell_z^{(2)})^\top  Y_u  \\
\vdots \\
(\ell_z^{(J)})^\top  Y_u  
\end{pmatrix} =  L_z^{-1} Y_u. 
\label{eqn:scriptYdef}
\end{align}
Working with the transformed vectors provides a convenient expression for the posterior (see Proposition \ref{prop:posterior} below). The formula is analogous to ones that appear in treatments of Bayesian linear regression \citep{agrawal2013thompson,rasmussen2006gaussian}, where $\ell_z^{(1)},\ldots, \ell_z^{(J)}$ act as fixed “context feature vectors'' derived from the prior covariance $V_z$. 

A subtlety in deriving this result is that entries of $Y_u$ are observed progressively, rather than all at once. The proof shows that, since both $L_z$ and $L_z^{-1}$ are lower triangular, the first $j$ entries of $\MC{Y}_u$ can be computed from the first $j$ entries of $Y_u$, and vice versa. The components of the transformed engagements $\MC{Y}_u$ become `uncensored' at the same time as the raw engagements $Y_u$ and provide identical information to the system. A detailed proof is given in Appendix \ref{app:posterior}.

\begin{algorithm}
    \label{alg:posterior}
    \caption{Posterior Update} 
    \begin{algorithmic}[1]
        \State \bo{Inputs:} $\big\{ \mu_{t,a}, \Sigma_{t,a}, V_{Z_a} \big\}_{a \in \MC{A}}, ~ \big\{ \tau_u, A_u, \tilde{Y}_{u,t} \big\}_{u \in \Uc_{\leq t}}$ \Comment{Posterior parameters, Previous data}
        \For{$a \in \MC{A}$}
            \State For all $u \in \Uc_{\leq t}$ and $j \in [1 \colon J]$ such that $t = \tau_u + d_j$, compute
            \begin{align*}
                \MC{Y}_u^{(j)} \gets ( \ell_z^{(j)} )^\top \begin{bmatrix}
                    Y_u^{(1:j)} \\
                    \bs{0}_{J-j}
                \end{bmatrix}
                \quad \TN{ where } \bs{0}_{J-j} \in \real^{J-j} \TN{ is a vector of zeros}
            \end{align*}
            \State $\Sigma_{t+1,a} \gets \left( \Sigma_{t,a}^{-1} + \sum_{u \in \Uc_{\leq t}} \ind\{A_u = a\} \sum_{j=1}^J \ind\{t = \tau_u + d_j\} \ell_z^{(j)} (\ell_z^{(j)})^\top \right)^{-1}$
            \State $\mu_{t+1,a} \gets \Sigma_{t+1,a} \left( \Sigma_{t,a}^{-1} \mu_{t,a} + \sum_{u \in \Uc_{\leq t}} \ind\{A_u = a\} \sum_{j=1}^J \ind\{t = \tau_u + d_j\} \ell_z^{(j)} \MC{Y}_u^{(j)} \right)$
        \EndFor
        \State \bo{Output:} $\{ \mu_{t+1,a}, \Sigma_{t+1,a} \}_{a \in \MC{A}}$ \Comment{Posterior parameters}
    \end{algorithmic}
\end{algorithm}

\begin{proposition}[Posterior Distribution]
    Let Assumptions \ref{assum:exchangeable} and \ref{assum:gaussian} hold, and let $V_z$ and $\Sigma_{0,z}$ be positive definite. Then, $\theta_a \mid \HH_{t} \sim N( \mu_{t,a}, \Sigma_{t,a} )$ where
    \begin{align*}
        \Sigma_{t,a} \triangleq \bigg( \Sigma_{1,z}^{-1} + \sum_{u \in \Uc_{<t}} \ind\{A_u = a\} \sum_{j=1}^J \ind\{t > \tau_u + d_j\} \ell_z^{(j)} (\ell_z^{(j)})^\top \bigg)^{-1}
    \end{align*}
    \vspace{-3mm}
    and
    \vspace{-3mm}
    \begin{align*}
        \mu_{t,a} \triangleq \Sigma_{t,a} \bigg( \Sigma_{1,z}^{-1} \mu_{1,z} + \sum_{u \in \Uc_{<t}} \ind\{A_u = a\} \sum_{j=1}^J \ind\{t > \tau_u + d_j\} \ell_z^{(j)} \MC{Y}_u^{(j)} \bigg).
    \end{align*}
    \label{prop:posterior}
\end{proposition}

\begin{proof}
It is sufficient to show that if $\theta_a \mid (Z_a = z) \sim N( \mu_{1,z}, \Sigma_{1,z} )$ then, for any $k \in [1 \colon J]$,
\begin{align*}
    \theta_a \mid \big( Y_u^{(1)}, Y_u^{(2)}, \dots, Y_u^{(k)}, A_u = a \big) 
    \sim N \left( \mupost, \Sigmapost \right)
\end{align*}
where for $\MC{Y}_u^{(j)}$ defined as in \eqref{eqn:scriptYdef},
\begin{align}
    \Sigmapost \triangleq \bigg\{ \Sigma_{1,z}^{-1} + \sum_{j=1}^k \ell_z^{(j)} (\ell_z^{(j)})^\top \bigg\}^{-1} 
    \quad \TN{and} \quad
    \mupost \triangleq \Sigmapost \bigg( \Sigma_{1,z}^{-1} \mu_{1,z} + \sum_{j=1}^k \ell_z^{(j)} \MC{Y}_u^{(j)} \bigg).
    \label{eqn:SigmaMuPost}
\end{align}

Since the covariance matrix $V_z$ is invertible, by the Cholesky decomposition there is a unique, positive definite lower triangular matrix $L_z \in \real^{J \by J}$ such that $V_z = L_z L_z^\top$. Using the matrix $L_z$, we can define a version of $Y_u$, whose noise covariance matrix is ``de-correlated'':
\begin{align*}
    \MC{Y}_u \triangleq L_z^{-1} Y_u
    = L_z^{-1} \big( Y_u - \theta_a \big) + L_z^{-1} \theta_a.
\end{align*}
Above $\theta_a \mid (Z_a = z) \sim N( \mu_{1,z}, \Sigma_0 )$ and $L_z^{-1} \big( Y_u - \theta_a \big) \mid ( A_u = a, Z_a = z, \theta_a ) \sim N \big( 0, I_J \big)$ where $I_J \in \real^{J \by J}$ is the identity matrix; note that $L_z^{-1} V_z (L_z^{-1})^{\top} = I_J$ since $V_z = L_z L_z^\top$ by definition.

Thus $\MC{Y}_u^{(j)}$, the $j^{\TN{th}}$ element of the vector $\MC{Y}_u$, can be written as
\begin{align*}
    \MC{Y}_u^{(j)} \triangleq (\ell_z^{(j)})^\top Y_u \underbrace{=}_{(a)} (\ell_z^{(j)})^\top \begin{bmatrix}
        Y_u^{(1:j)} \\
        \bs{0}_{J-j}
    \end{bmatrix} \quad \TN{where} \quad \MC{Y}_u^{(j)} \mid ( A_u = a, Z_a = z, \theta_a ) \sim N \big( (\ell_z^{(j)})^\top \theta_a, ~ 1 \big).
\end{align*}
Above, $(\ell_z^{(j)})^\top$ is the $j^{\TN{th}}$ row of the matrix $L_z^{-1}$. We now explain why equality (a) above holds. By the lower-triangular structure of the non-singular matrix $L_z$ (due to the Cholesky composition), the inverse matrix $L_z^{-1}$ is also lower-triangular \citep{meyer1970generalized}. Thus, the last $J-j$ dimensions of $(\ell_z^{(j)})^\top$, the $j^{\TN{th}}$ row of $L_z^{-1}$, must all be zeros.

Since $\theta_a \mid (Z_a = z) \sim N( \mu_{1,z}, \Sigma_{1,z} )$ and $ \MC{Y}_u^{(j)} \mid ( A_u = a, Z_a = z, \theta_a ) \sim N \big( (\ell_z^{(j)})^\top \theta_a, ~ 1 \big)$ by the display above, by conditioning properties of multivariate Gaussians (see Section 2.2 and Appendix A.1 in \cite{agrawal2013thompson} for details), we have 
\begin{align*}
    &\theta_a \mid \big( \MC{Y}_u^{(1)}, \MC{Y}_u^{(2)}, \dots, \MC{Y}_u^{(k)}, A_u = a \big) \sim N \left( \mupost, \Sigmapost \right),
\end{align*}
for $\mupost$ and $\Sigmapost$ as defined in \eqref{eqn:SigmaMuPost}.

The final step is to show that the following are equal in distribution:
\begin{align*}
    \theta_a \mid \big( \MC{Y}_u^{(1)}, \MC{Y}_u^{(2)}, \dots, \MC{Y}_u^{(k)}, A_u = a, Z_a = z \big)
    ~~~ \overset{D}{=} ~~~ 
    \theta_a \mid \big( Y_u^{(1)}, Y_u^{(2)}, \dots, Y_u^{(k)}, A_u = a, Z_a = z \big).
\end{align*}
Note it is sufficient to show that $\MC{Y}_u^{(1:k)} \triangleq \big( \MC{Y}_u^{(1)}, \MC{Y}_u^{(2)}, \dots, \MC{Y}_u^{(k)} \big)$ is a one-to-one (injective), non-random transformation of $Y_u^{(1:k)} \triangleq \big(Y_u^{(1)}, Y_u^{(2)}, \dots, Y_u^{(k)} \big)$. We prove this below.

Since $\MC{Y}_u^{(j)} = (\ell_z^{(j)})^\top \begin{bmatrix}
        Y_u^{(1:j)} \\
        \bs{0}_{J-j}
    \end{bmatrix}$, we have that $\begin{bmatrix}
        \MC{Y}_u^{(1:k)} \\
        \bs{0}_{J-k}
    \end{bmatrix} = L_z^{-1} \begin{bmatrix}
        Y_u^{(1:k)} \\
        \bs{0}_{J-k}
    \end{bmatrix}$.
It is sufficient to show that the function $f(x) = L_z^{-1} x$ for any vector $x \in \real^J$ is injective. Suppose that $f(x) = f(x')$, i.e., $L_z^{-1} x = L_z^{-1} x'$. Since $L_z^{-1}$ is positive definite, by multiplying both sides by $L_z$ it must be that $x = x'$. This $f$ is an injective function.

\end{proof}

\section{Impatient Bandit Algorithm with Context}
\label{app:contextVersion}

In this section, we write the impatient bandit algorithm extension to linear contextual bandits. We introduce new notation and redefine previous notation that will be used in this section alone. Each user $u \in \Uc$ has an associated context vector $X_u \in \real^k$ that is observed prior to making the recommendation decision. The algorithm posits the following data generating process for each item $a \in \MC{A}$:
\begin{align*}
    \theta_a \mid ( Z_a = z ) \sim N \left( \mu_{1, z} \,,\, \Sigma_{1, z} \right)
\end{align*}
Above $\theta_a, \mu_{1,z} \in \real^{k \cdot J}$ and $\Sigma_{1,z} \in \real^{(k \cdot J) \by (k \cdot J)}$. We use the notation $\theta_a = \big( \theta_a^{(1)}, \theta_a^{(1)}, \dots, \theta_a^{(J)} \big)$ where each $\theta_a^{(j)} \in \real^k$. Similarly, we will use the notation $\mu_{1,z} = \big( \mu_{1,z}^{(1)}, \mu_{1,z}^{(2)}, \dots, \mu_{1,z}^{(J)} \big)$ where each $\mu_{1,z}^{(j)} \in \real^k$. Potential outcomes are now distributed as follows:
\begin{align*}
    Y_u(a) \mid (\theta_a, X_u, Z_a = z) \sim N \left( \begin{bmatrix}
        X_u^\top \theta_a^{(1)} \\
        \vdots \\
        X_u^\top \theta_a^{(J)}
    \end{bmatrix} \,,\, V_z \right).
\end{align*}
Above $V_z \in \real^{J \by J}$. We use $Y_u(a) = \big( Y_u^{(1)}(a), Y_u^{(2)}(a), \dots, Y_u^{(J)}(a) \big) \in \real^J$. We continue to use $\tilde{Y}_{u,t}^{(j)}$ to refer to the censored outcomes, as defined in \eqref{eqn:censoredYs}. We now redefine the history to include the past observed user contexts $X_u$:
\begin{align*}
    \HH_{t} \triangleq \bigcup_{u \in \Uc_{<t}} \left\{ A_{u}, X_u, \tilde{Y}_{u,t}^{(1)}(A_u), \tilde{Y}_{u,t}^{(2)}(A_u), \dots, \tilde{Y}_{u,t}^{(J)}(A_u) \right\} \cup \{ Z_a : a \in \MC{A} \}.
\end{align*}

\begin{algorithm}
    \caption{Impatient Bandit Thompson Sampling with Context $\piPS$}
    \begin{algorithmic}[1]
    \Require Priors and noise covariance matrices $\big\{ \mu_{1,Z_a}, \Sigma_{1,Z_a}, V_{Z_a} \big\}_{a \in \MC{A}}$
    \For{$a \in \MC{A}$}
        \State $(\mu_{1,a}, \Sigma_{1,a}) \gets (\mu_{1, Z_a}, \Sigma_{1, Z_a})$ \Comment{Initialize beliefs.}
    \EndFor
    \State $\MC{H}_1 \gets \{Z_a : a \in \MC{A}\}$ \Comment{Initialize history.}
    \For{$t = 1, 2, \dots, T$}
        \For{$u \in \Uc_t$}
            \State $\tau_u \gets t$ \Comment{Set user decision time.}
            \For{$a \in \MC{A}$} \label{line:startsample}
                \State $\theta_a' \sim N \left( \mu_{t,a} \,,\, \Sigma_{t,a} \right)$ \Comment{Sample from belief.}
            \EndFor
            \State $A_u \gets \argmax_{a \in \MC{A}} \left\{ R( \theta_{a, X_u}' ) \right\}$ for $\theta_{a, X_u}' \gets \begin{bmatrix}
                X_u^\top (\theta_a')^{(1)} \\
                \vdots \\
                X_u^\top (\theta_a')^{(J)}
            \end{bmatrix}$ \Comment{Take action.}
            \label{line:takeactionContext}
        \EndFor
        \State $\MC{H}_{t+1} \gets \MC{H}_t \cup \left\{ Y_u^{(j)}(A_u) : u \in \Uc_{\leq t}, \text{$j$ s.t. $t = \tau_u + d_j$} \right\}$ \Comment{Update history.}
        \For{$a \in \MC{A}$}
            \State $(\mu_{t+1,a}, \Sigma_{t+1,a}) \gets \textrm{PosteriorUpdate}(\mu_{t, Z_a}, \Sigma_{t, Z_a}, V_{Z_a}, \MC{H}_{t+1})$ \Comment{Update belief.}
        \EndFor
    \EndFor
    \end{algorithmic}
    \label{alg:banditcontext}
\end{algorithm}

Similar to the non-contextual case, we assume the covariance matrix $V_z$ is invertible. Since the covariance matrix $V_z$ is invertible, by the Cholesky decomposition there is a unique, positive definite lower triangular matrix $L_z \in \real^{J \by J}$ such that $V_z = L_z L_z^\top$.  We use $(\ell_z^{(j)})^\top$ to refer to the $j^{\TN{th}}$ row of the matrix $L_z^{-1}$. Furthermore, we use $\ell_z^{(j,i)}$ to refer to the $i^{\TN{th}}$ element of the vector $(\ell_z^{(j)})^\top$. The posterior update formula for the contextual version of impatient bandits will use a new featurization function of both the user context $X_u$ and vector $(\ell_z^{(j)})^\top$:
\begin{align}
    \phi \big( X_u, \ell_z^{(j)} \big) \triangleq \begin{bmatrix}
        X_u \ell_z^{(j, 1)} \\
        \vdots \\
        X_u \ell_z^{(j, J)}
    \end{bmatrix} \in \real^{J \cdot k}
    \label{eqn:phivector}
\end{align}

\begin{algorithm}[H]
    \label{alg:posteriorContext}
    \caption{Posterior Update with Context} 
    \begin{algorithmic}[1]
        \State \bo{Inputs:} $\big\{ \mu_{t,a}, \Sigma_{t,a}, V_{Z_a} \big\}_{a \in \MC{A}}, ~ \big\{ \tau_u, A_u, X_u, \tilde{Y}_{u,t+1} \big\}_{u \in \Uc_{\leq t}}$ \Comment{Posterior parameters, Previous data}
        \For{$a \in \MC{A}$}
            \State For all $u \in \Uc_{\leq t}$ and $j \in [1 \colon J]$ such that $t = \tau_u + d_j$, compute
            \begin{align*}
                \MC{Y}_u^{(j)} \gets ( \ell_z^{(j)} )^\top \begin{bmatrix}
                    Y_u^{(1:j)} \\
                    \bs{0}_{J-j}
                \end{bmatrix}
                \quad \TN{ where } \bs{0}_{J-j} \in \real^{J-j} \TN{ is a vector of zeros}
            \end{align*}
            \State $\Sigma_{t+1,a} \gets \left( \Sigma_{t,a}^{-1} + \sum_{u \in \Uc_{\leq t}} \ind\{A_u = a\} \sum_{j=1}^J \ind\{t = \tau_u + d_j\} \phi(X_u, \ell_z^{j}) \phi(X_u, \ell_z^{j})^\top \right)^{-1}$
            \State $\mu_{t+1,a} \gets \Sigma_{t+1,a} \left( \Sigma_{t+1,a}^{-1} \mu_{t,a} + \sum_{u \in \Uc_{\leq t}} \ind\{A_u = a\} \sum_{j=1}^J \ind\{t = \tau_u + d_j\} \phi(X_u, \ell_z^{j}) \MC{Y}_u^{(j)} \right)$
        \EndFor
        \State \bo{Output:} $\{ \mu_{t+1,a}, \Sigma_{t+1,a} \}_{a \in \MC{A}}$ \Comment{Posterior parameters}
    \end{algorithmic}
\end{algorithm}

\paragraph{Posterior Update Formula Derivation}
We now provide an informal justification for the posterior update formula, which extends the proof of Proposition \ref{prop:posterior}. Using the matrix $L_z$ (recall $L_z$ is lower triangular and $V_z = L_z L_z^\top$), we can define a version of $Y_u$, whose noise covariance matrix is “de-correlated'':
\begin{align*}
    \MC{Y}_u \triangleq L_z^{-1} Y_u
    = L_z^{-1} \left( Y_u - \begin{bmatrix}
        X_u^\top \theta_a^{(1)} \\
        \vdots \\
        X_u^\top \theta_a^{(J)}
    \end{bmatrix} \right) + L_z^{-1} \begin{bmatrix}
        X_u^\top \theta_a^{(1)} \\
        \vdots \\
        X_u^\top \theta_a^{(J)}
    \end{bmatrix}.
\end{align*}
Above $\theta_a \mid (X_u, Z_a = z) \sim N( \mu_{1,z}, \Sigma_{0,z} )$ and 
$$L_z^{-1} \left( Y_u - \begin{bmatrix}
        X_u^\top \theta_a^{(1)} \\
        \vdots \\
        X_u^\top \theta_a^{(J)}
    \end{bmatrix} \right) \bigg| ( X_u, A_u = a, Z_a = z, \theta_a ) \sim N \big( 0, I_J \big)$$
    where $I_J \in \real^{J \by J}$
    is the identity matrix; note that $L_z^{-1} V_z (L_z^{-1})^{\top} = I_J$ since $V_z = L_z L_z^\top$ by definition.

Thus $\MC{Y}_u^{(j)}$, the $j^{\TN{th}}$ element of the vector $\MC{Y}_u$, can be written as
\begin{align*}
    \MC{Y}_u^{(j)} \triangleq (\ell_z^{(j)})^\top Y_u \underbrace{=}_{(a)} (\ell_z^{(j)})^\top \begin{bmatrix}
        Y_u^{(1:j)} \\
        \bs{0}_{J-j}
    \end{bmatrix}
\end{align*}
where
\begin{align*}
    \MC{Y}_u^{(j)} \mid ( X_u, A_u = a, Z_a = z, \theta_a ) \sim N \left( (\ell_z^{(j)})^\top \begin{bmatrix}
        X_u^\top \theta_a^{(1)} \\
        \vdots \\
        X_u^\top \theta_a^{(j)} \\
        \bs{0}_{J-j}
    \end{bmatrix}, ~ 1 \right)
\end{align*}
Above $(\ell_z^{(j)})^\top$ is the $j^{\TN{th}}$ row of the matrix $L_z^{-1}$. We now explain why equality (a) above holds. By the lower-triangular structure of the non-singular matrix $L_z$ (due to the Cholesky composition), the inverse matrix $L_z^{-1}$ is also lower-triangular \citep{meyer1970generalized}. Thus, the last $J-j$ dimensions of $(\ell_z^{(j)})^\top$, the $j^{\TN{th}}$ row of $L_z^{-1}$, must all be zeros.

Furthermore, note that by taking the transpose and rearranging terms,
\begin{align*}
    (\ell_z^{(j)})^\top \begin{bmatrix}
        X_u^\top \theta_a^{(1)} \\
        \vdots \\
        X_u^\top \theta_a^{(j)} \\
        \bs{0}_{J-j}
    \end{bmatrix}
    = \sum_{i=1}^j (\theta_a^{(i)})^\top X_u \ell_z^{(j, i)}
    = (\theta_a)^\top \phi \big( X_u, \ell_z^{(j)} \big),
\end{align*}
where above we use $\ell_z^{(j, i)}$ to refer to the $i^{\TN{th}}$ element of the vector $\ell_z^{(j)}$.
The final equality uses $\theta_a = \big( \theta_a^{(1)}, \theta_a^{(2)}, \dots \theta_a^{(J)} \big)$ and the featurization defined in \eqref{eqn:phivector}.

An argument equivalent to that made at the end of the Proposition \ref{prop:posterior} proof can be used to show that
\begin{align*}
    \theta_a \mid \big( \MC{Y}_u^{(1:k)}, X_u, A_u = a, Z_a = z \big)
    \quad \overset{D}{=} \quad 
    \theta_a \mid \big( Y_u^{(1:k)}, X_u, A_u = a, Z_a = z \big).
\end{align*}

\section{Regret Bounds}
\label{app:regretBounds}

\subsection{Reduction to estimation error}

The next lemma establishes a sense in which Thompson sampling always `exploits' what it knows about arms' performance, attaining low regret once uncertainty is low. Developing appropriate measures of remaining uncertainty is key to the analysis. We adopt one from \cite{qin2023adaptive}, which looks at the  posterior variance of arms' reward distributions averaged according to the optimal action probabilities: $\sum_{a\in \MC{A}} p_{t,a}^* \sigma^2_{t, a}$. Critically, this measure allows an arm to be poorly estimated (large $\sigma^2_{t,a}$) if it has a very low chance of being optimal (low $p_{t,a}^*)$.

\begin{lemma}
\label{lem:regret-to-estimation} Define $\sigma_{t,a}^2 \triangleq {\rm Var}(\bar{R}_a \mid \HH_{t})$. Recall from \eqref{eqn:RbarDef} that $\bar{R}_a \triangleq \E \left[ R\big( Y_u(a) \big) \mid \theta_a \right]$. Under Assumptions \ref{assum:exchangeable}-\ref{assump:round},
\begin{multline*}
    \E \left[ \Delta_t(\pirnd) \right] \leq \sqrt{ 2 \log(|\MC{A}|) \E_{\pirnd} \bigg[ \sum_{a\in \MC{A}} p_{t,a}^* \sigma^2_{t, a} \bigg] } \\
    + \underbrace{ \epsilonrnd |\MC{A}| \max_{a \in \MC{A}} \E_{\pirnd} \left[ R \big(Y_u(A^*)\big) - R\big( Y_u(a) \big) \right] }_{= O(\epsilonrnd)}
\end{multline*}
where $p_{t,a}^* = \PP( A^* = a \mid \HH_{t} )$.
\end{lemma}

\begin{proof}
Note by the definition of $\Delta_t(\pi)$ from \eqref{eqn:DeltaDef}, for any policy $\pi$,
\begin{align*}
    \E \left[ \Delta_t(\pi) \mid \HH_{t} \right] 
    &= \E \bigg[ \E_{\pi} \bigg[ \frac{1}{|\Uc_t|} \sum_{u \in \Uc_t} \bigg\{ R \big(Y_u(A^*)\big)  - R\big( Y_u(A_u) \big) \bigg\} ~ \bigg| ~ \{ \theta_a \}_{a \in \MC{A}}, \HH_{t} \bigg] \, \bigg| \, \HH_{t} \bigg] \\
    &\underbrace{=}_{(a)} \E_{\pi} \bigg[ \frac{1}{|\Uc_t|} \sum_{u \in \Uc_t} \bigg\{ R \big(Y_u(A^*)\big)  - R\big( Y_u(A_u) \big) \bigg\} \, \bigg| \, \HH_{t} \bigg] \\
    &= \frac{1}{|\Uc_t|} \sum_{u \in \Uc_t} \sum_{a \in \MC{A}} \prnd_{t,a} \E \left[ R \big(Y_u(A^*)\big) - R\big( Y_u(a) \big) \mid \HH_{t} \right] \\
    &\underbrace{=}_{(b)} \sum_{a \in \MC{A}} \prnd_{t,a} \E \left[ R \big(Y_u(A^*)\big) - R\big( Y_u(a) \big) \mid \HH_{t} \right]  ~~~ \TN{for~any}~ u \in \Uc_t
\end{align*}
Above equality (a) holds by law of iterated expectations by averaging over the draw of latent item features $\{ \theta_a \}_{a \in \MC{A}}$. Equality (b) holds because of exchangeability Assumption \ref{assum:exchangeable}.

Below, we consider any $u \in \Uc_t$. By the result above, we have that
\begin{align*}
    &\E \left[ \Delta_t(\pirnd) \mid \HH_{t} \right]
    = \sum_{a \in \MC{A}} \prnd_{t,a} \E \left[ R \big(Y_u(A^*)\big) - R\big( Y_u(a) \big) \mid \HH_{t} \right] \\
    &= \sum_{a \in \MC{A}} \p_{t,a} \E \left[ R \big(Y_u(A^*)\big) - R\big( Y_u(a) \big) \mid \HH_{t} \right] 
    + \sum_{a \in \MC{A}} \left( \prnd_{t,a} - \p_{t,a} \right) \E \left[ R \big(Y_u(A^*)\big) - R\big( Y_u(a) \big) \mid \HH_{t} \right] \\
    &\leq \sum_{a \in \MC{A}} \p_{t,a} \E \left[ R \big(Y_u(A^*)\big) - R\big( Y_u(a) \big) \mid \HH_{t} \right]
    + \sum_{a \in \MC{A}} \left| \prnd_{t,a} - \p_{t,a} \right| \E \left[ R \big(Y_u(A^*)\big) - R\big( Y_u(a) \big) \mid \HH_{t} \right]
\end{align*}
Above recall that $\prnd_{t,a}$ is the number of users allocated to action $a$ in round $t$ according to $\pirnd$. $\p_{t,a}$ is the probability action $a$ is selected at batch $t$ under Thompson sampling $\piPS$. The final inequality above holds because the regret $\E \left[ R \big(Y_u(A^*)\big) - R\big( Y_u(a) \big) \mid \HH_{t} \right]$ is non-negative.

We take an expectation of the above expression to get the following inequality:
\begin{multline*}
    \E \left[ \Delta_t(\pirnd) \right] \leq  \underbrace{ \sum_{a \in \MC{A}} \E_{\pirnd} \left[ \p_{t,a} \E \left[ R \big(Y_u(A^*)\big) - R\big( Y_u(a) \big) \mid \HH_{t} \right] \right] }_{(i)} \\
    + \underbrace{ \sum_{a \in \MC{A}} \E_{\pirnd} \left[ \left| \prnd_{t,a} - \p_{t,a} \right| \E \left[ R \big(Y_u(A^*)\big) - R\big( Y_u(a) \big) \mid \HH_{t} \right] \right] }_{(ii)}
\end{multline*}
We will bound each of the two terms above separately.

\paragraph{Bounding term (i)} 
Below we consider any $u \in \Uc_t$. First, note the following:
\begin{align*}
    \sum_{a \in \MC{A}} \p_{t,a} \E \left[ R \big(Y_u(A^*)\big) - R\big( Y_u(a) \big) \mid \HH_{t} \right] 
    &= \E \left[ R \big(Y_u(A^*)\big) \mid \HH_{t} \right] - \sum_{a \in \MC{A}} \p_{t,a} \E \left[ R\big( Y_u(a) \big) \mid \HH_{t} \right] \\
    &= \E \left[ R \big(Y_u(A^*)\big) \mid \HH_{t} \right] - \sum_{a \in \MC{A}} p_{t,a}^* \E \left[ R\big( Y_u(a) \big) \mid \HH_{t} \right]
\end{align*}
The last equality above holds because $p_{t,a}^* = \PP( A^* = a \mid \HH_{t} ) = \PP(A_u = a \mid \HH_{t}) = \p_{t,a}$. Next, using the linearity of the function $R$ (Assumption \ref{assum:gaussian}),
\begin{align*}
    = \E \left[ R \big(Y_u(A^*)\big) \mid \HH_{t} \right] - \sum_{a \in \MC{A}} p_{t,a}^* R\big( \mu_{t,a} \big) 
    &= \E \left[ R \big(Y_u(A^*)\big) \mid \HH_{t} \right] - \E \left[ R\big( \mu_{t,A^*} \big) \mid \HH_{t} \right] \\
    &= \E \left[ R \big(Y_u(A^*)\big) - R\big( \mu_{t,A^*} \big) \mid \HH_{t} \right] \\
    &= \E \left[ R \big(\theta_{A^*}\big) - R\big( \mu_{t,A^*} \big) \mid \HH_{t} \right]
\end{align*}
The final equality above holds by the linearity of function $R$ and the law of iterated expectations.

By taking the expectation on both sides of the expression above, we get that:
\begin{align*}
    &\sum_{a \in \MC{A}} \E_{\pirnd} \left[ \p_{t,a} \E \left[ R \big(Y_u(A^*)\big) - R\big( Y_u(a) \big) \mid \HH_{t} \right] \right] 
    = \E_{\pirnd} \left[ \E \left[ R \big(Y_u(A^*)\big) - R\big( \mu_{t,A^*} \big) \mid \HH_{t} \right] \right] \\
    &\leq \sqrt{ \E_{\pirnd} \left[ \sigma_{t,A^*}^2 \right] 2 \log(|\MC{A}|) } 
    = \sqrt{ \E_{\pirnd} \bigg[ \sum_{a \in \MC{A}} p_{t,a}^* \sigma_{t,a}^2 \bigg] 2 \log(|\MC{A}|) }     
\end{align*}
The last inequality above holds by Proposition 1 of \cite{qin2023adaptive}. See the discussion of Proposition 1 of \cite{qin2023adaptive} for more commentary. 

\paragraph{Bounding term (ii)} 
Consider any $u \in \Uc_t$.
\begin{align*}
    &\sum_{a \in \MC{A}} \E_{\pirnd} \left[ \left| \prnd_{t,a} - \p_{t,a} \right| \E \left[ R \big(Y_u(A^*)\big) - R\big( Y_u(a) \big) \mid \HH_{t} \right] \right] \\
    &\underbrace{\leq}_{(d)} \epsilonrnd \sum_{a \in \MC{A}} \E_{\pirnd} \bigg[ \E \left[ R \big(Y_u(A^*)\big) - R\big( Y_u(a) \big) \mid \HH_{t} \right] \bigg] \\
    &\leq \epsilonrnd |\MC{A}| \max_{a \in \MC{A}} \E_{\pirnd} \left[ R \big(Y_u(A^*)\big) - R\big( Y_u(a) \big) \right] 
\end{align*}
Above inequality (d) holds by rounding Assumption \ref{assump:round}. 
\end{proof}

\subsection{Helpful Lemmas}

By Lemma \ref{lem:regret-to-estimation}, in order to bound regret, it is enough to bound $\E_{\pirnd} \big[ \sum_{a\in \MC{A}} p_{t,a}^* \sigma^2_{t, a} \big]$. Note,
\begin{align*}
    \E_{\pirnd} \bigg[ \sum_{a\in \MC{A}} p_{t,a}^* \sigma^2_{t, a} \bigg] 
    = \E_{\pirnd}\bigg[ \sum_{a \in \MC{A}} \ind_{A^*=a} \sigma_{t,a}^2 \bigg] 
    = \E_{\pirnd}\bigg[ \sum_{a \in \MC{A}} \ind_{A^*=a}r_1^\top {\rm Cov}\left(\theta_a \mid \HH_{t} \right) r_1 \bigg].
\end{align*}
Furthermore, note that 
\begin{align*}
    {\rm Cov}\left(\theta_a \mid \HH_{t} \right) 
    = \bigg\{ \Sigma_{1,Z_a}^{-1} + \sum_{u \in \Uc_{<t}} \ind_{A_u=a} \sum_{j=1}^{J} \ind\{ t > \tau_u + d_j \} \ell_{Z_a}^{(j)} (\ell_{Z_a}^{(j)})^\top \bigg\}^{-1} \\
    = \bigg\{ \Sigma_{1,Z_a}^{-1} + m \sum_{t'=1}^{t-1} \prnd_{t',a} \sum_{j=1}^{J} \ind\{ t > t' + d_j \} \ell_{Z_a}^{(j)} (\ell_{Z_a}^{(j)})^\top \bigg\}^{-1}
\end{align*}
The final equality above holds by re-indexing because $\prnd_{t',a} = \frac{1}{m} \sum_{u \in \Uc_t} \ind(A_u = a)$ and because if $u \in \Uc_{t'}$ then $\tau_u = t'$.

The following result (Lemma \ref{lemma:qin} below) is used to bound ${\rm Cov}\left(\theta_a \mid \HH_{t} \right)$. This bound will use $C_{t,a}^{\TN{Full}}$, where
\begin{align}
    C_{t,a}^{\TN{Full}} &\triangleq {\rm Cov}\left(\theta_a \mid Z_a, \{ \tilde{Y}_{u,t}(a) \}_{u \in \Uc_{< t} } \right) 
    = \bigg\{ \Sigma_{1,Z_a}^{-1} + m \sum_{t'=1}^{t-1} \sum_{j=1}^{J} \ind\{ t > t' + d_j \} \ell_{Z_a}^{(j)} (\ell_{Z_a}^{(j)})^\top \bigg\}^{-1}.
    \label{eqn:Cfulldef}
\end{align}
We can interpret $C_{t,a}^{\TN{Full}}$ as the covariance of $\theta_a$ given the data up to time $t$ in a “full feedback'' setting, and interpret ${\rm Cov}\left(\theta_a \mid \HH_{t} \right)$ as the covariance of $\theta_a$ given the data up to time $t$ in a “bandit feedback'' setting. Specifically, the difference between ${\rm Cov}\left(\theta_a \mid \HH_{t} \right)$ and $C_{t,a}^{\TN{Full}}$ is that ${\rm Cov}\left(\theta_a \mid \HH_{t} \right)$ only conditions on the outcomes $\big\{ \tilde{Y}_{u,t}(a) : A_u = a \big\}_{u \in \Uc_{< t}}$, according to $\HH_{t}$. In contrast, $C_{t,a}^{\TN{Full}}$ conditions on all of $\big\{ \tilde{Y}_{u,t}(a) \big\}_{u \in \Uc_{< t}}$.

\begin{lemma}[Lemma 8 of \cite{qin2023adaptive}]
    \label{lemma:qin}
    Under Assumptions \ref{assum:exchangeable} and \ref{assum:gaussian}, for any $t$ and $a \in \MC{A}$, with probability $1$,
    \begin{equation}
        \label{eq:ipwQin}
        {\rm Cov}\left(\theta_a \mid \HH_{t} \right) \preceq C^{\rm Full}_{t,a} \bigg( \Sigma_{1,Z_a}^{-1} + m \sum_{t'=1}^{t-1} \frac{\sum_{j=1}^{J} \ind\{ t > t' + d_j \} \ell_{Z_a}^{(j)} (\ell_{Z_a}^{(j)})^\top}{\prnd_{t',a}} \bigg) C^{\rm Full}_{t,a}.
    \end{equation}
\end{lemma}

The upper bound in Lemma \ref{lemma:qin} above relates ${\rm Cov}\left(\theta_a \mid \HH_{t} \right)$ to an expression that involves $C_{t,a}^{\TN{Full}}$ on the right-hand-side of \eqref{eq:ipwQin}. The Lemma follows from an identical proof to the argument in \citet[Lemma 8]{qin2023adaptive}.

In general, the inverse propensity weights $1/\prnd_{t',a}$ in \eqref{eq:ipwQin} can be extremely large. Thankfully, by Lemma \ref{lem:regret-to-estimation} the term we need to bound, $\E_{\pirnd} \big[ \sigma_{t,A^*}^2 \big] = \E_{\pirnd} \big[ \sum_{a\in \MC{A}} p_{t,a}^* \sigma^2_{t, a} \big]$, depends on only the posterior variance \emph{evaluated at the optimal arm} $A^*$. The next lemma controls the inverse propensity assigns to the optimal arm, $1/\prnd_{t',A^*}$.

\begin{lemma}
    Under Assumptions \ref{assum:exchangeable}, \ref{assum:gaussian}, and \ref{assump:round}, for any $t \in [1 \colon T]$,
    \begin{align*}
        \E_{\pirnd} \left[ \frac{\ind(A^*=a)}{ \prnd_{t,a} } \,\bigg|\, \HH_{t} \right] \leq 1 + 2 \epsilonrnd.
    \end{align*}
    \label{lemma:ipw}
\end{lemma}
\begin{proof}
    \begin{align*}
        &\E_{\pirnd} \left[ \frac{\ind(A^*=a)}{ \prnd_{t,a} } \,\bigg|\, \HH_{t} \right]
        = \E_{\pirnd} \left[ \frac{\ind(A^*=a)}{ p_{t,a}^* } \frac{ p_{t,a}^* }{ \prnd_{t,a} } \,\bigg|\, \HH_{t} \right] \\
        &\underbrace{=}_{(a)} \E_{\pirnd} \left[ \frac{\ind(A^*=a)}{ p_{t,a}^* } \frac{ \p_{t,a} }{ \prnd_{t,a} } \,\bigg|\, \HH_{t} \right] 
        \underbrace{\leq}_{(b)} \frac{1}{1 - \epsilonrnd} \E_{\pirnd} \left[ \frac{\ind(A^*=a)}{ p_{t,a}^* } \,\bigg|\, \HH_{t} \right] \\
        &= \frac{1}{1 - \epsilonrnd} \cdot 1
        \underbrace{\leq}_{(c)} 1 + 2 \epsilonrnd
    \end{align*}
    \begin{itemize}[leftmargin=*]
        \item Above, equality (a) holds because $p_{t,a}^* = \PP( A^* = a \mid \HH_{t} ) = \PP(A_u = a \mid \HH_{t}) = \p_{t,a}$.
        \item Inequality (b) holds by rounding Assumption \ref{assump:round}.
        \item Inequality (c) holds since $\epsilonrnd \leq 0.5$ by Assumption \ref{assump:round} and since by Taylor Series expansion for any $\epsilon$ with $0 < \epsilon < 1$, \quad
        $\frac{1}{1-\epsilon} = \sum_{k=0}^\infty \epsilon^k = 1 + \epsilon + \sum_{k=2}^\infty \epsilon^k \leq 1 + 2\epsilon$.
    \end{itemize}    
\end{proof}

\begin{lemma}[Simplifying Value of Progressive Feedback Term]
    \label{lemma:VoPF}
    Under Assumptions \ref{assum:exchangeable} and \ref{assum:gaussian},
    \begin{align}
        \label{eqn:simplifyingVoPF}
        \TN{VoPF}(t,z) = \frac{1}{2} \log \frac{ \Var \big( \bar{R}_a \mid Z_a=z, \{ Y_u(a) \}_{u \in \Uc_{< (t-d_{\rm max})} } \big) }{ \Var \big( \bar{R}_a \mid Z_a=z, \{ \tilde{Y}_{u,t}(a) \}_{u \in \Uc_{< t} } \big)  }.
    \end{align}
\end{lemma}

\begin{proof}
Note, by definition of VoPF from \eqref{eqn:priceofcensoring},
\begin{align*}
    &{\rm VoPF}(t, z)
    = I \left( \bar{R}_a ; \, \big\{ \tilde{Y}_{u,t}(a) \big\}_{u \in \Uc_{< t} } \mid Z_a = z, \big\{ Y_{u}(a) \big\}_{ u \in \Uc_{< (t-d_{\rm max})} } \right) \\
    &\underbrace{=}_{(a)} H\left( \bar{R}_a \mid Z_a = z, \big\{ Y_{u}(a) \big\}_{ u \in \Uc_{< (t-d_{\rm max})} } \right) - H \left( \bar{R}_a \mid Z_a = z, \big\{ \tilde{Y}_{u,t}(a) \big\}_{u \in \Uc_{< t} }  \right) 
\end{align*}
\begin{align*}
    &\underbrace{=}_{(b)} \E \left[ \frac{1}{2} \log \left\{ 2 \pi e {\rm Var} \left( \bar{R}_a \mid Z_a = z, \big\{ Y_{u}(a) \big\}_{u \in \Uc_{< (t-d_{\rm max})}} \right) \right\} \right] \\
    &\quad \quad \quad \quad \quad \quad \quad \quad \quad \quad \quad \quad \quad \quad \quad - \E \left[ \frac{1}{2} \log \left[2 \pi e {\rm Var} \left( \bar{R}_a \mid Z_a = z, \big\{ \tilde{Y}_{u,t-1}(a) \big\}_{u \in \Uc_{< t} } \right) \right] \right] \\
    &= \frac{1}{2} \E \left[ \log \frac{ \Var \big( \bar{R}_a \mid Z_a = z, \{ Y_{u,t}(a) \}_{u \in \Uc_{< (t-d_{\rm max})} } \big) }{ \Var \left( \bar{R}_a \mid Z_a = z, \{ \tilde{Y}_{u,t}(a) \}_{u \in \Uc_{< t} } \right)  } \,\bigg|\, Z_a = z \right] \\
    &\underbrace{=}_{(c)} \frac{1}{2} \log \frac{ \Var \big( \bar{R}_a \mid Z_a = z, \{ Y_{u,t}(a) \}_{u \in \Uc_{< (t-d_{\rm max})} } \big) }{ \Var \left( \bar{R}_a \mid Z_a = z, \{ \tilde{Y}_{u,t}(a) \}_{u \in \Uc_{< t} } \right)  }
\end{align*}
\begin{itemize}[leftmargin=*]
    \item Equality (a): This holds by standard results in information theory. Note as is standard in information theory, the mutual information $I \left( \bar{R}_a ; \, \big\{ \tilde{Y}_{u,t}(a) \big\}_{u \in \Uc_{< t} } \mid Z_a = z, \big\{ Y_{u}(a) \big\}_{ u \in \Uc_{< (t-d_{\rm max})} } \right)$ is a constant as it marginalizes over $\big\{ Y_{u}(a) \big\}_{ u \in \Uc_{< (t-d_{\rm max})}}$. \\ 
    Similarly, $H\left( \bar{R}_a \mid Z_a = z, \big\{ Y_{u}(a) \big\}_{ u \in \Uc_{< (t-d_{\rm max})} } \right)$ denotes the conditional entropy and is marginalizes over $\big\{ Y_{u}(a) \big\}_{ u \in \Uc_{< (t-d_{\rm max})}}$.
    \item Equality (b): This holds by the formula for entropy of a Gaussian random variable.
    \item Equality (c): Under our Gaussian assumption, the variance terms $\Var \big( \bar{R}_a \mid Z_a=z, \{ Y_{u,t}(a) \}_{u \in \Uc_{< (t-d_{\rm max})} } \big)$ and $\Var \left( \bar{R}_a \mid Z_a = z, \{ \tilde{Y}_{u,t}(a) \}_{u \in \Uc_{< t} } \right)$ are constants.
\end{itemize}
\end{proof}

\begin{lemma}[Simplifying Mean Reward Variance Given Progressive Feedback]
Under Assumptions \ref{assum:exchangeable}-\ref{assum:gaussian},
\begin{align*}
    \E \left[ \Var \left( \bar{R}_a \mid Z_a, \{ \tilde{Y}_{u,t}(a) \}_{u \in \Uc_{< t} } \right) \right]
    = \E \left[ \frac{ \sigma_{R}^{2}(Z_a) \cdot \exp \big( -2 \cdot {\rm VoPF}(t, Z_a) \big) }{ \sigma_{R}^2(Z_a) \cdot (r_1^\top \Sigma_{Z_a} r_1)^{-1} + |\Uc_{< (t-d_{\rm max})}| } \right]
\end{align*}
\label{lemma:simplifyingCov}
\end{lemma}

\begin{proof}
For any $a \in \MC{A}$,
\begin{align*}
    &\E \left[ \Var \left( \bar{R}_a \mid Z_a, \{ \tilde{Y}_{u,t}(a) \}_{u \in \Uc_{< t} } \right) \right]  \\
    &= \E \bigg[ \Var \left( \bar{R}_a \mid Z_a, \{ Y_{u,t}(a) \}_{u \in \Uc_{< (t-d_{\rm max})} } \right) \frac{ \Var \left( \bar{R}_a \mid Z_a, \{ \tilde{Y}_{u,t}(a) \}_{u \in \Uc_{< t} } \right) }{ \Var \big( \bar{R}_a \mid Z_a, \{ Y_{u,t}(a) \}_{u \in \Uc_{< (t-d_{\rm max})} } \big) } \bigg] \\
    &\underbrace{=}_{(i)} \E \bigg[ \frac{ \sigma_{R}^{2}(Z_a) }{ \sigma_{R}^2(Z_a) \cdot (r_1^\top \Sigma_{Z_a} r_1)^{-1} + |\Uc_{< (t-d_{\rm max})}| } \cdot \frac{ \Var \left( \bar{R}_a \mid Z_a, \{ \tilde{Y}_{u,t}(a) \}_{u \in \Uc_{< t} } \right) }{ \Var \big( \bar{R}_a \mid Z_a, \{ Y_{u,t}(a) \}_{u \in \Uc_{< (t-d_{\rm max})} } \big) } \bigg] \\
    &\underbrace{=}_{(ii)} \E \left[ \frac{ \sigma_{R}^{2}(Z_a) \cdot \exp \big( -2 \cdot {\rm VoPF}(t, Z_a) \big) }{ \sigma_{R}^2(Z_a) \cdot (r_1^\top \Sigma_{Z_a} r_1)^{-1} + |\Uc_{< (t-d_{\rm max})}| } \right]
\end{align*}
Equality (i) holds since with probability $1$,
    \begin{align}
        \label{eqn:inequalityg}
        \Var \left( \bar{R}_a \mid Z_a, \{ Y_{u,t}(a) \}_{u \in \Uc_{< (t-d_{\rm max})} } \right)
        = \frac{ \sigma_{R}^2(Z_a) }{ \sigma_{R}^2(Z_a) \cdot (r_1^\top \Sigma_{1,Z_a} r_1)^{-1} + |\Uc_{< (t-d_{\rm max})}| }.
    \end{align}
    The equality above holds by applying the formula for updating a scalar Gaussian random variable $\bar{R}_a$ based on i.i.d noisy measurements $\{ R(Y_u) \}_{u\in \Uc_{< t} }$; recall that we use the notation $\sigma^2_{R}(z) = \Var \big( R(Y_u(a)) \mid \bar{R}_a, Z_a = z \big)$.
Equality (ii) above holds by Lemma \ref{lemma:VoPF}.
\end{proof}

\subsection{Completing the proof of Theorem \ref{thm:main}}
\label{app:proofTheoremMain}

\begin{customthm}{\ref{thm:main}}[Regret with Progressive Feedback]
Let Assumptions \ref{assum:exchangeable}-\ref{assump:round} hold, and let $\Sigma_{1,z}$ and $V_z$ be invertible for all $z \in \MC{Z}$. Then under Impatient Bandit Thompson Sampling algorithm $\pirnd$ (Algorithm \ref{alg:banditrnd}), for any $t>1$,
\begin{align*}
    \E \left[ \Delta_t(\pirnd) \right]  
    &\leq \sqrt{ \E \left[ \frac{ \exp \big( -2 \cdot {\rm VoPF}(t, Z_a) \big) \cdot \sigma^2_{R}(Z_a) }{ \sigma_{R}^2(Z_a) \cdot (r_1^\top \Sigma_{Z_a} r_1)^{-1} + |\Uc_{< (t-d_{\rm max})}|} \right] } \times \sqrt{ 2 |\MC{A}| \log(|\MC{A}|) } + O(\epsilon_{\rm rnd}).
\end{align*}
\end{customthm}

\begin{proof}
Recall that, by Lemma \ref{lem:regret-to-estimation}, that
\begin{align*}
    \E \left[ \Delta_t(\pirnd) \right] \leq \sqrt{ 2 \log(|\MC{A}|) \E_{\pirnd} \bigg[ \sum_{a\in \MC{A}} p_{t,a}^* \sigma^2_{t, a} \bigg] }
    + O(\epsilonrnd)
\end{align*}
where $p_{t,a}^* = \PP( A^* = a \mid \HH_{t} )$. By the above result, it is sufficient for the Theorem to show
\begin{align}
    \label{eqn:lastPartTheorem}
    \E_{\pirnd} \bigg[ \sum_{a\in \MC{A}} p_{t,a}^* \sigma^2_{t, a} \bigg]
    \leq (1+2\epsilon_{\rm rnd}) | \MC{A} | \cdot \E \left[ \frac{ \sigma_{R}^{2}(Z_a) \cdot \exp \big( -2 \cdot {\rm VoPF}(t, Z_a) \big) }{ \sigma_{R}^2(Z_a) \cdot (r_1^\top \Sigma_{Z_a} r_1)^{-1} + |\Uc_{< (t-d_{\rm max})}| } \right].
\end{align}
The remainder of the proof will be showing \eqref{eqn:lastPartTheorem} holds. Note that
\begin{align*}
    &\E_{\pirnd} \bigg[ \sum_{a\in \MC{A}} p_{t,a}^* \sigma^2_{t, a} \bigg] 
    = \E_{\pirnd}\bigg[ \sum_{a \in \MC{A}} \ind(A^*=a) \sigma_{t,a}^2 \bigg] \\
    &= \E_{\pirnd}\bigg[ \sum_{a \in \MC{A}} \ind(A^*=a) r_1^\top {\rm Cov}\left(\theta_a \mid \HH_{t} \right) r_1 \bigg] 
    = \sum_{a \in \MC{A}} r_1^\top \E_{\pirnd}\left[ \ind(A^*=a) {\rm Cov}\left(\theta_a \mid \HH_{t} \right) \right] r_1 \\
    &\leq \sum_{a \in \MC{A}} r_1^\top \E_{\pirnd}\bigg[ \ind(A^*=a) C^{\rm Full}_{t,a} \bigg( \Sigma_{1,Z_a}^{-1} + m \sum_{t'=1}^{t-1} \frac{\sum_{j=1}^{J} \ind\{ t > t' + d_j \} \ell_{Z_a}^{(j)} (\ell_{Z_a}^{(j)})^\top}{\prnd_{t',a}} \bigg) C^{\rm Full}_{t,a} \bigg] r_1
\end{align*}
The last inequality above holds by Lemma \ref{lemma:qin}. Additionally, by \eqref{eqn:Cfulldef}, $C_{t,a}^{\rm Full}$ is non-random given $\{ Z_a \}_{a \in \MC{A}}$. Thus, by the law of iterated expectations, the above equals the following:
\begin{align*}
    = \sum_{a \in \MC{A}} r_1^\top \E \bigg[ C^{\rm Full}_{t,a} \E_{\pirnd}\bigg[ \ind(A^*=a) \bigg( \Sigma_{1,Z_a}^{-1} + m \sum_{t'=1}^{t-1} \frac{\sum_{j=1}^{J} \ind\{ t > t' + d_j \} \ell_{Z_a}^{(j)} (\ell_{Z_a}^{(j)})^\top}{\prnd_{t',a}} \bigg) \bigg| \{ Z_a \}_{a \in \MC{A}} \bigg] C^{\rm Full}_{t,a} \bigg] r_1
\end{align*}
By moving terms that are constant given $\{ Z_a \}_{a \in \MC{A}}$ out of the expectation, we get that
\begin{align*}
    = \sum_{a \in \MC{A}} r_1^\top \E \bigg[ C^{\rm Full}_{t,a} \bigg( \Sigma_{1,Z_a}^{-1} + m \sum_{t'=1}^{t-1} \E_{\pirnd} \bigg[ \frac{ \ind(A^*=a) }{\prnd_{t',a}} \bigg| \{ Z_a \}_{a \in \MC{A}} \bigg] \sum_{j=1}^{J} \ind\{ t > t' + d_j \} \ell_{Z_a}^{(j)} (\ell_{Z_a}^{(j)})^\top \bigg) C^{\rm Full}_{t,a} \bigg] r_1
\end{align*}
By Lemma \ref{lemma:ipw} (recall that $\{ Z_a \}_{a \in \MC{A}}$ is part of the history $\HH_{t}$),
\begin{align*}
    \leq (1 + 2 \epsilonrnd) \sum_{a \in \MC{A}} r_1^\top \E \bigg[ C^{\rm Full}_{t,a} \bigg( \Sigma_{1,Z_a}^{-1} + m \sum_{t'=1}^{t-1} \sum_{j=1}^{J} \ind\{ t > t' + d_j \} \ell_{Z_a}^{(j)} (\ell_{Z_a}^{(j)})^\top \bigg) C^{\rm Full}_{t,a} \bigg] r_1
\end{align*}
By the definition of $C_{t,a}^{\rm Full}$ from \eqref{eqn:Cfulldef},
\begin{align*}
    = (1 + 2 \epsilonrnd) \sum_{a \in \MC{A}} r_1^\top \E \left[ C^{\rm Full}_{t,a} \right] r_1 
    &= (1+2\epsilon_{\rm rnd}) \sum_{a \in \MC{A}} \E \left[ r_1^\top {\rm Cov}\left(\theta_a \mid Z_a, \{ \tilde{Y}_{u,t}(a) \}_{u \in \Uc_{< t} } \right) r_1 \right] \\
    &\underbrace{=}_{(i)} (1+2\epsilon_{\rm rnd}) | \MC{A} | 
    \E \left[ \Var \left( \bar{R}_a \mid Z_a, \{ \tilde{Y}_{u,t}(a) \}_{u \in \Uc_{< t} } \right) \right] \\
    &\underbrace{=}_{(ii)} (1+2\epsilon_{\rm rnd}) | \MC{A} | 
    \E \left[ \frac{ \sigma_{R}^{2}(Z_a) \cdot \exp \big( -2 \cdot {\rm VoPF}(t, Z_a) \big) }{ \sigma_{R}^2(Z_a) \cdot (r_1^\top \Sigma_{Z_a} r_1)^{-1} + |\Uc_{< (t-d_{\rm max})}| } \right].
\end{align*}
Equality (i) above holds for any action $a \in \MC{A}$, since by Assumption \ref{assum:items}, $a \in \MC{A}$ are i.i.d. from the same distribution. Finally, equality (ii) holds by Lemma \ref{lemma:simplifyingCov}.
\end{proof}

\subsection{Proof of Corollary \ref{corr:mainNoZ}}

\begin{customcor}{\ref{corr:mainNoZ}}[Regret of Impatient Bandit Algorithm without Action Features $Z_a$]
Let Assumptions \ref{assum:exchangeable}-\ref{assump:round} hold, and let $\Sigma_{1}$ and $V$ be invertible. Then under the Impatient Bandit Thompson Sampling algorithm $\pirnd$ (Algorithm \ref{alg:banditrnd}), for any $t>1$,
\begin{align*}
    \E \left[ \Delta_t(\pirnd) \right]  \leq
    \underbrace{%
    \vphantom{ \sqrt{ \frac{ \lvert \MC{A} \rvert}{\sigma_{R}^2} } }
    \exp \big[ - {\rm VoPF}(t) \big]
}_{\substack{\TN{Decrease from} 
 \\ \TN{Progressive Feedback}}}
\cdot \underbrace{%
    \sigma_{R} \sqrt{ \frac{ 2 \lvert \MC{A} \rvert \log \lvert \MC{A} \rvert}{\sigma_{R}^2 (r_1^\top \Sigma_1 r_1)^{-1} + \lvert \Uc_{<(t-d_{\rm max})}\rvert} }
}_{\TN{Regret under Delayed Rewards}}
+ \underbrace{%
    \vphantom{ \sqrt{ \frac{ \lvert \MC{A} \rvert}{\sigma_{R}^2} } }
    O(\epsilon_{\rm rnd})
}_{\TN{Rounding error}}.
\end{align*}
\end{customcor}

\begin{proof}
In the case in which there are no action features $Z_a$, recall from \eqref{eqn:priceofcensoringNoZ} that ${\rm VoPF}(t) = I \big( \bar{R}_a ; ~ \big\{ \tilde{Y}_{u,t}(a) : u \in \Uc_{< t} \big\} \mid \big\{ Y_{u}(a) : u \in \Uc_{< (t-d_{\rm max})} \big\} \big)$.
By Theorem \ref{thm:main}, we have that
\begin{align*}
    \E \left[ \Delta_t(\pirnd) \right] 
    &\leq \sqrt{ \E \left[ \exp \big( -2 \cdot {\rm VoPF}(t) \big) \cdot \sigma^2_{R}(Z_a) \right] } \times \sqrt{ \frac{ 2 |\MC{A}| \log(|\MC{A}|)}{| \Uc_{<(t-d_{\rm max})}|} } + O(\epsilon_{\rm rnd}) \\
    &= \sqrt{ \exp \big( -2 \cdot {\rm VoPF}(t) \big) \cdot \sigma^2_{R} } \times \sqrt{ \frac{ 2 |\MC{A}| \log(|\MC{A}|)}{| \Uc_{<(t-d_{\rm max})}|} } + O(\epsilon_{\rm rnd}).
\end{align*}
Note for last equality above we use the fact when there are no action features $Z_a$, $\sigma_R^2 = \sigma_R^2(Z_a)$ is a constant and so is $\TN{VoPF}(t)$. 
\end{proof}

\begin{customprop}{\ref{prop:extremes}}[Regret Under Extremes] 
Let the conditions of Corollary \ref{corr:mainNoZ} hold.

\noindent \under{\TN{(1) Completely Uninformative Intermediate Feedback (Delayed Rewards).}} In the case that the intermediate feedback is completely uninformative, i.e., ${\rm VoPF}\big( t \big) = 0$ for all $t$,
    \begin{multline*}
        \E \bigg[ \frac{1}{T} \sum_{t=1}^T \Delta_t(\pirnd) \bigg] \, \leq \,
        \frac{d_{\max}}{T} \sqrt{ 2 (r_1^\top \Sigma_1 r_1) \log (|\MC{A}|) } 
        + 3 \sigma_R \sqrt{ \frac{ 2 |\MC{A}| \log (|\MC{A}|)}{Tm} } 
        + O(\epsilon_{\rm rnd}).
    \end{multline*}

\noindent \under{\TN{(2) Perfect Intermediate Feedback (No Delay).}} If we have a perfect surrogate, i.e., \\
$\TN{Corr}\big(Y_u^{(1)}, R(Y_u) \big) = 1$ 
    and $d_1 = 0$,
    \begin{align*}
        \E \bigg[ \frac{1}{T} \sum_{t=1}^T \Delta_t(\pirnd) \bigg]
        \, \leq \,  3 \sigma_R \sqrt{ \frac{ 2 |\MC{A}| \log (|\MC{A}|)}{ Tm} } + O(\epsilonrnd).
    \end{align*}
\end{customprop}

\begin{proof}

\noindent \under{(1) Completely Uninformative Intermediate Feedback (Delayed Rewards).}
We first bound the regret under the first $d_{\max}$ batches. Let $\pi_{\rm uniform}$ denote the policy that randomly selects an action uniformly from $\MC{A}$ to play exclusively for the first $d_{\max}$ batches:
\begin{multline}
     \E \bigg[ \frac{1}{T} \sum_{t=1}^{d_{\max}} \Delta_t(\pirnd) \bigg]
     \leq \E \bigg[ \frac{1}{T} \sum_{t=1}^{d_{\max}} \Delta_t(\pi_{\rm uniform}) \bigg]
     \underbrace{\leq}_{(a)} \frac{d_{\max}}{T} \left( \E \left[ \argmax_{a \in \MC{A}} \big\{ r_1^\top \theta_a \big\} \right] - r_1^\top \mu_1 \right) \\
     = \frac{d_{\max}}{T} \E \left[ \argmax_{a \in \MC{A}} \big\{ r_1^\top (\theta_a-\mu_1) \big\} \right] 
     \underbrace{\leq}_{(b)} \frac{d_{\max}}{T} \sqrt{ 2 (r_1^\top \Sigma_1 r_1) \log (|\MC{A}|) }
     \label{eqn:initialBatches}
\end{multline}
Inequality (a) above uses that the reward function $R(Y_u(a)) = r_1^\top Y_u(a) + r_2$ is a linear function, which means $\E[ R(Y_u(a)) \mid \theta_a] = r_1^\top \theta_a$. Recall $\theta_a \sim N( \mu_1, \Sigma_1)$ according to Lemma \ref{lem:joint_normal}. Inequality (b) uses that $\E \left[ \argmax_{a \in \MC{A}} \big\{ r_1^\top (\theta_a-\mu_1) \big\} \right] = \E \left[ \argmax_{a \in \MC{A}} W_a \right]$ for $W_a \iidsim N( 0, r_1^\top \Sigma_1 r_1)$ over $a \in \MC{A}$ and basic inequality bounds on the expectation of the maxima of independent Gaussian random variables.

We now consider the subsequent batches (after the first $d_{\max}$ batches). By Corollary \ref{corr:mainNoZ},
\begin{multline}
    \E \bigg[ \frac{1}{T} \sum_{t=d_{\max}+1}^T \Delta_t(\pirnd) \bigg]  \leq 
    \sigma_{R} \sqrt{ 2 |\MC{A}| \log(|\MC{A}|) } \cdot \frac{1}{T} \sum_{t=d_{\max}+1}^T \sqrt{ \frac{ \exp \big( - 2 \cdot {\rm VoPF}\big( t \big) \big) }{ \sigma_{R}^2 \cdot (r_1^\top \Sigma_1 r_1)^{-1} + | \Uc_{<(t-d_{\rm max})}|} } \\
    + O(\epsilon_{\rm rnd}).
    \label{eqn:sumAcrossBatchesDmax}
\end{multline}
Note when ${\rm VoPF}\big( t \big) = 0$ for all $t$,
\begin{align*}
    \frac{1}{T} \sum_{t=d_{\rm max}+1}^T \sqrt{ \frac{ \exp \big( - 2 \cdot {\rm VoPF}\big( t \big) \big) }{\sigma_{R}^2 (r_1^\top \Sigma_1 r_1)^{-1} + |\Uc_{<t-d_{\rm max}}| }}
    &\leq \frac{1}{T} \sum_{t=d_{\rm max}+1}^T \frac{1}{\sqrt{ |\Uc_{<t-d_{\rm max}}| }} \\
    &\underbrace{=}_{(a)} \frac{1}{ T \sqrt{m} } \sum_{t=d_{\rm max}+1}^T \frac{1}{\sqrt{ t - d_{\rm max}}} = \frac{1}{ T \sqrt{m} } \sum_{t=1}^{T-d_{\rm max}} \frac{1}{\sqrt{ t}} \\
    &\underbrace{\leq}_{(b)} \frac{2 \sqrt{T-d_{\rm max}} + 1}{ T \sqrt{m}}
    = \frac{2}{\sqrt{Tm}} + \frac{1}{T \sqrt{m}}
    \leq \frac{3}{\sqrt{Tm}}
\end{align*}
For equality (a) above, we use that $|\Uc_{<t-d_{\rm max}}| = m \cdot \max(0, t-d_{\rm max})$. Equality (b) holds since $\frac{1}{\sqrt{t}}$ is decreasing in $t$, $\sum_{t=1}^T \frac{1}{\sqrt{T}} \leq \int_0^T \frac{1}{\sqrt{t}} dt = 2 \sqrt{T}$. By the above result, \eqref{eqn:initialBatches}, and \eqref{eqn:sumAcrossBatchesDmax}, we have shown  \eqref{eqn:regretPt1} holds. \\

\noindent \under{(2) Perfect Intermediate Feedback (No Delay).}
By Corollary \ref{corr:mainNoZ},
\begin{align*}
    &\E \bigg[ \frac{1}{T} \sum_{t=1}^T \Delta_t(\pirnd) \bigg] 
    \leq \sigma_{R} \sqrt{ 2 |\MC{A}| \log(|\MC{A}|) } \cdot \frac{1}{T} \sum_{t=1}^T \sqrt{ \frac{ \exp \big( - 2 \cdot {\rm VoPF}\big( t \big) \big) }{ \sigma_{R}^2 \cdot (r_1^\top \Sigma_1 r_1)^{-1} + | \Uc_{<(t-d_{\rm max})}|} } + O(\epsilon_{\rm rnd}) \\
    &\underbrace{=}_{(a)} \sigma_{R} \sqrt{ 2 |\MC{A}| \log(|\MC{A}|) } \cdot \frac{1}{T} \sum_{t=1}^T \sqrt{  \Var \big( \bar{R}_a \mid Z_a=z, \{ \tilde{Y}_{u,t}(a) \}_{u \in \Uc_{< t} } \big) }
    + O(\epsilon_{\rm rnd}) \\
    &\underbrace{=}_{(b)} \sigma_{R} \sqrt{ 2 |\MC{A}| \log(|\MC{A}|) } \cdot \frac{1}{T} \sum_{t=1}^T \sqrt{ \frac{1}{\sigma_{R}^2 \cdot (r_1^\top \Sigma_1 r_1)^{-1} + | \Uc_{<t}|} } + O(\epsilon_{\rm rnd}) \\
    &\underbrace{\leq}_{(c)} \sigma_{R} \sqrt{ 2 |\MC{A}| \log(|\MC{A}|) } \cdot \frac{1}{ T \sqrt{m} } \sum_{t=1}^T \sqrt{ \frac{ 1 }{ t } }
    + O(\epsilon_{\rm rnd}) \\
    &\underbrace{\leq}_{(d)} \sigma_{R} \sqrt{ 2 |\MC{A}| \log(|\MC{A}|) } \cdot \frac{2 \sqrt{T} + 1}{T \sqrt{m}} + O(\epsilon_{\rm rnd})
    \leq \sigma_{R} \sqrt{ 2 |\MC{A}| \log(|\MC{A}|) } \cdot \frac{3}{\sqrt{Tm}} + O(\epsilon_{\rm rnd})
\end{align*}
Above (a) holds since $\Var ( \bar{R}_a \mid Z_a=z, \{ Y_{u,t}(a) \}_{u \in \Uc_{< (t-d_{\rm max})} } ) = \{ (r_1^\top \Sigma_1 r_1)^{-1} + | \Uc_{<(t-d_{\rm max})}| \}^{-1}$ and that by Lemma \ref{lemma:VoPF}, 
    \begin{align*}
        \exp \big( - 2 \cdot {\rm VoPF}\big( t \big) \big) 
        = \frac{ \Var \big( \bar{R}_a \mid Z_a=z, \{ \tilde{Y}_{u,t}(a) \}_{u \in \Uc_{< t} } \big)  }{ \Var \big( \bar{R}_a \mid Z_a=z, \{ Y_{u,t}(a) \}_{u \in \Uc_{< (t-d_{\rm max})} } \big) }
    \end{align*}
Equality (b) uses the formula for the posterior variance of a Gaussian, as well as the fact that we have a perfect surrogate, i.e., that $\TN{Corr}(R_u, Y_u^{(1)}) = 1$.
For inequality (c) above, we use that $|\Uc_{\leq t}| = m \cdot t$.
Inequality (d) holds since $\frac{1}{\sqrt{t}}$ is decreasing in $t$, $\sum_{t=1}^T \frac{1}{\sqrt{T}} \leq \int_0^T \frac{1}{\sqrt{t}} dt = 2 \sqrt{T}$. Thus we have shown that \eqref{eqn:perfect} holds.
\end{proof}

\subsection{Derivation of Remark \ref{rem:interpretVoPF}}
\label{app:derivationRemark}

By Lemma \ref{lemma:VoPF}, under Assumptions \ref{assum:exchangeable} and \ref{assum:gaussian},
\begin{align}
    \label{eqn:simplifyingVoPFRemark}
    \TN{VoPF}(t) = \frac{1}{2} \log \frac{ \Var \big( \bar{R}_a \mid \{ Y_u(a) \}_{u \in \Uc_{< (t-d_{\rm max})} } \big) }{ \Var \big( \bar{R}_a \mid \{ \tilde{Y}_{u,t}(a) \}_{u \in \Uc_{< t} } \big)  }.
\end{align}
Recall that $\theta_a \sim N \left( \mu_1, \Sigma_1 \right)$ and $Y_u(a) \mid \theta_a \sim N \left( \theta_a, V \right)$, where
\begin{align*}
\Sigma_1 = \begin{bmatrix}1 & \rho \\ \rho & 1\end{bmatrix} \qquad \TN{and} \qquad
V = \sigma^2_R \begin{bmatrix}1 & \rho \\ \rho & 1\end{bmatrix}.
\end{align*}

\paragraph{Numerator of \eqref{eqn:simplifyingVoPFRemark}}
Note that the posterior distribution $\theta_a \mid \{ Y_{u,t}(a) \}_{u \in \Uc_{< (t-d_{\rm max})} }$ is multivariate Gaussian with the following posterior covariance matrix:
\begin{align*}
C \triangleq (\Sigma_1^{-1} + m(t-d_{\max})V^{-1})^{-1} = (1 + m(t-d_{\max}) \sigma^{-2}_R)^{-1} \Sigma_1.
\end{align*}
Above, recall we use $m$ to refer to the number of users per batch. Thus, the posterior variance of the expected reward $\bar{R}_a$ is 
\begin{align*}
\mathbf{Var}[\bar{R}_a \mid \{ Y_{u,t}(a) \}_{u \in \Uc_{< (t-d_{\rm max})} } ] = (1 + m(t-d_{\max}) \sigma^{-2}_R)^{-1}
\end{align*}

\paragraph{Denominator of \eqref{eqn:simplifyingVoPFRemark}}
Now note that the posterior distribution $\theta_a \mid \{ \tilde{Y}_{u,t}(a) \}_{u \in \Uc_{<t} }$ is multivariate Gaussian with the following posterior covariance matrix:
\begin{align*}
&C - ((1 + m(t-d_{\max}) \sigma^{-2}_R)^{-1} + \sigma^2_R)^{-1} \bm{c} \bm{c}^\top \\
&\qquad = (1 + n\sigma^{-2}_R)^{-1} \left(\begin{bmatrix}1 & \rho \\ \rho & 1\end{bmatrix}
    - (1 + \sigma^2_R + m(t-d_{\max}))^{-1}\begin{bmatrix}1 & \rho \\ \rho & \rho^2\end{bmatrix} \right) 
\end{align*}
where $\bm{c} = (1 + m(t-d_{\max})\sigma^2_R)^{-1} \begin{bmatrix} 1 & \rho \end{bmatrix}$ is the first row of $C$. As a result,
\begin{align*}
\mathbf{Var}[\bar{R}_a \mid \{ \tilde{Y}_{u,t}(a) \}_{u \in \Uc_{<t} }] 
&\qquad = (1 + m(t-d_{\max})\sigma^{-2}_R)^{-1}\left(1 - \frac{\rho^2}{\sigma^2_R + m(t-d_{\max}) + 1} \right) \\
&\qquad = (1 + m(t-d_{\max}) \sigma^{-2}_R)^{-1}\left(\frac{\sigma^2_r + m(t-d_{\max}) + 1 - \rho^2}{\sigma^2_R + m(t-d_{\max}) + 1} \right)
\end{align*}
Thus, we can rewrite \eqref{eqn:simplifyingVoPFRemark} as follows:
\begin{align*}
\TN{VoPF}(t) =
\frac{1}{2} \log\left( \frac{\sigma^2_R + m(t-d_{\max}) + 1}{\sigma^2_r + m(t-d_{\max}) + 1 - \rho^2} \right) \quad \forall t \geq d_{\max}.
\end{align*}

\section{Rounding Procedure}
\label{app:rounding}

\begin{algorithm}
    \caption{Iterative Decrement Rounding Algorithm}
    \begin{algorithmic}[1]
        \State \bo{Inputs:} Probabilities $\{\p_{t,a}\}_{a \in \MC{A}}$, Granularity $m$
        \For{$a \in \MC{A}$}
            \State $\prnd_{t,a} \gets \frac{\lceil m \p_{t,a} \rceil}{m}$
        \EndFor
        \While{$\sum_{a \in \MC{A}} \prnd_{t,a} > 1$}
            \State $\alpha \gets \argmax_{a \in \MC{A}} \frac{\prnd_{t,a} - 1/m}{\p_{t,a}}$
            \State $\prnd_{t,\alpha} \gets \prnd_{t,\alpha} - \frac{1}{m}$
        \EndWhile
        \State \bo{Output:} $\{ \prnd_{t,a} \}_{a \in \MC{A}}$
    \end{algorithmic}
    \label{alg:rounding-dec}
\end{algorithm}

\begin{proposition}[Iterative Decrement Rounding Procedure]
Rounding procedure Algorithm \ref{alg:rounding-dec} satisfies Assumption \ref{assump:round} holds as long as $|\MC{A}| \geq 2$ and $m \geq 2 |\MC{A}|$. Specifically, for any value of $m$, Assumption \ref{assump:round} will hold for $\epsilonrnd = \frac{|\MC{A}|}{m}$.
\end{proposition}

\begin{proof}
Let $q^{(0)}_a \triangleq \frac{\lceil m \p_{t,a} \rceil}{m}$ denote the initially rounded-up probabilities. Since $\p_{t,a} \leq q^{(0)}_a < \p_{t,a} + \frac{1}{m}$, we have that
\begin{align*}
    1 \leq \sum_{a \in \MC{A}} q^{(0)}_a < 1 + \frac{|\MC{A}|}{m}.
\end{align*}
Moreover, $\sum_{a \in \MC{A}} q^{(0)}_a$ lies on the grid $\{j/m : j \in \mathbb{Z}\}$. Therefore the excess mass $\sum_{a \in \MC{A}} q^{(0)}_a - 1$ is an integer multiple of $1/m$ and is strictly less than $|\MC{A}|/m$. Hence Algorithm \ref{alg:rounding-dec} performs at most $|\MC{A}|$ decrement steps, and after these steps the output $\prnd_t$ satisfies $\sum_{a \in \MC{A}} \prnd_{t,a} = 1$.

\paragraph{Absolute Error Bound.} For each $a$, $0 \leq q^{(0)}_a - \p_{t,a} < \frac{1}{m}$.
The algorithm only decreases the initial rounded up value $q^{(0)}_a$, and the total number of decrements is at most $|\MC{A}|$. Therefore no coordinate can move downward by more than $|\MC{A}|/m$, which implies
\begin{align*}
    \left| \prnd_{t,a} - \p_{t,a} \right|
    \leq \frac{|\MC{A}|}{m}.
\end{align*}

\paragraph{Relative Error Bound.} 
If $q^{(0)}_a$ is never decremented, then $\prnd_{t,a} = q^{(0)}_a \geq \p_{t,a}$, so $\frac{\prnd_{t,a}}{\p_{t,a}} \geq 1$.

Now suppose coordinate $a$ is decremented at least once, and consider the final time at which $a$ is decremented. Let $\tilde q = (\tilde q_a, \dots, \tilde q_{|\MC{A}|})$ denote the vector immediately before this decrement. By the definition of the decrement rule in line 5 of Algorithm \ref{alg:rounding-dec},
\begin{align*}
    a \in \argmax_{\alpha \in \MC{A}}
    \frac{\tilde q_{\alpha} - 1/m}{\p_{t,\alpha}}.
\end{align*}
Immediately after this decrement, and since this is the final decrement of coordinate $a$, $\frac{\prnd_{t,a}}{\p_{t,a}} = \frac{\tilde q_a - 1/m}{\p_{t,a}}$. Thus,
\begin{align*}
    \frac{\prnd_{t,a}}{\p_{t,a}} \geq
    \max_{\alpha \in \MC{A}} \frac{\tilde q_{\alpha} - 1/m}{\p_{t,\alpha}}.
\end{align*}
Since the algorithm only decreases $\tilde q_{\alpha}$ as it runs, the final rounded probability satisfies $\prnd_{t,\alpha} \leq \tilde q_{\alpha}$,
and therefore
\begin{align}
     \frac{\prnd_{t,a}}{\p_{t,a}} \geq
    \frac{\prnd_{t,\alpha} - 1/m}{\p_{t,\alpha}}
    \qquad \text{for all } \alpha \in \MC{A}.
    \label{eqn:rnd-inequality}
\end{align}
Thus, we have that
\begin{align*}
    \frac{\prnd_{t,a}}{\p_{t,a}}
    = \frac{\prnd_{t,a}}{\p_{t,a}} \sum_{\alpha \in \MC{A}} \p_{t,\alpha}
    \underbrace{\geq}_{(i)} \sum_{\alpha \in \MC{A}}
    \left(\prnd_{t,\alpha} - \frac{1}{m}\right)
    = 1 - \frac{|\MC{A}|}{m}.
\end{align*}
Inequality (i) above holds by \eqref{eqn:rnd-inequality} by multiplying both sides by $\prnd_{t,\alpha}$.
\end{proof}

\bibliography{main}

\end{document}